\documentclass{article}

\PassOptionsToPackage{numbers, compress}{natbib}

\usepackage[preprint]{neurips_2026}

\usepackage[utf8]{inputenc}
\usepackage[T1]{fontenc}
\usepackage{hyperref}
\usepackage{url}
\usepackage{booktabs}
\usepackage{amsfonts}
\usepackage{amsmath}
\usepackage{nicefrac}
\usepackage{microtype}
\usepackage{xcolor}
\usepackage{graphicx}
\usepackage{array}
\usepackage{tabularx}
\usepackage{placeins}
\usepackage{listings}
\usepackage[most]{tcolorbox}
\usepackage{longtable} 

\newcommand{\repopy}{https://github.com/powerzoojax/PowerZooPy}

\definecolor{lstframe}{HTML}{2C5F8C}
\definecolor{lstkw}{HTML}{0B5394}
\definecolor{lstcomment}{HTML}{6A7787}
\definecolor{lststring}{HTML}{8B5E00}

\lstdefinestyle{pzjpython}{%
  language=Python,
  basicstyle=\ttfamily\footnotesize,
  keywordstyle=\color{lstkw}\bfseries,
  commentstyle=\color{lstcomment}\itshape,
  stringstyle=\color{lststring},
  showstringspaces=false, breaklines=true, breakatwhitespace=false,
  columns=fullflexible, keepspaces=true, upquote=true,
  aboveskip=2pt, belowskip=2pt,
}

\lstdefinestyle{pzjyaml}{%
  basicstyle=\ttfamily\footnotesize,
  keywordstyle=\color{lstkw}\bfseries,
  commentstyle=\color{lstcomment}\itshape,
  stringstyle=\color{lststring},
  showstringspaces=false, breaklines=true,
  columns=fullflexible, keepspaces=true, upquote=true,
  morecomment=[l]{\#}, morestring=[b]",
  aboveskip=2pt, belowskip=2pt,
}

\lstdefinestyle{pzjbash}{%
  language=bash,
  basicstyle=\ttfamily\footnotesize,
  keywordstyle=\color{lstkw}\bfseries,
  commentstyle=\color{lstcomment}\itshape,
  showstringspaces=false, breaklines=true,
  columns=fullflexible, keepspaces=true, upquote=true,
  aboveskip=2pt, belowskip=2pt,
}

\lstnewenvironment{bashplain}{%
  \lstset{style=pzjbash, frame=leftline, framesep=4pt, framerule=1pt,
          rulecolor=\color{lstframe}, xleftmargin=10pt}%
}{}

\newtcolorbox{lstbox}[2][]{%
  enhanced, breakable,
  colback=white, colframe=lstframe,
  colbacktitle=lstframe, coltitle=white,
  fonttitle=\bfseries\small,
  arc=1.5mm, boxrule=0.6pt,
  title={#2},
  left=6pt, right=6pt, top=3pt, bottom=3pt,
  #1
}

\makeatletter
\@ifundefined{NCY}{\newcolumntype{Y}{>{\raggedright\arraybackslash}X}}{}
\makeatother

\title{PowerZooJax: A JAX-based Power System Benchmark for Reinforcement Learning}

\author{%
  \parbox{\dimexpr\textwidth-2\tabcolsep\relax}{\centering
    Zhanhua Pan$^{1\dagger}$, Xiao Liu$^{2\dagger}$, Zhilong Cao$^{1}$, Jianhong Wang$^{3\ddagger}$, Dawei Qiu$^{1\ddagger}$\thanks{Corresponding author: \texttt{dawei.qiu@ntu.edu.sg}. $^{\dagger}$These authors contributed equally. $^{\ddagger}$Co-supervisors.}\\[10pt]
    \normalfont\small
    $^{1}$Nanyang Technological University, Singapore. 
    $^{2}$Cornell University, USA. 
    $^{3}$University of Bristol, UK.}
}

\begin{document}

\maketitle

\begin{abstract}
Power system operation is a safety-critical sequential decision-making problem, making it a natural testbed for reinforcement learning (RL). However, existing RL environments for power systems are often narrow in scope and computationally limited by CPU-based simulation workflows, making large-scale evaluation difficult. We introduce PowerZooJax, a JAX-based benchmark suite for RL in power system operation. It provides five constrained Markov decision process tasks spanning generation, transmission, distribution, distributed energy resources, and data center microgrid. By rewriting power flow, economic dispatch, market clearing, and device dynamics as JAX computation graphs, PowerZooJax keeps the entire training and evaluation loop on the GPU. Experiments show substantial speedups over CPU-based simulations and demonstrate standardized evaluation of policy returns, safety violations, and out-of-distribution stress conditions. Our open-source benchmark is available at: \url{https://github.com/powerzoojax/PowerZooJax}.
\end{abstract}

\begin{figure}[h!]
    \centering
    \includegraphics[width=1.00\textwidth]{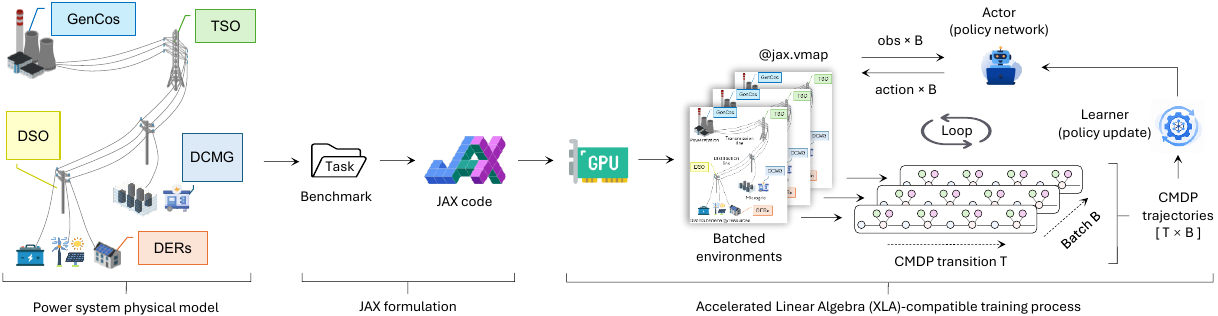}
    \caption{Execution paradigm of PowerZooJax.}
    \label{fig:paradigm} 
\end{figure}

\section{Introduction}
\label{sec:1}
Power systems are undergoing a fundamental transformation driven by the 3Ds: decarbonization, decentralization, and digitalization~\citep{HowDecarbonizationDigitalization2018disilvestre}. Traditionally, power systems were operated around centralized generation, long-distance transmission, and largely passive demand. This paradigm is increasingly challenged by the growing integration of renewable generation, distributed energy resources (DERs), battery storage, electric vehicle (EV) charging, flexible loads, and emerging critical infrastructure such as data centers~\citep{DecarbonizationElectricitySystems2021strbac}. Furthermore, the rise of artificial intelligence (AI) reinforces the central role of power systems: AI development depends on reliable, affordable, and sustainable electricity supply~\citep{EnergyPolicyConsiderations2019strubell}, while AI-driven data centers are becoming a major source of new electricity demand~\citep{EnvironmentalImpactNetzero2025xiao}.

Operating power systems requires sequential decision-making under uncertainty while satisfying strict physical, operational, and security constraints~\citep{TutorialStochasticOptimization2016powell, SecurityConstrainedUnitCommitment2023chen}. This structure makes power system operation a natural domain formulated as a constrained Markov decision process (CMDP) to study reinforcement learning (RL) algorithms. As national critical infrastructure, the power system is one of the most complex and safety-critical physical systems, where operational failures can disrupt economic activity and public services~\citep{SmallVulnerableSets2017yang, FragmentationOutageClusters2022wu}. System operators must continuously balance supply and demand, manage constrained power flows, maintain voltage and power quality, coordinate DERs, and ensure the reliable operation of local microgrids~\citep{Achieving100Renewable2017kroposki}.

On the other hand, applying RL to power system operation remains challenging from both modeling and computational perspectives. Specifically, power systems rely on different physical approximations and optimization formulations, ranging from DC optimal power flow (DC OPF) in transmission networks to AC optimal power flow (AC OPF) in distribution networks and device-level DER coordination at the grid edge. These differences motivate benchmarks that preserve task-specific power system physics while providing a consistent interface and standardized evaluation protocol for RL algorithms. Beyond these domain-specific challenges, existing power system RL environments are often limited by their computational structure. Most environments run power flow calculations and system dynamics through CPU-based simulators, while RL policies are commonly trained on GPUs. This separation makes large-scale training inefficient: rollouts are constrained by CPU simulation speed, and repeated CPU-GPU communication introduces additional overhead. As a result, evaluation over tremendous conditions, random seeds and trajectories will strongly delay the procedure of developing RL algorithms for real-world power system applications.

To mitigate this issue, we aim to develop a more efficient benchmark for evaluating RL algorithms in power system operation. Inspired by the recent proliferation of JAX-based RL environments~\citep{gymnax2022github,JaxMARLMultiAgentRL2024bandyopadhyay}, we introduce PowerZooJax, a novel power system RL benchmark fully implemented in JAX~\citep{jax2018github}, so it enables compiled and vectorized rollouts completely on GPUs. This design reduces the CPU-GPU communication overhead present in many existing power system RL benchmarks~\citep{PowerGymReinforcementLearning2022fan,SustainGymReinforcementLearning2023yeh,RL2GridBenchmarkingReinforcement2025marchesini}.

The main contributions of this work are threefold. First, we present a unified benchmark scope that covers five major tasks in daily power system operation, enabling RL algorithms to be evaluated across the entire power system decision-making problems. Second, we formulate these operational problems under a consistent CMDP interface, enabling systematic comparison of RL algorithms across five tasks with different physical models, operational constraints, and objectives. Third, we provide JAX-based benchmark environments, baseline RL algorithms, and evaluation metrics that support scalable and reproducible assessment of RL-based power system control methods. 

\section{Related Work}
\label{sec:2}
\textbf{Power System RL Benchmarks.}
Existing power system RL benchmarks cover different parts of the power grid. For grid operation, MAPDN~\citep{wang2021multi} studies voltage control, ANDES-Gym~\citep{Andes_gymVersatileEnvironment2022cui} supports dynamic power system simulation, PowerGridworld~\citep{PowerGridworld2022biagioni} examines distribution system control, and RL2Grid~\citep{RL2GridBenchmarkingReinforcement2025marchesini} and its multi-agent extension MARL2Grid-TR~\citep{MARL2GridTRMultiAgentRL2025marchesini} target transmission topology and redispatching control. At the end-use DER level, CityLearn v2~\citep{CityLearnV2OpenAI2023nweye}, GridLearn~\citep{GridLearnMultiagentReinforcement2022pigott}, and SustainGym~\citep{SustainGymReinforcementLearning2023yeh} provide environments for flexible consumption, building energy management, demand response, and grid-interactive energy communities. At the microgrid level, pymgrid~\citep{PymgridOpenSourcePython2020henri} and CommonPower~\citep{CommonPowerFrameworkSafe2025eichelbeck} focus on local generation, storage, and load scheduling. Collectively, these benchmarks demonstrate that power systems provide realistic sequential decision-making problems with uncertainty, physical constraints, and operational objectives. 

However, most existing benchmarks focus on a specific operational task. Many also rely on CPU-based solvers, making large-scale parallel rollouts costly: MARL2Grid-TR, for example, reports roughly 120{,}000 CPU hours to train and evaluate its baselines. In addition, they often lack standardized task protocols, violation accounting, and evaluation records, which makes cross-layer comparison and independent verification difficult. PowerZooJax addresses these limitations through a JAX-based architecture that provides physically grounded tasks across generation, transmission, distribution, end-use DER coordination, and microgrid operation. It further offers fixed task protocols, user-defined reward functions and safety costs, and reproducible evaluation records for scalable and standardized RL evaluation.

\textbf{JAX-based RL Benchmarks.}
Recent years have seen a growing number of RL benchmarks implemented in JAX, motivated by the need for fast, vectorized, and accelerator-compatible environment simulation. Brax~\citep{brax2021github} is an early and influential JAX-based benchmark for physics-based RL, demonstrating the benefits of differentiable and massively parallel simulation. Gymnax~\citep{gymnax2022github} further re-implements classic Gym control tasks in JAX, providing lightweight environments for benchmarking RL algorithms. Jumanji~\citep{bonnet2024jumanji} extends this direction to combinatorial optimization and structured decision-making problems, while Pgx~\citep{PgxHardwareAcceleratedParallel2023koyamada} focuses on classic board games.

Beyond single-agent RL, JAX-based benchmarks have also been developed for multi-agent and long-horizon settings. JaxMARL~\citep{JaxMARLMultiAgentRL2024bandyopadhyay} re-implements a collection of popular multi-agent RL environments in JAX, enabling accelerated simulation for MARL research. More recently, Craftax~\citep{CraftaxLightningFastBenchmark2024matthews} builds on ideas from MiniHack and Crafter to provide a fast grid-world benchmark for studying long-horizon exploration and generalization of RL algorithms. Octax~\citep{radji2025octax} re-implements a suite of CHIP-8 games in JAX for benchmarking different aspects of capabilities of RL algorithms.

PowerZooJax belongs to this growing family of JAX-based RL benchmarks, but targets a distinct and underrepresented domain: real-world power system operation. Unlike existing JAX benchmarks that primarily focus on classic control, combinatorial optimization, or games, PowerZooJax provides tasks grounded in practical power system operation while exposing core RL challenges such as partial observability, mixed action spaces, heterogeneous agents, long-horizon scheduling, and multi-objective control. In this sense, PowerZooJax complements existing JAX-based benchmarks by extending accelerator-compatible RL evaluation to large-scale, safety-critical power system operation.

\section{Background and Problem Setting}
\label{sec:3}
\begin{figure}[h!]
    \centering
    \includegraphics[width=1.00\textwidth]{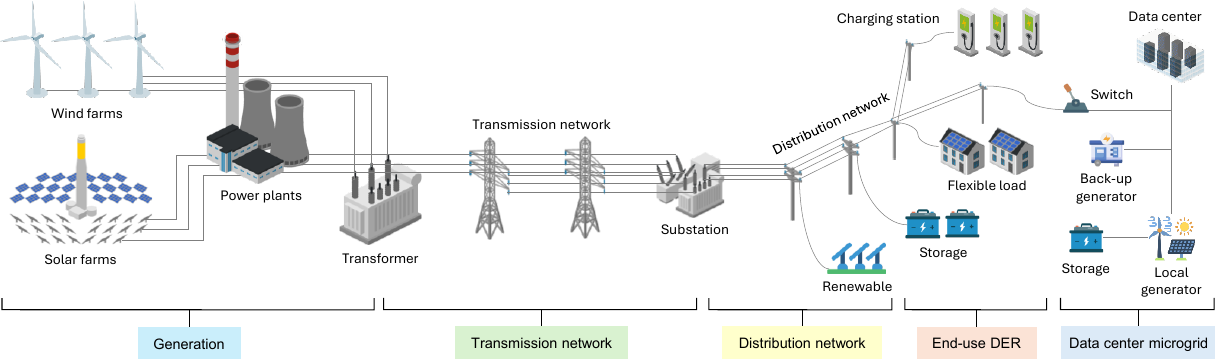}
    \caption{Overview of a modern power system model spanning generation, transmission, distribution, end-use DERs, and a data center microgrid.}
    \label{fig:powersystem} 
\end{figure}

The daily operation of a modern power system can be decomposed into five tasks, as illustrated in Figure~\ref{fig:powersystem}: generation, transmission, distribution, end-use DER, and data center microgrid. These tasks span dispatching conventional and renewable generation resources to meet time-varying demand, operating the high-voltage transmission network via DC OPF, operating the lower-voltage distribution network via AC OPF, coordinating user-side DER flexibility, and managing local generation, battery storage, backup resources, and critical loads in a data center microgrid.

Despite their different physical scales and modeling details, these five tasks share a common sequential decision-making structure under uncertainties and safety constraints, which can be formulated as constrained Markov decision processes (CMDPs)~\citep{ReinforcementLearningSelective2022chen}. At each step, the system state may include: demand, renewable availability, generator status, network conditions, voltage profiles, storage states of charge, and microgrid operating modes. Each agent's observation space is varied among different tasks. Agents select actions from task-specific action spaces, such as generation dispatch, power flow control, DER scheduling, storage charging or discharging, and data center workload scheduling. The system then evolves according to the physical power flow dynamics, device constraints, and stochastic exogenous factors (e.g. load variation and renewable uncertainty). Reward functions are designed by encoding operational objectives, including cost minimization, carbon reduction, and service continuity, while safety costs capture constraint violations such as power imbalance, voltage violations, and line overloads. The following five paragraphs provide an overview of the CMDP formulation for each task; the formal CMDP tuple, notation, and full details on state and action spaces, transition dynamics, reward functions, and safety costs are deferred to Appendix~\ref{app:C} and Appendix~\ref{app:E}.

\textbf{GenCos (Generation Companies) --- competitive market bidding.}
Five generation companies bid into the wholesale electricity market on the IEEE 5-bus transmission network \texttt{case5} over a 24-hour horizon at 30-minute resolution, using Great Britain (GB) demand profiles\footnote{\url{https://www.neso.energy/data-portal/historic-demand-data} National Energy System Operator (NESO).}. Each agent operating a generation company observes its own generation status and public market price signals, submits a piecewise-linear bid curve, and receives realized profit as the reward after security constrained economic dispatch (SCED)-based market clearing. Thermal line overloads are recorded as the safety cost. This is a competitive multi-agent RL task with partial observability and strategic interactions.

\textbf{TSO (Transmission System Operator) --- transmission unit commitment.}
A transmission system operator commits and dispatches 54 generators on the IEEE 118-bus transmission network \texttt{case118} using GB demand profiles over a 24-hour horizon at 30-minute resolution. The task enforces ramping limit, minimum up/down-time, and DC OPF feasibility constraints at each step. The transmission system operator agent observes the full system state and receives negative operating cost as the reward, while thermal line overloads and reserve shortfalls are recorded as two separate safety costs. This is a single-agent RL task with a large mixed continuous-binary action space.

\textbf{DSO (Distribution System Operator) --- distribution demand response.}
A distribution system operator coordinates six flexible loads on the IEEE 33-bus distribution network \texttt{case33bw}, using real Ausgrid distribution demand profiles\footnote{\url{https://www.ausgrid.com.au/Industry/Our-Research/Data-to-share/}, Ausgrid.} over a 24-hour horizon at 30-minute resolution. The distribution system operator agent observes the full system state, including bus voltages, branch flows, nodal demands, and deferred load-shifting buffers, and selects curtailment and load-shifting fractions for each controllable load. The reward is negative network loss, while voltage limit violations are recorded as the safety cost. This is a single-agent RL task for distribution-level demand response and voltage-aware load coordination.

\textbf{DERs (Distributed Energy Resources) --- cooperative DER coordination.}
Twelve DERs, including four batteries, four PV inverters\footnote{\url{https://bmrs.elexon.co.uk/generation-by-fuel-type}, Elexon.}, and four flexible loads, are placed at fixed buses of the 141-bus distribution network \texttt{case141}. Each distributed energy resource agent observes its own bus voltage, the voltages of its electrically connected neighboring buses, a compact system-wide voltage summary, and its local device state. Each agent takes a 2-dimensional device-specific action, such as active/reactive power control for batteries, curtailment/reactive power control for PV inverters, and curtailment/load shifting for flexible loads. All agents share a negative network loss reward, while voltage, thermal, and device feasibility violations are recorded as three separate safety costs. This is a cooperative multi-agent RL task with heterogeneous agents and local observations.

\textbf{DCMG (Data Center Microgrid) --- long-horizon multi-objective scheduling.}
A data center microgrid powers an AI data center using on-site PV, battery storage, a diesel generator, and a connection to the main grid. A microgrid agent controls the system over 288 five-minute steps using real Google workload traces~\citep{GoogleClusterdata2026}, jointly scheduling training and fine-tuning jobs, cooling, battery operation, diesel generation, and grid interaction based on workload, thermal, and energy states. Under normal conditions, the microgrid can operate in grid-connected mode; during disturbances or outages, it can switch to islanded mode to maintain critical load service. The reward combines energy use, operating cost, and carbon emissions, while power imbalance, server-zone over-temperature, and workload incompletion are recorded as three separate safety costs. This is a single-agent RL task for long-horizon, multi-objective scheduling of critical computing infrastructure.

\section{Power System Problem, RL Challenge and Evaluation Scope}
\label{sec:4}
From the perspective of RL, the five tasks can be viewed as either single-agent RL or multi-agent RL problems. 
This leads to one of our aims in proposing PowerZooJax: to reflect how RL techniques can address the emergent real-world problems, while providing a clean, unified testbed for AI researchers to easily pinpoint the key challenges of RL problems. 
Each task is grounded in a realistic operational setting, but is also designed to highlight a distinct RL challenge, such as partial observability, mixed action spaces, heterogeneous agents, long-horizon scheduling, or multi-objective control.

Table~\ref{tab:rl_difficulties} summarizes the interrelationship between power system problems embedded in the tasks we defined and the corresponding RL challenges in the modern RL literature. We categorize the tasks into centralized and decentralized settings according to their operational roles. Centralized tasks correspond to decisions that are typically executed by system operators to maintain secure and reliable network operation, while decentralized tasks correspond to settings in which multiple users, assets, or market participants make individual decisions based on local information.

From the perspective of RL evaluation, Table~\ref{tab:rl_difficulties} clearly captures the basic assumptions of popular RL problems for each task. As a result, researchers can confidently get the information about what features of their algorithms (RL challenges) are expected to be meaningfully assessed. More importantly, due to that our benchmark tasks follow the practical settings of power system problems, algorithms that work well on our benchmark can be explained as possessing the potential to handle real-world problems. This is what most of previous benchmarks that focused on either games or simplified grid-world settings lack.

\begin{table}[h!]
\centering
\small
\setlength{\tabcolsep}{4pt}
\caption{Interrelationship between power system problems and RL challenges.}
\label{tab:rl_difficulties}
\begin{tabular}{@{}lp{6.4cm}p{6.0cm}@{}}
    \toprule
    Tasks & Power System Problems & RL Challenges \\
    \midrule
    GenCos & \textbf{Decentralized} electricity market bidding with a 3-dim action space and a 12-dim observation space. & \textbf{Multi-agent RL} with partial observability (each agent's local information). \\
    \midrule
    TSO & \textbf{Centralized} unit commitment with a 108-dim action space and a 410-dim observation space. & \textbf{Single-agent RL} with a large, mixed action space of continuous and binary variables. \\
    \midrule
    DSO & \textbf{Centralized} control of flexible loads with a 12-dim action space and a 195-dim observation space. & \textbf{Single-agent RL} with an easy-to-start setup to verify initial design ideas. \\
    \midrule
    DERs & \textbf{Decentralized} control of three types of DERs (battery, PV, flexible loads) with a 2-dim action space and a 15-dim observation space. & \textbf{Multi-agent RL} with partial observability (each agent's local and neighbor information) and heterogeneous agents. \\
    \midrule
    DCMG & \textbf{Centralized} long-horizon scheduling with a 5-dim action space and a 24-dim observation space. & \textbf{Single-agent RL} with a long-horizon sequential decision making process and multi-objectives. \\
    \bottomrule
\end{tabular}
\begin{flushleft}
\footnotesize \textbf{Note:} ``dim'' is short for dimensional.
\end{flushleft}
\end{table}

\section{How Does PowerZooJax Work?}
\label{sec:5}
Each of the five tasks in Section~\ref{sec:3} runs a power flow (PF) or optimal power flow (OPF) computation at every environment step. DSO and DERs use radial AC PF, TSO and GenCos rely on DC OPF clearing, and DCMG combines local power-balance checks with device dynamics. These computations involve fixed-shape numerical operations, such as matrix updates, constraint evaluation, and batched linear-algebra routines, which are well suited to compilation and fusion by accelerated linear algebra (XLA). Conventional pipelines typically write them as host-side Python loops that traverse a feeder tree or call a CPU solver for a linear program, incurring a host round trip at every step~\citep{LearningRunPower2021marot}. In contrast, our PowerZooJax rewrites each iteration as a fixed-shape JAX computation graph and runs the full rollout, including batched parallel scenarios, inside a compiled accelerator program, as shown in Figure~\ref{fig:paradigm}.

This design enables PowerZooJax to keep the whole training loop on the GPU. From episode reset, through transitions, rewards, safety costs, action sampling, and gradient updates, the main computation operates on GPU-resident arrays. Episode boundaries do not require returning to a Python control loop: when one episode terminates, the next reset can be executed as the next operation in the same compiled JAX program. As a result, PowerZooJax avoids per-step CPU-GPU synchronization during rollout generation, reducing overhead and enabling efficient large-scale parallel training.

\subsection{Rewriting Power Flow as a JAX Computation Graph}
\label{sec:5.1}
A traditional radial PF solver, e.g.\ as implemented in pandapower~\citep{PandapowerOpenSourcePython2018thurner}, is a host-side fixed-point iteration. At each iteration $k$, the solver walks the feeder tree backward to aggregate downstream loads into branch flows $\mathbf{P}^{(k)},\mathbf{Q}^{(k)}$, walks it forward to update bus voltages $\mathbf{v}^{2,(k)}$, and checks convergence condition $\max_n |\mathbf{v}^{2,(k)}_n - \mathbf{v}^{2,(k-1)}_n| < \varepsilon$ inside a Python \texttt{while} loop. Since both sweeps and the convergence test are controlled on the host, each RL environment step incurs host-side control overhead and potentially a device-host synchronization. PowerZooJax rewrites the same iterative procedure as a fixed-shape JAX computation graph through three structural changes. In each change below, the expression on the left of $\rightarrow$ denotes the traditional implementation pattern, while the expression on the right denotes the PowerZooJax replacement.

\textbf{Tree walking $\rightarrow$ dense matrix-vector products.}
In a radial network, the backward and forward sweeps can be represented using fixed topology matrices. PowerZooJax precomputes two such matrices at environment construction time: a downstream-aggregation matrix $\mathbf{D}\in\mathbb{R}^{N_\ell\times N}$, where $N$ and $N_\ell$ are the numbers of buses and branches and $\mathbf{D}_{\ell,n}=1$ if bus $n$ is downstream of line $\ell$, and a path matrix $\boldsymbol{\Pi}\in\mathbb{R}^{N\times N_\ell}$, where $\boldsymbol{\Pi}_{n,\ell}=1$ if line $\ell$ lies on the unique path from the slack bus to bus $n$. The backward sweep then becomes a matrix-vector multiplication that aggregates downstream active and reactive loads into branch flows, while the forward sweep becomes a matrix-vector multiplication that accumulates voltage drops along each root-to-bus path. One PF iteration can therefore be written as
\begin{equation}
\label{eq:pf-graph}
\begin{aligned}
    &\mathbf{P}^{(k)} = \mathbf{D}\,\bigl(\mathbf{p}^{\mathrm{load}} + \mathbf{p}^{\mathrm{loss},(k-1)}\bigr), \\
    &\mathbf{v}^{2,(k)} = V_{\mathrm{slack}}^{2}\,\mathbf{1} - \boldsymbol{\Pi}\,\boldsymbol{\delta}^{2,(k)},
\end{aligned}
\end{equation}
with $\mathbf{Q}^{(k)}$ computed analogously. Here, $\mathbf{p}^{\mathrm{load}}$ is the vector of nodal active loads, $\mathbf{p}^{\mathrm{loss},(k-1)}$ is the per-bus loss correction recomputed from the previous iterate, and $\boldsymbol{\delta}^{2,(k)}$ denotes the per-line squared-voltage drop computed from $(\mathbf{P}^{(k)},\mathbf{Q}^{(k)},\mathbf{v}^{2,(k-1)})$ using the standard DistFlow recurrence~\citep{NetworkReconfigurationDistribution1989baran}. Since $\mathbf{D}$ and $\boldsymbol{\Pi}$ depend only on the network topology, they are precomputed once and reused throughout training. This replaces explicit feeder-tree traversal with fixed-shape matrix-vector products that are directly compatible with JAX compilation and batching.

\textbf{Python \texttt{while} $\rightarrow$ \texttt{jax.lax.while\_loop}.}
Equation~\eqref{eq:pf-graph} defines a fixed-point iteration body $f:(\mathbf{P},\mathbf{Q},\mathbf{v}^{2})\mapsto(\mathbf{P}',\mathbf{Q}',\mathbf{v}'^{2})$, whose output depends only on its input arrays. In PowerZooJax, this body is implemented entirely with \texttt{jnp} and \texttt{lax} operations, without host-side mutation or external state updates. JAX therefore traces the iteration body once into a static XLA computation graph and runs the fixed-point loop with \texttt{jax.lax.while\_loop}, replacing the host-controlled Python \texttt{while} loop with a device-side loop that can be compiled, fused, and batched across parallel environments.

\textbf{Per-step solver call $\rightarrow$ batched PF rollout via \texttt{jax.vmap}/\texttt{jax.lax.scan}.}
Once the radial PF iteration is written as fixed-shape matrix operations and a device-side \texttt{lax.while\_loop}, PowerZooJax can evaluate many PF solves without launching a separate solver at every environment step. The precomputed topology matrices $\mathbf{D}$ and $\boldsymbol{\Pi}$ are shared across batched scenarios, so the same PF kernel is applied in parallel over a leading batch dimension with \texttt{vmap}. Across time, \texttt{lax.scan} composes the per-step PF solve with device updates, rewards, and safety-cost calculations, producing a single compiled rollout for an entire RL episode. In our tasks, this means that the 48 PF evaluations required by one DSO or DERs episode are executed as part of one accelerator-resident computation graph, rather than as repeated Python calls to an external solver.

\subsection{Optimal Power Flow as a Computation Graph}
\label{sec:5.2}
The same computation-graph principle applies to OPF kernels. We illustrate this with DC OPF, which is used by the TSO and GenCos tasks. DC OPF is a linear program that minimizes generation cost subject to system-wide power balance, generator capacity bounds, and transmission line-flow limits. In a conventional pipeline, this linear program is passed to an external CPU solver once per environment step, making the OPF solve a host-side bottleneck. PowerZooJax instead implements OPF solvers directly inside the JAX graph: an ADMM~\citep{DistributedOptimalPower2014erseghe} solver for TSO's DC OPF, and a PDIPM~\citep{ComputationalIssuesMarketBased2007wang} solver for GenCos market clearing. For example, one iteration of the ADMM DC OPF solver can be written as
\begin{equation}
\label{eq:opf-graph}
\begin{aligned}
    &\begin{bmatrix} \mathbf{p}^{(k+1)} \\ \nu^{(k+1)}_{\mathrm{eq}} \end{bmatrix} = A_{\mathrm{aug}}^{-1}\begin{bmatrix} \mathbf{r}^{(k)} \\ \mathbf{1}^\top\mathbf{p}^{d} \end{bmatrix},\\
    &\mathbf{z}_1^{(k+1)} = \mathrm{clip}\bigl(\mathbf{p}^{(k+1)} + \mathbf{y}_1^{(k)}/\rho,\ \underline{\mathbf{p}},\ \overline{\mathbf{p}}\bigr),
\end{aligned}
\end{equation}
where $\mathbf{p}$ is the generator dispatch vector with output bounds $(\underline{\mathbf{p}},\overline{\mathbf{p}})$, $\mathbf{p}^{d}$ is the nodal demand vector, $\nu_{\mathrm{eq}}$ is the multiplier on the system-wide power-balance equality, $\mathbf{z}_1$ and $\mathbf{y}_1$ are the ADMM auxiliary and dual variables associated with the generator bounds, $\rho>0$ is the ADMM penalty parameter, and $\mathbf{r}^{(k)}$ is the residual vector assembled from the current auxiliary and dual iterates. The projection step for the line-flow auxiliary variable $\mathbf{z}_2$ and the dual updates for $\mathbf{y}_1,\mathbf{y}_2$ complete one iteration (Appendix~\ref{app:D.2}, Equation~\eqref{eq:admm}). The augmented KKT matrix $A_{\mathrm{aug}}$ depends only on grid parameters and does not change with the per-step load, so its inverse is precomputed once at setup, similar to the topology matrices $\mathbf{D}$ and $\boldsymbol{\Pi}$ used in the PF kernel. Each ADMM iteration is then reduced to fixed-shape matrix operations and elementwise projections, making the iteration body fully traceable by JAX. The complete OPF solve is executed as a fixed-length \texttt{lax.fori\_loop}, so batched OPF solves can be compiled and run on GPU device without repeated calls to an external CPU solver.

\section{Evaluation}
\label{sec:6}
We evaluate PowerZooJax on the five benchmark tasks introduced in Section~\ref{sec:3}. Section~\ref{sec:6.1} first evaluates the computational efficiency of the JAX-based implementation described in Section~\ref{sec:5}, measuring wall-clock speedups under matched training budgets. Sections~\ref{sec:6.2}--\ref{sec:6.4} then present representative task-level results for GenCos, TSO, and DCMG, including training returns, safety violations, comparisons with task-specific baselines, and performance on out-of-distribution (OOD) splits. Analogous results for DSO and DERs follow the same protocol and are reported in Appendix~\ref{app:I.3}--\ref{app:I.4}.

\textbf{Hardware.} All experiments are run on a single workstation equipped with an NVIDIA RTX~4500 Ada (24~GB) GPU and an AMD Ryzen Threadripper PRO 7985WX host with 256~GB of RAM.

\textbf{Evaluation Setup.} We evaluate each task with 5 random seeds. Training budgets are task-specific: 5M environment steps for GenCos, 20M for TSO, 3M for DSO, 10M for DERs, and 1M for DCMG. We use PPO~\citep{ProximalPolicyOptimization2017schulman} for the single-agent tasks (TSO, DSO, and DCMG), and IPPO~\citep{IndependentLearningAll2020witt} for the multi-agent tasks (GenCos and DERs). DCMG is additionally trained with SAC~\citep{SoftActorCriticAlgorithms2018haarnoja} for the representative episode in Section~\ref{sec:6.4}. After training, each seed is evaluated on a held-out split, with task-specific OOD stress splits when applicable. For each seed, results are averaged over 10--50 evaluation episodes depending on the task, and final results are reported as mean values with 95\% CIs across seeds. Full hyperparameters and network architectures are in Appendix~\ref{app:H}.

\subsection{Execution Speed and Scaling Performance}
\label{sec:6.1}
We compare three training backends on the five benchmark tasks: (i) PowerZooJax, which runs both the environment simulation and policy training in JAX on the GPU; (ii) Stable-Baselines3 (SB3)~\citep{stable-baselines3}, which uses PowerZooPy\footnote{\url{\repopy}, the companion object-oriented Python implementation.}, our CPU-based Python implementation of the same tasks, for environment simulation and trains a PyTorch policy on the GPU; and (iii) SBX~\citep{AraffinSbx2026raffin}, the JAX port of SB3, which trains a JAX policy on the GPU while using the same CPU-based PowerZooPy environment simulation.
 
\begin{figure}[h!]
  \centering
  \includegraphics[width=1.00\linewidth]{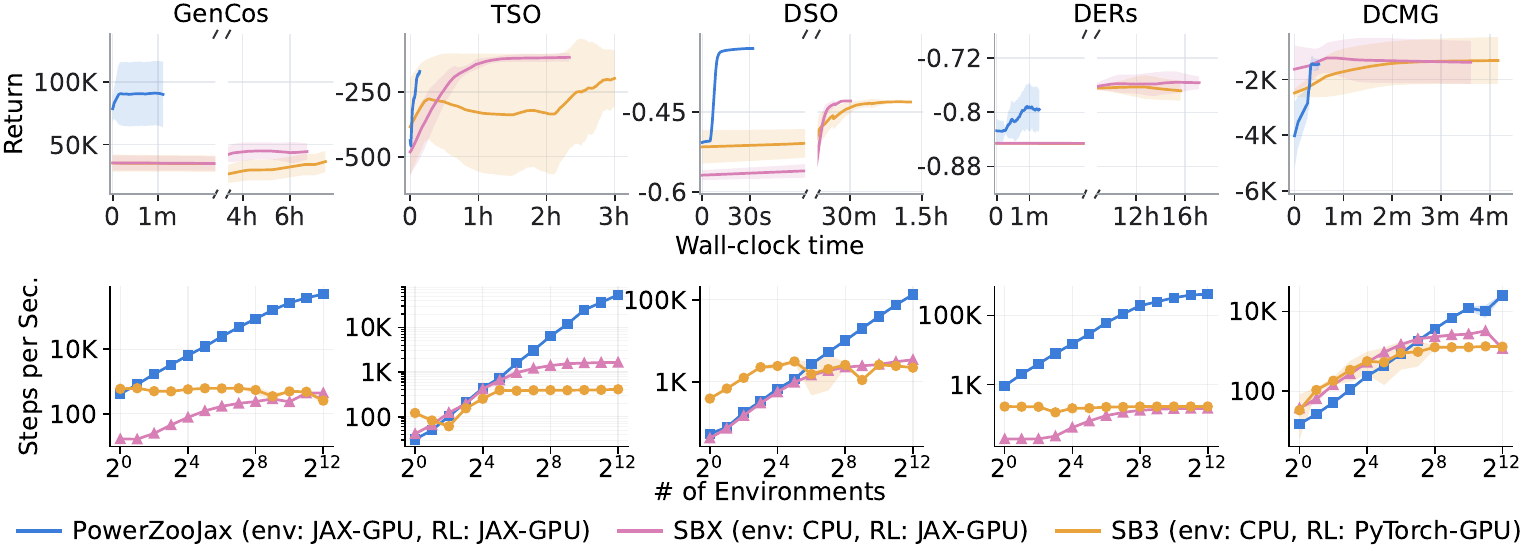}
  \caption{Cross-backend speed for five benchmark tasks. Top: wall-clock training time under matched budgets. Bottom: environment throughput with increasing parallelism, shown on log--log axes.}
  \label{fig:speed}
\end{figure}
\begin{table}[h!]
\centering
\small
\setlength{\tabcolsep}{3pt}
\renewcommand{\arraystretch}{0.95} 
\caption{Cross-backend speed per benchmark task for three training backends.}
\label{tab:speed}
\begin{tabular}{l r r r r r r r r r r}
    \toprule
    & & \multicolumn{5}{c}{Wall-clock to matched budget (s)} & \multicolumn{4}{c}{Throughput at $2^{12}$ envs (steps/s)} \\
    \cmidrule(lr){3-7} \cmidrule(lr){8-11}
    Task & Budget & $N_\text{envs}$ & PowerZooJax & SBX & SB3 & speedup & PowerZooJax & SBX & SB3 & speedup \\
    \midrule
    GenCos & 5M  & 256 & 67  & 9{,}602  & 10{,}868 & 161$\times$ & 506{,}134 & 442     & 244     & 2{,}077$\times$ \\
    TSO    & 20M & 256 & 515 & 10{,}799 & 10{,}801 & 21$\times$  & 53{,}724  & 1{,}649 & 411     & 131$\times$    \\
    DSO    & 3M  & 128 & 32  & 1{,}667  & 3{,}801  & 117$\times$ & 134{,}660 & 3{,}435 & 2{,}207 & 61$\times$     \\
    DERs   & 10M & 128 & 79  & 61{,}883 & 59{,}778 & 780$\times$ & 414{,}506 & 203     & 233     & 2{,}040$\times$ \\
    DCMG   & 1M  & 64  & 33  & 221      & 257      & 8$\times$   & 25{,}132  & 1{,}167 & 1{,}298 & 22$\times$    \\
    \bottomrule
\end{tabular}
\end{table}

\textbf{Wall-clock and Throughput.} Figure~\ref{fig:speed} evaluates speed from two perspectives. The wall-clock measures how long each backend takes to finish the same training budget for each task, including episode resets, periodic evaluations, and gradient updates. The throughput measures how many environment steps each backend can generate per second as the number of parallel environments increases from $2^{0}$ to $2^{12}$. Table~\ref{tab:speed} reports the final wall-clock time and the throughput at $2^{12}$ parallel environments, and the corresponding speedup of PowerZooJax over the CPU-based baselines.

\textbf{Speedup vs Per-step Cost.} The largest gains occur when CPU baselines perform solver-heavy work. GenCos and DERs call CPU-based OPF or PF routines in the CPU implementations, while PowerZooJax executes these kernels inside the compiled JAX graph, yielding throughput speedups above $2{,}000\times$. TSO and DSO implemented in JAX also benefit substantially from moving power system computation onto the GPU. DCMG has the smallest speedup because its transitions are mostly arithmetic device updates without an external solver call, so the CPU baseline is already fast. 

\subsection{GenCos: Bidding under Partial Observability}
\label{sec:6.2}
\begin{figure}[h!]
  \centering
  \includegraphics[width=1.00\linewidth]{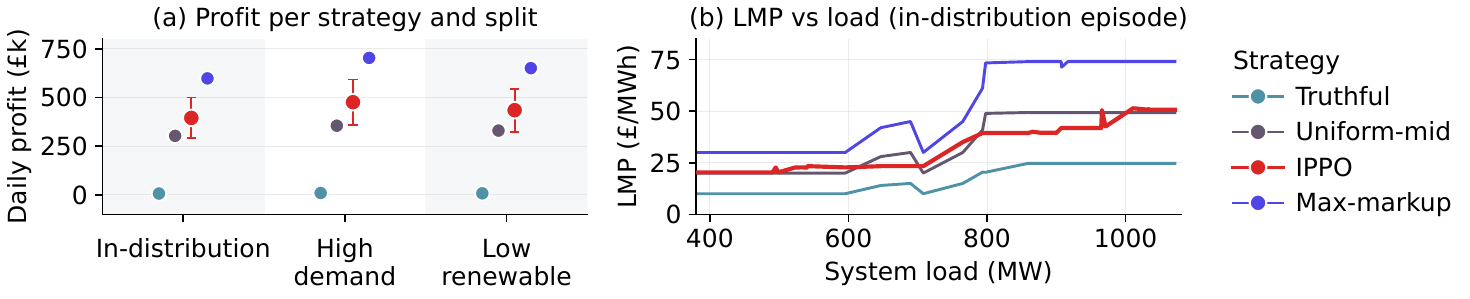}
  \caption{Generation company profit and market clearing price on \texttt{case5}. (a) Total daily profit for each strategy and evaluation split. (b) Locational marginal price (LMP) vs system load on a representative evaluation. Error bars in (a) show 95\% CIs across 5 seeds.}
  \label{fig:gencos}
\end{figure}

Figure~\ref{fig:gencos} evaluates GenCos under four bidding strategies and three evaluation splits. The two extreme baselines define the strategic range: Truthful bidding at marginal cost achieves a daily profit of \pounds 6.9k/day, while Max-markup bidding at the maximum allowed price achieves \pounds 598k/day. Uniform-mid, which uses an intermediate markup between Truthful and Max-markup bidding, achieves \pounds 303k/day. IPPO learns a more profitable strategy than Uniform-mid, reaching \pounds 395k/day on the in-distribution split, while remaining below the Max-markup upper baseline. The same ordering is preserved under the two OOD stress splits: high demand, which increases load by 10\%, and low renewable, which decreases wind generation by 5\%. The learned strategy also produces distinct locational marginal prices (LMPs). When system load exceeds 800~MW, Max-markup clears at an LMP near \pounds 75/MWh, Uniform-mid near \pounds 50/MWh, and IPPO around \pounds 45/MWh, indicating that IPPO learns to increase market profit while avoiding the most aggressive markup strategy.

\subsection{TSO: Economic--Safety Cost Trade-off}
\label{sec:6.3}
\begin{figure}[h!]
  \centering
  \includegraphics[width=1.00\linewidth]{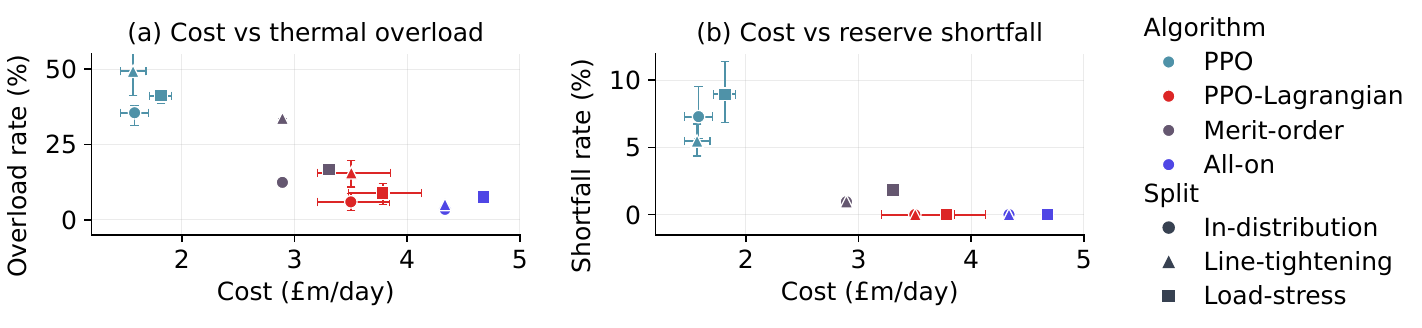}
  \caption{TSO economic--safety cost trade-off on \texttt{case118}. (a) Operating cost vs thermal overload rate. (b) Operating cost vs reserve shortfall rate. Error bars show 95\% CIs across 5 seeds.}
  \label{fig:tso}
\end{figure}

Figure~\ref{fig:tso} shows the economic--safety cost trade-off for transmission operation. The merit-order baseline, which commits generators from cheapest to most expensive until capacity covers demand plus a 5\% reserve margin, while ignoring thermal line limits, achieves an operating cost of \pounds 2.89M/day. The all-on baseline keeps every generator online, improving safety but increasing cost to \pounds 4.34M/day. PPO, trained with negative operating cost as the reward, reduces to \pounds 1.58M/day but violates thermal line limits in 36\% and reserve margins in 7\% over all evaluations. PPO-Lagrangian~\citep{ray2019benchmarking}, which applies adaptive penalties to each violation type, eliminates reserve shortfalls and reduces the thermal overload rate to 6\%, at a higher cost of \pounds 3.50M/day. The same pattern holds under the two OOD stress splits: line-tightening, where line ratings are reduced to 85\%, and load-stress, where demand is scaled by 1.15. Under these stresses, PPO reaches thermal overload rates of 49\% and 41\%, respectively, while PPO-Lagrangian reduces them to 16\% and 9\% and maintains zero reserve shortfall. These results show that the Lagrangian formulation substantially improves safety, fully eliminating reserve violations on this network, but thermal overloads remain a harder constraint to satisfy under stressed conditions.

\subsection{DCMG: Long-Horizon Scheduling for a Data Center}
\label{sec:6.4}
\begin{figure}[h!]
  \centering
  \includegraphics[width=1.00\linewidth]{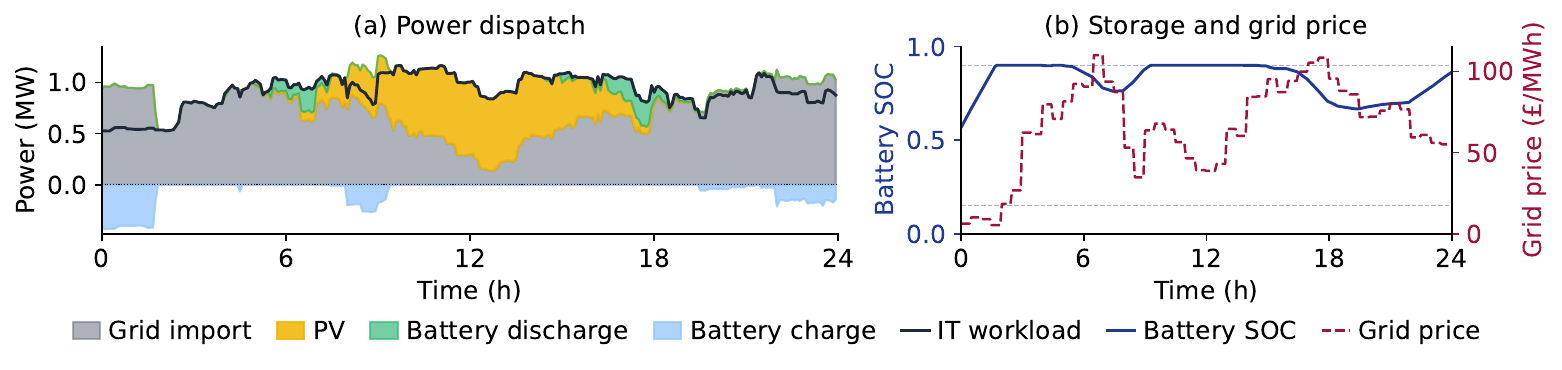}
  \caption{DCMG daily scheduling on a representative day. (a) Power supply to the IT workload. (b) Daily battery state of charge (SoC) and grid electricity price.}
  \label{fig:dcmg}
\end{figure}

Figure~\ref{fig:dcmg} shows a representative 24-hour SAC policy trajectory over 288 five-minute steps. SAC charges the battery from the grid during the first two hours when electricity prices are low, discharges the battery through the morning price peak above \pounds 100/MWh, recharges during the midday when PV generation is available, and cycles again through the evening peak. This repeated charge--discharge pattern indicates that SAC learns to coordinate storage schedules across the full 288-step horizon. Throughout the episode, the IT workload is served at every step, showing that SAC learns a price-aware and constraint-respecting scheduling policy directly from the multi-objective reward.

\section{Conclusion}
\label{sec:7}
PowerZooJax shows that power system RL benchmarks can simultaneously balance physical modeling and computational scalability. By integrating training and evaluation into a GPU-accelerated pipeline, the benchmark supports fast rollouts while retaining operational metrics that matter for real grid decision making. The five tasks demonstrate how RL algorithms can be evaluated across diverse power system settings under a common constrained Markov decision process (CMDP) interface, with standardized reporting of returns, safety costs, and stress-test performance.

\textbf{Limitations.} PowerZooJax is a benchmark suite, not a deployment-ready power system simulator or controller. Although it covers five representative tasks, the current version is still limited in grid scale, uncertainty modeling, operational constraints, and task diversity. Finally, the current environments are built from public or synthetic grid models and public data, and therefore cannot fully capture the complexity of real power system with real-time operational data. Strong performance on PowerZooJax should therefore be interpreted as evidence of an algorithm's ability to address representative power system RL challenges, not as a guarantee of safe real-world power system operation.

\textbf{Future Work.} Future work will extend PowerZooJax in three directions. First, we plan to broaden PowerZooJax with larger grid cases, richer uncertainty models, more realistic constraints, and additional tasks. Second, we envision PowerZooJax evolving into a more intelligent programming tool for power system RL. By integrating PowerZooJax with language models, users could specify power system requirements in natural language, and the tool could automatically generate the corresponding RL environment, task configuration, reward and safety cost definitions, and baseline algorithms. Third, we aim to collaborate with the power industry to access grid models and operational data, build digital-twin environments, and study how AI-based control methods can be evaluated and transferred toward practical power system applications.

\section*{Acknowledgement}
Dawei Qiu is supported by the Nanyang Technological University (NTU) Start-Up Grant project \#026749-00001 “Market Design for Low-Carbon Power Systems: Towards a Reliable, Affordable, and Resilient Transition”. Jianhong Wang is supported by the Engineering and Physical Sciences Research Council (EPSRC) [Grant Ref: EP/Y028732/1].

\newpage
\bibliographystyle{unsrtnat}
\bibliography{references}

\newpage
\appendix
\section*{Appendix}
\label{app:0}
This supplementary material is organized as follows.
Section~\ref{app:C} formalizes the CMDP tuple shared by all five tasks and fixes notation.
Section~\ref{app:D} gives the physical-model derivations (radial AC power flow, DC OPF with ADMM, bid-based market clearing with PDIPM, and the device plus microgrid couplings).
Section~\ref{app:E} gives the per-task observation, action, reward, and cost equations.
Section~\ref{app:F} documents external data sources and the split and stress-test taxonomy.
Section~\ref{app:G} provides a quickstart with install, a minimal Python rollout, configuration excerpts, and CLI commands.
Section~\ref{app:H} lists algorithm choices, hyperparameters, network architectures, and evaluation protocol.
Section~\ref{app:I} reports the per-task supplementary tables and figures.
Section~\ref{app:J} reports execution scaling and the backend comparison.
\medskip

\section{Constrained Markov Decision Process Formulation}
\label{app:C}

This section gives the formal constrained Markov decision process (CMDP) tuple shared by all five tasks, the two constraint forms used in the benchmark (discounted expected cost and zero-violation constraint), and the Lagrangian formulation used by the safe-RL baselines. The notation used throughout the appendix is fixed in Table~\ref{tab:notation} (Section~\ref{app:C.5}).

\subsection{The CMDP tuple}
\label{app:C.1}

Each task is an episodic CMDP \citep{ReinforcementLearningSelective2022chen,ConstrainedPolicyOptimization2017achiama}
\begin{equation}
\label{eq:cmdp-tuple}
    \mathcal{M} = \bigl(\mathcal{S},\ \mathcal{A},\ P,\ r,\ \mathbf{c},\ \gamma,\ \mathbf{b},\ \rho_0,\ T\bigr),
\end{equation}
where $\mathcal{S}$ and $\mathcal{A}$ are the state and action spaces, $P:\mathcal{S}\times\mathcal{A}\to\Delta(\mathcal{S})$ is the transition kernel, $r:\mathcal{S}\times\mathcal{A}\to\mathbb{R}$ is the scalar reward, $\mathbf{c}=(c_1,\dots,c_k):\mathcal{S}\times\mathcal{A}\to\mathbb{R}^k$ is the cost vector, $\gamma\in(0,1]$ is the discount factor, $\mathbf{b}\in\mathbb{R}_{\ge 0}^k$ is the constraint-threshold vector, $\rho_0$ is the initial-state distribution, and $T$ is the episode horizon (Table~\ref{tab:task-overview} lists per-task $T$). Reward and cost are produced as separate outputs by every task environment so that safety violations cannot be silently absorbed into the scalar return; the cost-vector order is fixed by the declared constraint names of each task. In the multi-agent tasks (GenCos and DERs), each agent observes only a projection of the state (Section~\ref{app:E}), so these tasks are constrained partially observable problems.

For a stationary policy $\pi:\mathcal{S}\to\Delta(\mathcal{A})$, define the reward and cost values
\begin{align}
  J_r(\pi) &\;=\; \mathbb{E}_\pi\Bigl[\textstyle\sum_{t=0}^{T-1}\gamma^t\,r(s_t,a_t)\Bigr], \label{eq:cmdp-Jr}\\
  J_{c_i}(\pi) &\;=\; \mathbb{E}_\pi\Bigl[\textstyle\sum_{t=0}^{T-1}\gamma^t\,c_i(s_t,a_t)\Bigr], \quad i=1,\dots,k. \label{eq:cmdp-Jc}
\end{align}

\subsection{Constraint forms used in the suite}
\label{app:C.2}

PowerZooJax tasks use one of two constraint forms. \textbf{Form 1 (budgeted expected cost)} is applied to the DSO voltage channel, the three DERs channels (voltage, thermal, resource), and the three DCMG channels (workload incompletion, server-zone over-temperature, power imbalance), as well as the TSO Lagrangian baselines. \textbf{Form 2 (zero-violation constraint)} is applied to the TSO reserve-shortfall and thermal-overload channels reported in Table~\ref{tab:tso-frontier-appendix}. The single GenCos thermal-overload channel is reported as a market-side diagnostic only.

\textbf{Form 1 (discounted expected-cost budget).}
\begin{equation}
\label{eq:cmdp-form1}
  \max_\pi\; J_r(\pi) \quad\text{s.t.}\quad J_{c_i}(\pi) \le b_i,\ i=1,\dots,k.
\end{equation}
Used by DSO, DERs, and TSO Lagrangian baselines, where the $b_i$ are operational risk budgets per episode.

\textbf{Form 2 (zero-violation constraint).}
\begin{equation}
\label{eq:cmdp-form2}
  \max_\pi\; J_r(\pi) \quad\text{s.t.}\quad \mathbb{P}_\pi\!\bigl[c_i(s_t,a_t) > 0\bigr] = 0,\ \forall t,\ \forall i\in I_\text{hard}.
\end{equation}
Used for the TSO reserve-shortfall and thermal-overload channels: a policy satisfies Form~2 only if both channels record zero cost at every step, so cost cannot be traded for safety. The TSO results in Section~\ref{app:I.2} are reported under this form.

\subsection{Lagrangian formulation}
\label{app:C.3}

For Form~1 with multipliers $\boldsymbol{\lambda}\in\mathbb{R}_{\ge 0}^k$, the Lagrangian
\begin{equation}
\label{eq:cmdp-lagrangian}
  \mathcal{L}(\pi,\boldsymbol{\lambda}) \;=\; J_r(\pi) \;-\; \boldsymbol{\lambda}^{\!\top}\bigl(J_{c}(\pi) - \mathbf{b}\bigr)
\end{equation}
gives the saddle-point reformulation $\max_\pi\min_{\boldsymbol{\lambda}\ge 0}\mathcal{L}(\pi,\boldsymbol{\lambda})$. PPO-Lagrangian and IPPO-Lagrangian baselines update $\boldsymbol{\lambda}$ by stochastic ascent on $J_{c}(\pi) - \mathbf{b}$ while $\pi$ is updated by PPO clipped-objective ascent on $\mathcal{L}$. Sauté PPO~\citep{SauteRLAlmost2022sootla} instead augments the observation with the per-channel remaining cost budget $b_i - \sum_{t'<t}\gamma^{t'}c_i(s_{t'},a_{t'})$ and trains plain PPO on the augmented state.

\subsection{Per-task instantiation}
\label{app:C.4}

Each task instantiates $\mathcal{M}$ through a \texttt{TaskSpec}/\texttt{ConstraintSpec} pair (the \texttt{TaskSpec} Protocol surface is shown in Section~\ref{app:G.2}). Concrete observation, action, reward, and cost equations are given in the task cards of Section~\ref{app:E}; per-task safety thresholds are part of each task's configuration file, and split definitions are listed in Section~\ref{app:F}.

\subsection{Notation}
\label{app:C.5}

\begingroup
\footnotesize
\setlength{\LTleft}{0pt}
\setlength{\LTright}{0pt}
\begin{longtable}{@{}>{\raggedright\arraybackslash}p{0.30\linewidth}>{\raggedright\arraybackslash}p{0.66\linewidth}@{}}
    \caption{Notation used across the main paper and the appendix.}
    \label{tab:notation}\\
    \toprule
    Symbol & Meaning \\
    \midrule
    \endfirsthead
    \toprule
    Symbol & Meaning \\
    \midrule
    \endhead
    \midrule
    \multicolumn{2}{r}{}\\
    \endfoot
    \bottomrule
    \endlastfoot
    \multicolumn{2}{l}{\emph{CMDP and policy}} \\
    $\mathcal{S},\,\mathcal{A}$ & State and action spaces of one task. \\
    $P$ & Transition kernel; \emph{deterministic given exogenous time-series}. \\
    $r,\,\mathbf{c}\in\mathbb{R}^k$ & Scalar reward and per-step cost vector with $k$ task-specific entries. \\
    $\gamma,\,T$ & Discount factor and episode horizon ($T=48$ or $T=288$). \\
    $\mathbf{b}\in\mathbb{R}_{\ge 0}^k$ & Cost thresholds (Form~1) or zero-violation set (Form~2). \\
    $\pi,\,\rho_0$ & Policy and initial-state distribution. \\
    $\boldsymbol{\lambda}\in\mathbb{R}_{\ge 0}^k$ & Lagrange multiplier vector. \\
    \midrule
    \multicolumn{2}{l}{\emph{Power system}} \\
    $\mathbf{P},\,\mathbf{Q}$ & Branch active and reactive power flows. \\
    $\mathbf{v}^2$ & Vector of squared bus voltage magnitudes (DistFlow uses squared form). \\
    $\mathbf{D}\in\mathbb{R}^{N_\ell\times N}$ & Downstream-aggregation matrix (radial PF). \\
    $\boldsymbol{\Pi}\in\mathbb{R}^{N\times N_\ell}$ & Path matrix from slack bus to each bus (radial PF). \\
    $\mathbf{H},\,\mathbf{H}_d,\,\mathbf{H}_\text{bus}$ & Generator-side, demand-side, and full bus-level PTDF (DC OPF). \\
    $A_\text{aug}$ & Augmented KKT matrix in ADMM DC OPF. \\
    $\mathbf{p},\,\underline{\mathbf{p}},\,\overline{\mathbf{p}}$ & Generator dispatch vector and its lower and upper output bounds (DC OPF). \\
    $\mathbf{z}_1,\,\mathbf{z}_2$ & ADMM auxiliary variables for the generator-bound and line-flow constraints. \\
    $\mathbf{y}_1,\,\mathbf{y}_2$ & ADMM dual variables for the generator-bound and line-flow constraints. \\
    $\mathbf{r}^{(k)}$ & ADMM residual vector at iteration $k$ (right-hand side of the augmented KKT solve). \\
    $\rho,\,\mu$ & ADMM penalty parameter and PDIPM barrier parameter. \\
    \midrule
    \multicolumn{2}{l}{\emph{JAX primitives used in PowerZooJax}} \\
    \texttt{jax.jit} & Compiles the full rollout (reset, step, reward, cost) into a single GPU program. \\
    \texttt{jax.vmap} & Runs the same environment in parallel across batched scenarios. \\
    \texttt{jax.lax.scan} & Unrolls the per-episode loop (48 or 288 steps) on the device. \\
    \texttt{jax.lax.while\_loop} & Drives the radial PF fixed-point iteration to convergence on the device. \\
    \texttt{jax.lax.fori\_loop} & Runs a fixed number of ADMM or PDIPM solver iterations on the device. \\
    PRNG key & Random seed passed to \texttt{reset}/\texttt{step} for reproducible scenario and policy sampling. \\
\end{longtable}
\endgroup

\FloatBarrier

\section{Physical Models}
\label{app:D}

The five tasks share five physical kernels (Table~\ref{tab:physical-cores-app}). Main paper Sections~5.1--5.2 give one equation each for radial PF and DC OPF; this appendix gives the full derivations and shows how each kernel becomes a fixed-shape JAX computation graph. Section~\ref{app:D.4} covers the device dynamics shared by DSO, DERs, and DCMG, and Section~\ref{app:D.5} covers the DCMG couplings.

\begin{table}[h]
  \caption{Physical kernels reused across tasks.}
  \label{tab:physical-cores-app}
  \centering
  \footnotesize
  \begin{tabularx}{\linewidth}{p{4.2cm}p{2.4cm}Y}
    \toprule
    Kernel & Used by & Output \\
    \midrule
    Radial AC PF (DistFlow recurrence) & DSO, DERs & Branch flows, bus voltages, network loss \\
    \midrule
    DC OPF (ADMM) & TSO & Generator dispatch, line flows, operating cost \\
    \midrule
    Bid-based market clearing (PDIPM) & GenCos & Cleared dispatch, locational marginal prices, profit \\
    \midrule
    Resource device dynamics & DSO, DERs, DCMG & SoC, PQ injections, flexible-load shifting, diesel output \\
    \midrule
    DCMG balance and thermal coupling & DCMG & IT power, cooling demand, grid import, SLA tracking, carbon emission \\
    \bottomrule
  \end{tabularx}
\end{table}

\subsection{Radial AC power flow as a JAX graph}
\label{app:D.1}

\textbf{Standard DistFlow recurrence.} Consider a radial feeder with $N$ buses and $N_\ell = N - 1$ branches, slack bus $0$, and per-branch resistance and reactance $r_\ell, x_\ell$. Each branch $\ell$ connects parent bus $p(\ell)$ to child bus $c(\ell)$. Let $p^{\mathrm{load}}_n, q^{\mathrm{load}}_n$ be the active and reactive load at bus $n$, and $P_\ell, Q_\ell$ be the active and reactive flow on branch $\ell$ (positive in the parent$\to$child direction). The DistFlow recurrence~\citep{NetworkReconfigurationDistribution1989baran} reads
\begin{align}
\label{eq:distflow-base}
  P_\ell &= p^{\mathrm{load}}_{c(\ell)} + \sum_{\ell'\in\text{children}(\ell)} P_{\ell'} + r_\ell\,\frac{P_\ell^2 + Q_\ell^2}{v^2_{p(\ell)}}, \\
  Q_\ell &= q^{\mathrm{load}}_{c(\ell)} + \sum_{\ell'\in\text{children}(\ell)} Q_{\ell'} + x_\ell\,\frac{P_\ell^2 + Q_\ell^2}{v^2_{p(\ell)}}, \\
  v^2_{c(\ell)} &= v^2_{p(\ell)} - 2\bigl(r_\ell P_\ell + x_\ell Q_\ell\bigr) + (r_\ell^2 + x_\ell^2)\,\frac{P_\ell^2 + Q_\ell^2}{v^2_{p(\ell)}},
\end{align}
where $\text{children}(\ell)$ denotes the set of branches whose parent bus is $c(\ell)$. Conventional pipelines walk the feeder tree backward (children $\to$ parent) to evaluate $P_\ell, Q_\ell$, then forward (root $\to$ leaves) to evaluate $v^2_n$, iterating until $\max_n |v^{2,(k)}_n - v^{2,(k-1)}_n| < \varepsilon$ (typically $\varepsilon = 10^{-6}$). On a host CPU each pass is a Python tree walk and each iteration triggers a host-device synchronization.

\textbf{Topology matrices.} For a radial network the backward and forward sweeps depend only on topology, so they can be precomputed once at environment construction time as two binary matrices:
\begin{equation}
\label{eq:topology-matrices}
\begin{aligned}
  \mathbf{D} &\in \{0,1\}^{N_\ell\times N},\quad
  D_{\ell, n}=
  \begin{cases}
    1 & \text{if bus } n \text{ is downstream of branch } \ell,\\
    0 & \text{else,}
  \end{cases}
  \\
  \boldsymbol{\Pi} &\in \{0,1\}^{N\times N_\ell},\quad
  \Pi_{n, \ell}=
  \begin{cases}
    1 & \text{if branch } \ell \text{ lies on the path from bus } 0 \text{ to bus } n,\\
    0 & \text{else.}
  \end{cases}
\end{aligned}
\end{equation}
The DistFlow iteration becomes
\begin{equation}
\label{eq:distflow-jax}
  \begin{aligned}
    \mathbf{P}^{(k)} &= \mathbf{D}\,\bigl(\mathbf{p}^{\mathrm{load}} + \mathbf{p}^{\mathrm{loss},(k-1)}\bigr),\\
    \mathbf{Q}^{(k)} &= \mathbf{D}\,\bigl(\mathbf{q}^{\mathrm{load}} + \mathbf{q}^{\mathrm{loss},(k-1)}\bigr),\\
    \boldsymbol{\delta}^{2,(k)} &= 2\,(\mathbf{r}\odot\mathbf{P}^{(k)} + \mathbf{x}\odot\mathbf{Q}^{(k)}) - (\mathbf{r}^2+\mathbf{x}^2)\odot\frac{(\mathbf{P}^{(k)})^2+(\mathbf{Q}^{(k)})^2}{\mathbf{v}^{2,(k-1)}_{p(\cdot)}},\\
    \mathbf{v}^{2,(k)} &= V_\text{slack}^2\,\mathbf{1} - \boldsymbol{\Pi}\,\boldsymbol{\delta}^{2,(k)},
  \end{aligned}
\end{equation}
with the per-bus loss-correction $\mathbf{p}^{\mathrm{loss}}, \mathbf{q}^{\mathrm{loss}}$ recomputed from the previous iterate. Equation~\eqref{eq:distflow-jax} expresses one PF iteration as fixed-shape \texttt{jax.numpy} matrix-vector products with no host-side branching.

\textbf{Compiled fixed-point loop and batching.} The whole iteration body $f:(\mathbf{P},\mathbf{Q},\mathbf{v}^2)\to(\mathbf{P}',\mathbf{Q}',\mathbf{v}'^2)$ is implemented with \texttt{jnp} and \texttt{lax} ops only, with no host-side mutation. PowerZooJax wraps it in \texttt{jax.lax.while\_loop} with the convergence predicate evaluated on traced arrays, then composes the per-step PF solve with reward and cost computation through \texttt{jax.lax.scan} for time and \texttt{jax.vmap} for batched scenarios. The 48 PF evaluations of one DSO or DERs episode are therefore one accelerator-resident computation, not 48 host-side solver calls. The same kernel powers DSO and DERs.

\subsection{DC optimal power flow with ADMM}
\label{app:D.2}

\textbf{LP formulation.} TSO uses a DC OPF with the bus-level PTDF $\mathbf{H}_\text{bus}\in\mathbb{R}^{N_\ell\times N}$ from the prebuilt case file; $\mathbf{H}\in\mathbb{R}^{N_\ell\times N_g}$ and $\mathbf{H}_d\in\mathbb{R}^{N_\ell\times N_d}$ select the generator and demand bus columns of $\mathbf{H}_\text{bus}$. Let $\mathbf{p}\in\mathbb{R}^{N_g}$ be the generator dispatch vector, $\mathbf{c}^g$ marginal cost, $\mathbf{p}^d\in\mathbb{R}^{N_d}$ nodal demand, $\overline{\mathbf{f}}$ line-flow limits, and $(\underline{\mathbf{p}},\overline{\mathbf{p}})$ generator output bounds. The DC OPF is the LP
\begin{equation}
\label{eq:dcopf}
  \min_{\mathbf{p}}\; \mathbf{c}^{g\,\top}\mathbf{p} \quad\text{s.t.}\quad
  \mathbf{1}^\top\mathbf{p} = \mathbf{1}^\top\mathbf{p}^d,\ \ |\mathbf{H}\mathbf{p} - \mathbf{H}_d\mathbf{p}^d| \le \overline{\mathbf{f}},\ \ \underline{\mathbf{p}} \le \mathbf{p}\le\overline{\mathbf{p}}.
\end{equation}
This DC OPF is the inner solver invoked at every step of the TSO task; commitment, ramp, and minimum up/down-time constraints are enforced by the environment (Section~\ref{app:E.2}).

\textbf{ADMM iteration.} We split the box and line-flow constraints into auxiliary variables $\mathbf{z}_1,\mathbf{z}_2$ with dual variables $\mathbf{y}_1,\mathbf{y}_2$ and penalty $\rho > 0$. The system-wide power balance is kept as a hard equality and handled by an augmented KKT system. After standard derivation \citep{DistributedOptimalPower2014erseghe} the ADMM update for iteration $k$ reads
\begin{equation}
\label{eq:admm}
  \begin{aligned}
    \begin{bmatrix} \mathbf{p}^{(k+1)} \\ \nu^{(k+1)}_\text{eq} \end{bmatrix}
      &= A_\text{aug}^{-1} \begin{bmatrix} \mathbf{r}^{(k)} \\ \mathbf{1}^\top\mathbf{p}^d \end{bmatrix},\\
    \mathbf{z}_1^{(k+1)} &= \mathrm{clip}\bigl(\mathbf{p}^{(k+1)} + \mathbf{y}_1^{(k)}/\rho,\ \underline{\mathbf{p}},\ \overline{\mathbf{p}}\bigr),\\
    \mathbf{z}_2^{(k+1)} &= \mathrm{clip}\bigl(\mathbf{H}\mathbf{p}^{(k+1)} - \mathbf{H}_d\mathbf{p}^d + \mathbf{y}_2^{(k)}/\rho,\ -\overline{\mathbf{f}},\ \overline{\mathbf{f}}\bigr),\\
    \mathbf{y}_1^{(k+1)} &= \mathbf{y}_1^{(k)} + \rho\bigl(\mathbf{p}^{(k+1)} - \mathbf{z}_1^{(k+1)}\bigr),\\
    \mathbf{y}_2^{(k+1)} &= \mathbf{y}_2^{(k)} + \rho\bigl(\mathbf{H}\mathbf{p}^{(k+1)} - \mathbf{H}_d\mathbf{p}^d - \mathbf{z}_2^{(k+1)}\bigr),
  \end{aligned}
\end{equation}
with the residual $\mathbf{r}^{(k)} = -\mathbf{c}^g + \rho\,\mathbf{z}_1^{(k)} - \mathbf{y}_1^{(k)} + \rho\mathbf{H}^\top(\mathbf{z}_2^{(k)} + \mathbf{H}_d\mathbf{p}^d) - \mathbf{H}^\top\mathbf{y}_2^{(k)}$, the multiplier $\nu_\text{eq}$ on $\mathbf{1}^\top\mathbf{p}=\mathbf{1}^\top\mathbf{p}^d$ recovered alongside $\mathbf{p}^{(k+1)}$ and discarded after extraction, and the augmented KKT matrix
\begin{equation}
\label{eq:admm-Aaug}
  A_\text{aug} = \begin{bmatrix} \rho\,(\mathbf{I} + \mathbf{H}^\top\mathbf{H}) & \mathbf{1} \\ \mathbf{1}^\top & 0 \end{bmatrix}
\end{equation}
depending only on the case parameters; PowerZooJax precomputes $A_\text{aug}^{-1}$ at construction time, so each ADMM iteration becomes one matrix-vector product plus elementwise clips. The whole solve runs as a fixed-length \texttt{jax.lax.fori\_loop} (TSO uses 200 iterations and tolerance $10^{-4}$), which compiles to a single XLA program. TSO uses this kernel.

\subsection{Bid-based market clearing with PDIPM}
\label{app:D.3}

GenCos clears a security-constrained economic dispatch (SCED) over piecewise-linear bid curves. Each agent $i$ submits up to $S=3$ break-points; the clearing problem with bid prices $c^b_{i,s}$ and segment caps $\overline{p}_{i,s}$ is
\begin{equation}
\label{eq:pdipm-lp}
  \min_{p_{i,s}\ge 0}\; \sum_{i,s} c^b_{i,s}\,p_{i,s}\quad\text{s.t.}\quad
  \sum_{i,s} p_{i,s} = D_t,\ \ |\mathbf{H}\mathbf{p} - \mathbf{H}_d\mathbf{p}^d|\le\overline{\mathbf{f}},\ \ p_{i,s}\le\overline{p}_{i,s}.
\end{equation}
PowerZooJax solves it with a primal-dual interior-point method~\citep{ComputationalIssuesMarketBased2007wang}. With barrier parameter $\mu$, the perturbed KKT system is
\begin{equation}
\label{eq:pdipm-kkt}
  \nabla_p\mathcal{L} = 0,\quad
  s_j\,\xi_j = \mu,\ \forall j,\quad
  s_j \ge 0,\ \xi_j \ge 0,
\end{equation}
where $s_j,\xi_j$ are slacks and multipliers for the inequality constraints. At each iteration the duality gap $\mu^{(k)} = \mathbf{s}^{(k)\top}\boldsymbol{\xi}^{(k)}/m$ is recomputed and shrunk via a centring term: the perturbed complementarity residual uses target $\sigma\,\mu^{(k)}$ with $\sigma=0.2$. Each Newton step solves a reduced KKT system. The loop runs as a fixed-length \texttt{jax.lax.fori\_loop} with early stopping when $\mu^{(k)} < 10^{-5}$. The dual variable $\nu^*$ on the power-balance equality is recovered analytically from the primal solution; locational marginal prices are
\begin{equation}
\label{eq:pdipm-lmp}
  \boldsymbol{\pi}^* = -\nu^*\,\mathbf{1} - \mathbf{H}_\text{bus}^\top\boldsymbol{\xi}^*_\text{flow},
\end{equation}
where $\boldsymbol{\xi}^*_\text{flow} = \boldsymbol{\xi}^*_\text{upper} - \boldsymbol{\xi}^*_\text{lower}$ is the net line-constraint dual built from the duals of the upper and lower line-flow inequalities in \eqref{eq:pdipm-lp}.

\subsection{Resource device dynamics}
\label{app:D.4}

\paragraph{Battery storage.} For battery $j$ with capacity $E^\text{max}_j$ (MWh) and rated power $P^\text{max}_j$ (MW), the per-step state-of-charge update with charge/discharge efficiencies $\eta^c_j, \eta^d_j$ and step length $\Delta t$ is
\begin{equation}
  \label{eq:soc-clip}
  \mathrm{SoC}_j^{t+1} = \mathrm{clip}\Bigl(\mathrm{SoC}_j^t + \frac{\Delta t}{E^\text{max}_j}\bigl(\eta^c_j p^c_{j,t} - \tfrac{1}{\eta^d_j}\,p^d_{j,t}\bigr),\ \mathrm{SoC}^\text{min}_j,\ \mathrm{SoC}^\text{max}_j\Bigr),
\end{equation}
with $p^c_{j,t}, p^d_{j,t}\in [0, P^\text{max}_j]$ and $p^c_{j,t}\,p^d_{j,t} = 0$ enforced by parameterizing the action as $p_j\in[-P^\text{max}_j, P^\text{max}_j]$ and splitting at zero. Tasks that include batteries (DERs, DCMG) terminate the episode by adding a quadratic penalty $(\mathrm{SoC}_j^T - \mathrm{SoC}^\text{target}_j)^2$.

\paragraph{PV inverter.} A PV inverter $j$ with rated apparent power $S^\text{max}_j$, available DC power $p^\text{avail}_{j,t}$, and active/reactive setpoints $p_{j,t},q_{j,t}$ enforces the box-projected PQ envelope
\begin{equation}
  \label{eq:pq-envelope}
  0 \le p_{j,t} \le p^\text{avail}_{j,t},\quad p_{j,t}^2 + q_{j,t}^2 \le (S^\text{max}_j)^2,
\end{equation}
implemented as a closed-form projection onto the inverter capability circle.

\paragraph{Flexible load.} Flexible load $j$ has nominal demand $d^\text{nom}_{j,t}$, curtailment cap $\eta^\text{cur}_j\in[0,1]$, shifting cap $\eta^\text{shift}_j$, and a shift horizon $H_j$ (steps). Curtailed energy is permanently lost; shifted energy must be repaid within $H_j$ steps. The realized demand and the per-load energy buffer evolve as
\begin{align}
  \label{eq:load-energy}
  d^\text{real}_{j,t} &= (1 - u^\text{cur}_{j,t})\,d^\text{nom}_{j,t} - u^\text{shift}_{j,t}\,d^\text{nom}_{j,t} + u^\text{recall}_{j,t},\\
  E^\text{shift}_{j,t+1} &= E^\text{shift}_{j,t} + u^\text{shift}_{j,t}\,d^\text{nom}_{j,t} - u^\text{recall}_{j,t},
\end{align}
with $u^\text{shift}_{j,t}\le \eta^\text{shift}_j$ and $u^\text{cur}_{j,t}\le \eta^\text{cur}_j$. DSO and DERs use this model.

\paragraph{Diesel generator.} Diesel generator $j$ has rated power $P^\text{max}_j$ and a linear fuel cost $C_j(p) = c^\text{fuel}_j\,p$ proportional to the delivered active power. The continuous normalized command $a^\text{dg}_{j,t}\in[0,1]$ maps to
\begin{equation}
  \label{eq:norm-cmd}
  p_{j,t} = \mathrm{clip}(a^\text{dg}_{j,t}\,P^\text{max}_j,\ 0,\ P^\text{max}_j),\qquad \text{cost}_{j,t} = C_j(p_{j,t})\,\Delta t,
\end{equation}
with an optional minimum-loading threshold ($a^\text{dg}_{j,t}<P^\text{min}_j/(2 P^\text{max}_j)$ snaps to $0$, in between snaps up to $P^\text{min}_j$) so the operating point respects diesel wet-stacking guidance. DCMG wraps the diesel as a same-step residual-slack actuator, so dispatch is realized after the workload, cooling, and battery decisions.

\subsection{DCMG coupling}
\label{app:D.5}

The DCMG task couples the device dynamics of Section~\ref{app:D.4} with a lumped-capacitance thermal model and a single-bus power balance. Let $W_t$ be the GPU-equivalent compute load, $\theta_t^\text{IT}$ the IT-zone temperature, $\theta_t^\text{out}$ the outdoor temperature, and $p^\text{cool}_t$ the cooling electrical power. The IT-zone thermal dynamics are
\begin{equation}
\label{eq:dcmg-thermal}
  \theta_{t+1}^\text{IT} = \theta_t^\text{IT} + \frac{\Delta t}{C_\text{th}}\bigl(\alpha_W\,W_t - \beta_t\,p^\text{cool}_t + \gamma_\text{out}(\theta_t^\text{out} - \theta_t^\text{IT})\bigr),
\end{equation}
with thermal capacity $C_\text{th}$, IT-to-heat coefficient $\alpha_W$ (so $\alpha_W W_t$ is the IT electrical power at step $t$, equal to its heat output), wall-to-ambient coefficient $\gamma_\text{out}$, and cooling COP $\beta_t = \beta(\theta_t^\text{out})$ that decreases with outdoor temperature. The over-temperature cost channel $c^\text{th}_t = (\theta_t^\text{IT} - \theta^\text{max})_+$ is added to the cost vector at every step. Cooling is implemented as a setpoint-tracking controller: the RL action $a^\text{cool}_t$ sets the IT-zone setpoint, and $p^\text{cool}_t$ equals the resulting heat-removal rate divided by $\beta_t$ (Section~\ref{app:E.5}).

The local power balance at the point of common coupling is
\begin{equation}
\label{eq:dcmg-balance}
  p^\text{DC}_t \;=\; p^\text{PV}_t + p^\text{batt}_t + p^\text{DG}_t + p^\text{grid}_t,
\end{equation}
where $p^\text{DC}_t = p^\text{IT}_t + p^\text{cool}_t + p^\text{aux}_t$ is the total data-center load and $p^\text{aux}_t$ is the auxiliary load (PSU losses and other site loads, modeled as a fixed fraction of $p^\text{IT}_t$). Grid import $p^\text{grid}_t\in[0, \overline{p}^\text{grid}]$ in grid-connected mode and $p^\text{grid}_t=0$ in islanded mode. SLA tracking compares served workload against committed jobs; the SLA-deficit cost channel is the per-step expired-task count normalized by the GPU count. The reward combines energy use, electricity cost (GB MID market price), and carbon emission (grid-import and DG carbon intensities); the weights are given in Section~\ref{app:E.5}.

\FloatBarrier

\section{Per-Task Cards}
\label{app:E}

This section gives the formal CMDP instantiation, observation, action, reward, cost vector, splits, baselines, and safety budgets for each of the five tasks. Concrete numerical thresholds, time-step lengths, and split sizes are taken from the per-task configuration files (Section~\ref{app:G.3} shows one such file). Table~\ref{tab:task-overview} gives a one-line summary of the suite.

\paragraph{Notation in the task cards.} We use bold lowercase for vectors and bold uppercase for matrices; symbols without a $t$ subscript are time-invariant. Per-task dimensions follow each task's case file and configuration; the explicit dimension counts under each observation/action equation in this section are evaluated at the canonical case (\texttt{case5}, \texttt{case118}, \texttt{case33bw}, \texttt{case141}, and the data center microgrid configuration shipped with the benchmark).

\begin{table}[h]
  \caption{One-line task overview.}
  \label{tab:task-overview}
  \centering
  \footnotesize
  \begin{tabularx}{\linewidth}{p{1.0cm}Yp{2.8cm}p{3.4cm}}
    \toprule
    Task & Decision problem & Primary metric & Cost channels \\
    \midrule
    GenCos      & 5-agent strategic bidding on 5-bus market, 48 half-hour steps & Total per-agent profit & Thermal overload (diagnostic) \\
    \midrule
    TSO         & 1-agent SCUC on 118-bus system, 48 half-hour steps & Operating cost & Reserve, thermal (Form 2) \\
    \midrule
    DSO         & 1-agent demand response on 33-bus feeder, 48 half-hour steps & Network loss (MWh) & Voltage band \\
    \midrule
    DERs        & 12-agent cooperative DER control on 141-bus feeder, 48 half-hour steps & Active power loss (MW) & Voltage band, thermal, resource \\
    \midrule
    DCMG        & 1-agent data center microgrid scheduling, 288 five-minute steps & Episode return & SLA, over-temperature, power balance \\
    \bottomrule
  \end{tabularx}
\end{table}

\subsection{Generation-company bidding task (GenCos)}
\label{app:E.1}
\begin{figure}[h]
  \centering
  \includegraphics[width=0.55\linewidth]{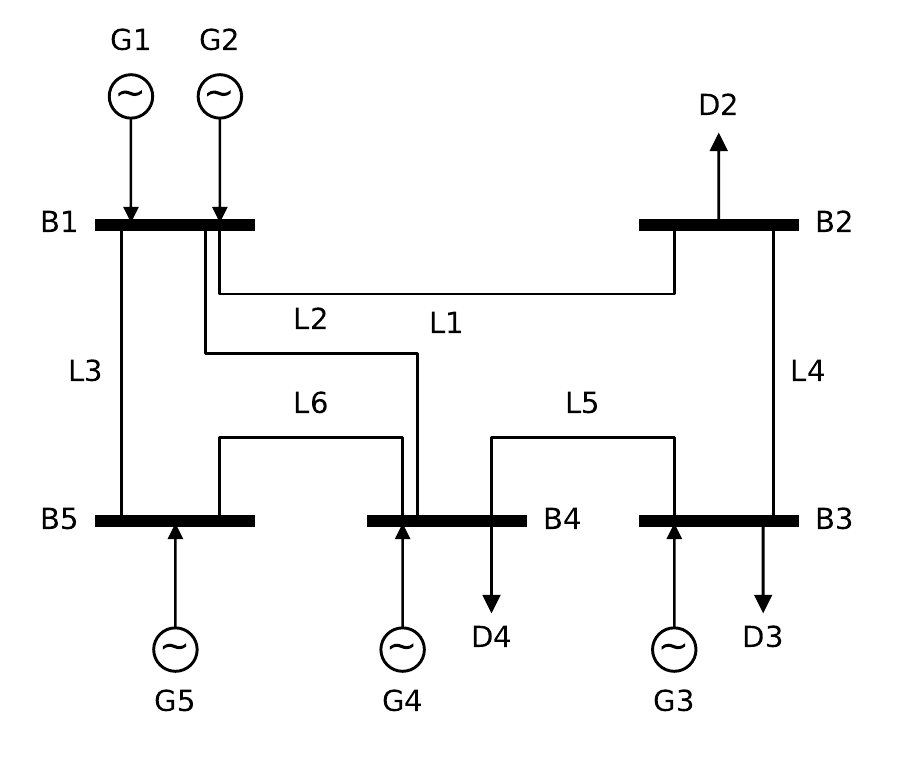}
  \caption{IEEE 5-bus market test system (\texttt{case5}) used by the GenCos task: 5 buses B1--B5 (horizontal bars), 6 transmission lines L1--L6, 5 thermal generators G1--G5 (circled symbols, one per agent; G1 and G2 share bus B1, while G3, G4, G5 sit at buses B3, B4, B5 respectively), and 3 load sites D2, D3, D4 (outward arrows at buses B2, B3, B4). Each step, the SCED clearing of Section~\ref{app:D.3} returns the per-bus LMPs and per-unit dispatches that drive each agent's profit.}
  \label{fig:gencos-case5}
\end{figure}

\textbf{Task.}
Five generation companies (GenCos) compete in a repeated wholesale electricity market on the IEEE 5-bus transmission grid (\texttt{case5}, GB demand profiles\footnote{Great Britain demand traces from the National Energy System Operator (NESO), \url{https://www.neso.energy/data-portal/historic-demand-data}.}; see Figure~\ref{fig:gencos-case5}). The horizon is 24 hours at 30-minute resolution, so an episode contains 48 sequential market clearing steps. At each step the market operator clears all bids via a security-constrained economic dispatch (SCED, an optimization that picks the cheapest feasible bid mix subject to grid limits), returning each unit's dispatch level $P^g_{i,t}$ and the locational marginal prices (LMPs, the per-bus electricity prices). Successive clearings are coupled by runtime ramp bounds (per-step limits on how fast a generator can change output) enforced inside the SCED linear program: at step $t$ each unit's box constraint is tightened to $p^{\min}_{i,t}=\max(p^{\min}_i, P^g_{i,t-1}-\Delta P^{\mathrm{down}}_i)$ and $p^{\max}_{i,t}=\min(p^{\max}_i, P^g_{i,t-1}+\Delta P^{\mathrm{up}}_i)$, so the episode is a rolling sequential market rather than 48 independent auctions. The market and the grid are shared across the five agents, which makes this a competitive multi-agent task.

\textbf{Observation and Action.}
Each agent owns one generator and observes only its own physical status plus a few public market signals. For agent $i$,
\begin{equation}
\label{eq:gencos-state-action}
\mathbf{o}_{i,t}=\bigl(c^{b}_{i},\ \widetilde{P}^{\max}_{i},\ P^{g}_{i,t-1},\ \widetilde{w}_{i,t-1},\ h^{\mathrm{ramp}}_{i,t},\ D^{\mathrm{fcst}}_{t+1},\ \overline{\boldsymbol{\pi}}^{\mathrm{hist}}_{t},\ \tau_t\bigr) \in \mathbb{R}^{12}.
\end{equation}
\begin{equation}
\mathbf{a}_{i,t}\in[-1,1]^{K_{\mathrm{seg}}}.
\end{equation}
Each unit's true cost is $\mathrm{TC}_i(P)=\tfrac{a_i}{3}P^3+\tfrac{b_i}{2}P^2+c_iP$, the integral of a quadratic marginal cost (the cost of producing one extra MW at output $P$) $\mathrm{MC}_i(P)=a_iP^2+b_iP+c_i$; truthful prices are obtained by evaluating $\mathrm{MC}_i$ at the midpoint of each of $K_{\mathrm{seg}}{=}3$ equal-width segments covering $[p^{\min}_i, p^{\max}_i]$. The eight observation entries are:
\begin{itemize}\setlength{\itemsep}{1pt}
  \item $c^{b}_i$: unit's truthful first-segment price (lowest-block marginal cost), normalized by the configured price scale;
  \item $\widetilde{P}^{\max}_i$: capacity relative to the largest unit;
  \item $P^{g}_{i,t-1}$: previous dispatch, normalized by the agent's capacity;
  \item $\widetilde{w}_{i,t-1}$: previous-step profit, normalized by a fixed maximum-revenue scale;
  \item $h^{\mathrm{ramp}}_{i,t}=\min(p^{\max}_i-P^g_{i,t-1},\Delta P^{\mathrm{up}}_i)/p^{\max}_i$: achievable ramp-up next step (the smaller of remaining capacity and the per-step ramp-up limit);
  \item $D^{\mathrm{fcst}}_{t+1}$: next-step total-demand forecast (perfect one-step look-ahead, normalized by total capacity);
  \item $\overline{\boldsymbol{\pi}}^{\mathrm{hist}}_{t}\in\mathbb{R}^{4}$: circular buffer of the four most recent system-mean LMPs (oldest to newest), giving every agent a coarse public price-dynamics signal;
  \item $\tau_t \in \mathbb{R}^{2}$: sin/cos time-of-day encoding.
\end{itemize}
The six task-level scalars, the four-element price history, and the two-component time encoding sum to 12.

The raw action $\mathbf{a}_{i,t}$ is mapped to a piecewise bid curve in three steps:
\begin{enumerate}\setlength{\itemsep}{1pt}
  \item Shift to non-negative magnitudes $m_{i,t}=(\mathbf{a}_{i,t}+1)/2\in[0,1]^{K_{\mathrm{seg}}}$.
  \item Sort along the segment axis to enforce a monotone offer (no later block cheaper than an earlier block).
  \item Apply a multiplicative markup capped by a configured maximum, so that the cleared offer for segment $k$ is $\mathrm{offer}_{i,k,t}=c^{\mathrm{seg}}_{i,k}\cdot\bigl(1+\mathrm{sort}(m_{i,t})_k\cdot\overline{m}\bigr)$ with $c^{\mathrm{seg}}_{i,k}$ the truthful segment price.
\end{enumerate}

\textbf{Reward and Violation Cost.}
Each agent's reward is its realized profit at the cleared dispatch, equal to cleared revenue $\boldsymbol{\pi}^*_{b(i),t}\,P^{g}_{i,t}\,\Delta t$ minus true production cost $\mathrm{TC}_i(P^{g}_{i,t})\,\Delta t$, where $b(i)$ is the bus (the network node where loads and generators connect) of unit $i$:
\begin{equation}
\label{eq:gencos-reward-cost}
r_{i,t}=\bigl(\boldsymbol{\pi}^*_{b(i),t}\,P^{g}_{i,t}-\mathrm{TC}_i(P^{g}_{i,t})\bigr)\,\Delta t,
\qquad
c_t=C^{\mathrm{thermal}}_t,
\end{equation}
with $\boldsymbol{\pi}^*$ the cleared LMP vector from Section~\ref{app:D.3}. The single violation channel $C^{\mathrm{thermal}}_t$ is the sum of post-clearing line-flow violations (in MW), shared across agents because line overloads are a property of the joint outcome rather than of any single bid. This channel is reported as a market-side diagnostic only.

\textbf{Baselines.}
The benchmark releases three non-learning strategic baselines (Truthful, Uniform-mid, and Max-markup) together with a learned IPPO baseline that uses one shared actor-critic across the five GenCos. The three heuristics bracket the strategic spectrum: Truthful (bidding at marginal cost) is the competitive-equilibrium reference (the outcome where every bidder offers at marginal cost), Max-markup (bidding at the configured cap) is the upper envelope of market power, and Uniform-mid (bidding at the midpoint markup) sits between them as a neutral oligopoly reference.

\textbf{Challenge.}
GenCos is a non-stationary, partially observable, competitive multi-agent task. Each agent only sees its own dispatch, profit, and a small public LMP window, and the market reward depends on the joint action through the SCED clearing. Policy improvement for one bidder shifts the optimization problem faced by every other bidder. These factors together define the standard hard regime for current MARL.

\subsection{Transmission dispatch task (TSO)}
\label{app:E.2}
\begin{figure}[h]
  \centering
  \includegraphics[width=1.00\linewidth]{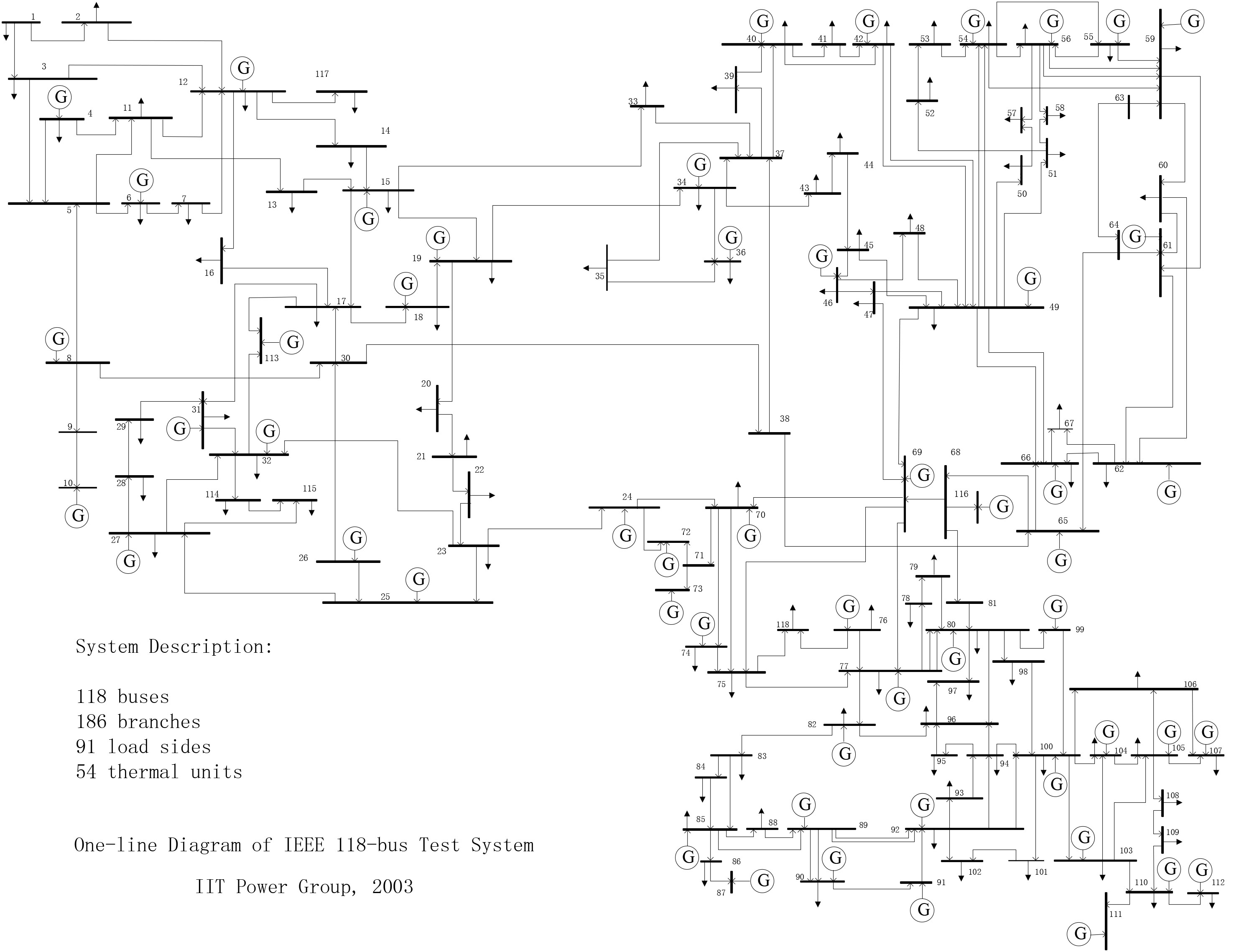}
  \caption{One-line diagram of the IEEE 118-bus transmission test system (\texttt{case118}) used by the TSO task: 118 buses (horizontal bars), 186 branches (lines connecting buses), 54 thermal generators (circled G; arrows pointing into their bus), and 91 load sides (arrows pointing out of their bus). The TSO task in Section~\ref{app:E.2} commits and dispatches the 54 generators on a 30-minute cadence under DC-OPF redispatch. Source: IIT Power Group, 2003.}
  \label{fig:tso-case118}
\end{figure}

\textbf{Task.}
A single transmission system operator (TSO) runs the 54-generator IEEE 118-bus transmission grid (\texttt{case118}, 186 lines; see Figure~\ref{fig:tso-case118}) over a 24-hour horizon at 30-minute resolution, which gives 48 sequential decision steps per episode and is driven by GB net-load profiles (gross demand minus must-take wind and solar)\footnote{Demand from the National Energy System Operator (NESO), \url{https://www.neso.energy/data-portal/historic-demand-data}; wind and solar from Elexon BMRS, \url{https://bmrs.elexon.co.uk/generation-by-fuel-type}.}. At each step the agent decides which generators are on (\emph{commitment}) and how much each produces (\emph{dispatch}); the environment then enforces ramp limits (per-step bounds on how fast a unit can change output), minimum up/down times (a unit must remain in its current state for at least this many consecutive steps after a transition), and a DC optimal power flow solution (a linear approximation that maps generator output to per-line flows). One centralized agent sees the whole grid, so this is a fully observable single-agent task with a large mixed continuous/binary action space.

\textbf{Observation and Action.}
The observation summarizes the whole grid; the action specifies a commitment intent and a dispatch preference per generator:
\begin{equation}
\label{eq:tso-state-action}
\begin{aligned}
\mathbf{o}_{t} = \bigl(&\mathbf{u}_{t},\ \mathbf{t}^{\mathrm{state}}_{t},\ \mathbf{P}^{g}_{t-1},\ \mathbf{C}^{g},\ \mathbf{f}_{t-1}, 
&D_{t},\ \eta^{\mathrm{rsv}}_{t},\ \tau_{t},\ \mathbf{D}^{\mathrm{fcst}}_{t}\bigr) \in \mathbb{R}^{410}.
\end{aligned}
\end{equation}
\begin{equation}
\mathbf{a}_{t}=\bigl(\mathbf{u}^{\mathrm{cmd}}_{t},\ \mathbf{P}^{\mathrm{pref}}_{t}\bigr) \in [-1,1]^{108}.
\end{equation}
The nine observation entries, with $N_g{=}54$ and $N_\ell{=}186$, are:
\begin{itemize}\setlength{\itemsep}{1pt}
  \item $\mathbf{u}_{t}\in\{0,1\}^{N_g}$: on/off status of all generators, carried over from the previous step;
  \item $\mathbf{t}^{\mathrm{state}}_{t}\in\mathbb{R}^{N_g}$: elapsed time each generator has been in its current on/off state (normalized), letting a learner respect minimum up/down constraints by inspection;
  \item $\mathbf{P}^{g}_{t-1}\in\mathbb{R}^{N_g}$: previous dispatch, normalized by per-unit capacity;
  \item $\mathbf{C}^{g}\in\mathbb{R}^{N_g}$: linear coefficient $b_i$ of each unit's quadratic marginal-cost function $\mathrm{MC}_i(P)=a_iP^2+b_iP+c_i$, normalized by its maximum across units;
  \item $\mathbf{f}_{t-1}\in\mathbb{R}^{N_\ell}$: active power flow on each transmission line, normalized by line capacity;
  \item $D_{t}\in\mathbb{R}$: aggregate system demand, normalized by total capacity;
  \item $\eta^{\mathrm{rsv}}_{t}\in\mathbb{R}$: committed-headroom reserve ratio (committed capacity minus current load, divided by load);
  \item $\tau_t\in\mathbb{R}^{2}$: sin/cos time-of-day encoding;
  \item $\mathbf{D}^{\mathrm{fcst}}_{t}\in\mathbb{R}^{4}$: four-step (two-hour) perfect look-ahead total-load forecast.
\end{itemize}
The dimension count is $4\,N_g + N_\ell + 4 + 4 = 410$.

The action is two-headed:
\begin{itemize}\setlength{\itemsep}{1pt}
  \item $\mathbf{u}^{\mathrm{cmd}}_{t}\in[-1,1]^{N_g}$: commitment intent applied by sign threshold (entries $u^{\mathrm{cmd}}_{i,t}>0$ request unit $i$ on for the next step), with the active minimum up/down constraints overriding any request that would violate them;
  \item $\mathbf{P}^{\mathrm{pref}}_{t}\in[-1,1]^{N_g}$: dispatch preference, denormalized linearly to each unit's ramp-adjusted feasible interval $[p^{\min}_{i,t}, p^{\max}_{i,t}]$ and added as a linear bias to its marginal-cost intercept inside the DC OPF, so the cleared dispatch follows the preference whenever it is feasible under all operational limits.
\end{itemize}
The action dimension is $2 N_g = 108$.

\textbf{Reward and Violation Cost.}
The reward is the negative total operating cost, summing fuel, startup, and no-load components, rescaled by a fixed factor (Section~\ref{app:H.1}). The two-channel violation cost separately reports thermal line overloads $C^{\mathrm{thermal}}_t$ (the cumulative MW above each line's capacity) and reserve-margin shortfalls $C^{\mathrm{reserve}}_t$ (the MW deficit of committed headroom relative to the configured 5\% margin), so that a safe-RL learner can trade them off explicitly rather than collapse them into a single penalty. Both channels are reported under Form~2 (zero-violation constraint):
\begin{equation}
\label{eq:tso-reward-cost}
r_{t}=-C^{\mathrm{operating}}_{t},
\qquad
\mathbf{c}_{t}=\bigl(C^{\mathrm{thermal}}_{t},\ C^{\mathrm{reserve}}_{t}\bigr).
\end{equation}

\textbf{Baselines.}
The benchmark provides two heuristic baselines (All-On, which leaves every unit committed, and Merit Order, which commits the cheapest unit first until total capacity covers demand plus a fixed 5\% reserve margin) and two learned baselines: PPO as the unconstrained reference and PPO-Lagrangian with a primal-dual update on the two cost channels.

\textbf{Challenge.}
TSO mixes 54 binary commitments with 54 continuous setpoints, so the action space is high-dimensional and combinatorial. Each step also requires the policy to keep an OPF feasible against ramp, minimum up/down, thermal, and reserve constraints simultaneously. Reserve violations are sparse and bursty, which hurts safe-RL credit assignment.

\subsection{Distribution demand response task (DSO)}
\label{app:E.3}
\begin{figure}[h]
  \centering
  \includegraphics[width=0.95\linewidth]{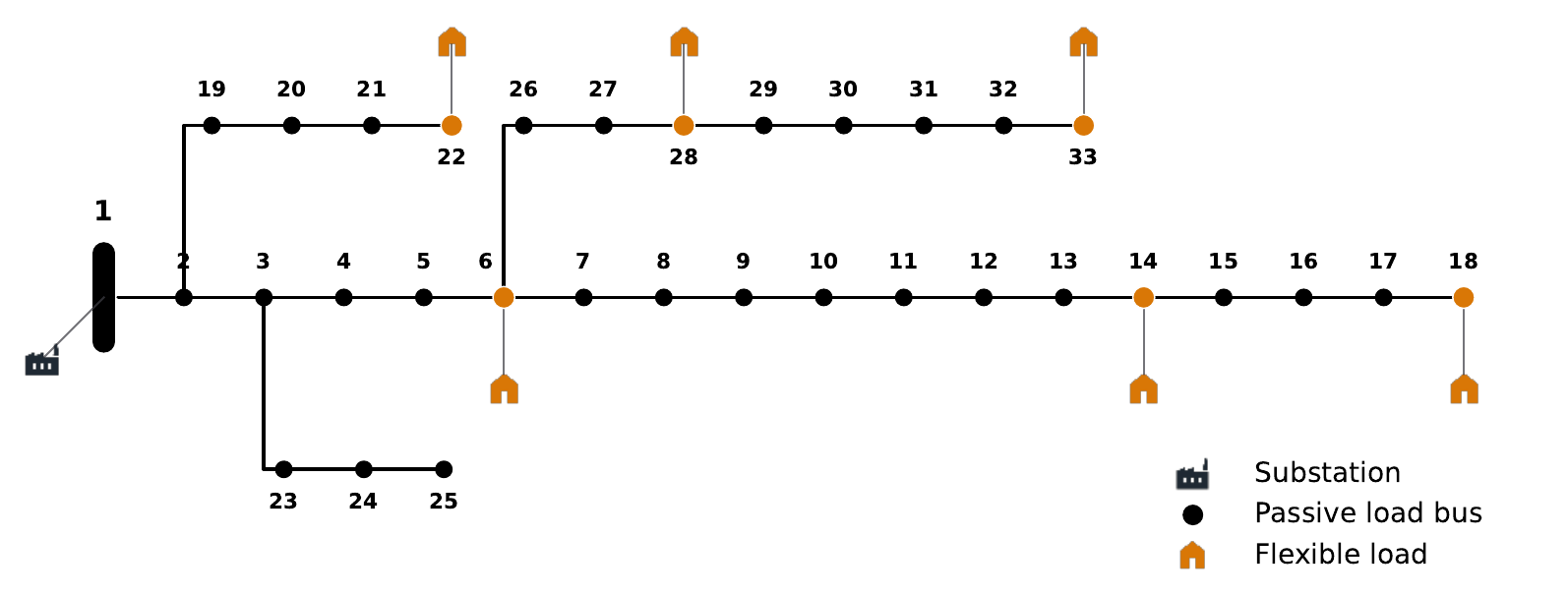}
  \caption{IEEE 33-bus radial distribution network (\texttt{case33bw}) used by the DSO task. The substation (factory icon) at bus 1 supplies the network through a main feeder (buses 2--18) and three lateral branches (19--22, 23--25, 26--33). Gray house icons mark passive loads; orange house icons mark the six controllable flexible loads at buses 6, 14, 18, 22, 28, and 33. The DSO task in Section~\ref{app:E.3} sets curtailment and load-shift fractions for these six devices on a 30-minute cadence under radial AC power flow.}
  \label{fig:dso-case33}
\end{figure}

\textbf{Task.}
A single distribution system operator (DSO) coordinates six flexible loads on the IEEE 33-bus distribution network (\texttt{case33bw}; see Figure~\ref{fig:dso-case33}) over a 24-hour episode at 30-minute resolution, which gives 48 sequential decision steps per episode and is driven by real Ausgrid demand profiles\footnote{Ausgrid distribution demand traces, \url{https://www.ausgrid.com.au/Industry/Our-Research/Data-to-share/}.}. The goal is to reduce total active power loss while keeping every bus voltage inside the safe band $[0.94,1.06]$~p.u.\ (per unit, the normalized voltage scale used in power systems). One centralized agent sees the whole network, so this is a fully observable single-agent task with smooth continuous control.

\textbf{Observation and Action.}
\begin{equation}
\label{eq:dso-state-action}
\mathbf{o}_{t}=\bigl(\mathbf{V}_{t},\ \mathbf{P}^{\mathrm{br}}_{t},\ \mathbf{Q}^{\mathrm{br}}_{t},\ \mathbf{P}^{\mathrm{load}}_{t},\ \mathbf{Q}^{\mathrm{load}}_{t},\ \tau_{t},\ \mathbf{z}^{\mathrm{flex}}_{t}\bigr) \in \mathbb{R}^{195}.
\end{equation}
\begin{equation}
\mathbf{a}_{t}=\bigl(\mathbf{d}^{\mathrm{curtail}}_{t},\ \mathbf{d}^{\mathrm{shift}}_{t}\bigr) \in [-1,1]^{12}.
\end{equation}
The seven observation entries, with $N{=}33$ buses and $N_\ell{=}32$ branches, are:
\begin{itemize}\setlength{\itemsep}{1pt}
  \item $\mathbf{V}_{t}\in\mathbb{R}^{N}$: per-bus voltage magnitude, centered and scaled as $(|V|-1)/0.1$ so nominal voltages sit near 0 and the band edges $0.94/1.06$~p.u.\ sit near $\pm 0.6$;
  \item $\mathbf{P}^{\mathrm{br}}_{t},\mathbf{Q}^{\mathrm{br}}_{t}\in\mathbb{R}^{N_\ell}$: active and reactive (the imaginary part of AC power, in MVAr) power flows on each branch, normalized by the reference load;
  \item $\mathbf{P}^{\mathrm{load}}_{t},\mathbf{Q}^{\mathrm{load}}_{t}\in\mathbb{R}^{N}$: active and reactive nodal demands, normalized by the reference load;
  \item $\tau_{t}\in\mathbb{R}^{2}$: sin/cos time-of-day encoding;
  \item $\mathbf{z}^{\mathrm{flex}}_{t}\in\mathbb{R}^{30}$: stacks five normalized variables per controllable load --- $[c^{\mathrm{cur}}, s^{\mathrm{out}}, s^{\mathrm{in}}, b^{\mathrm{fill}}, b^{\mathrm{eng}}]$, namely current curtailment level, outgoing/incoming shift fractions, deferred-buffer fill ratio, and deferred-buffer energy --- across the six flexible loads.
\end{itemize}
The dimensions sum to $3N + 2N_\ell + 2 + 6{\times}5 = 99 + 64 + 2 + 30 = 195$.

The action is two-headed and per device:
\begin{itemize}\setlength{\itemsep}{1pt}
  \item $\mathbf{d}^{\mathrm{curtail}}_{t}\in[-1,1]^{6}$: curtailment intent, clipped to $[0,1]$ inside each FlexLoad bundle and then multiplied by that device's curtailment cap, applied at the current slot (curtailed energy is permanently lost);
  \item $\mathbf{d}^{\mathrm{shift}}_{t}\in[-1,1]^{6}$: load-shift intent, clipped to $[0,1]$ and multiplied by that device's shift cap, postponing demand to a later slot inside the $H{=}4$-step (two-hour) shift horizon (shifted energy must be repaid within $H$ steps).
\end{itemize}

\textbf{Reward and Violation Cost.}
The reward is the negative total network loss $P^{\mathrm{loss}}_{t}$ (in MW summed over all $N_\ell$ branches). The violation cost $C^{\mathrm{voltage}}_{t}$ is the count of buses whose voltage falls outside the safe band on the current half-hour slot (an integer in $[0, N]$, then averaged into the per-step rate reported in Section~\ref{app:I.3}). This channel is reported under Form~1 (budgeted expected cost):
\begin{equation}
\label{eq:dso-reward-cost}
r_{t}=-P^{\mathrm{loss}}_{t},
\qquad
c_{t}=C^{\mathrm{voltage}}_{t}.
\end{equation}

\textbf{Baselines.}
Three heuristic baselines are released:
\begin{itemize}\setlength{\itemsep}{1pt}
  \item \emph{No Control}: zero curtailment, zero shift;
  \item \emph{TOU} (time-of-use): fixed peak-window rule that curtails and shifts demand during 16:00--21:00 each day;
  \item \emph{Droop}: a local voltage droop response that reacts only to each controllable load's own bus voltage.
\end{itemize}
Four learned baselines are trained on the same observation and reward: PPO, SAC, Sauté PPO, and PPO-Lagrangian.

\textbf{Challenge.}
DSO has a smooth and continuous action space, but the reward (loss) and the cost (voltage) point in opposite directions: lowering loss often pushes voltages closer to the band edges, and the band must hold for every bus on every half-hour slot. Voltage couples the six controllers through the shared power flow equations, so learning has to discover a joint policy that respects a global constraint from local actions, which is the standard hard regime for safe-RL methods even when control itself is smooth.

\subsection{Distributed energy resource coordination task (DERs)}
\label{app:E.4}
\begin{figure}[h]
  \centering
  \includegraphics[width=1.00\linewidth]{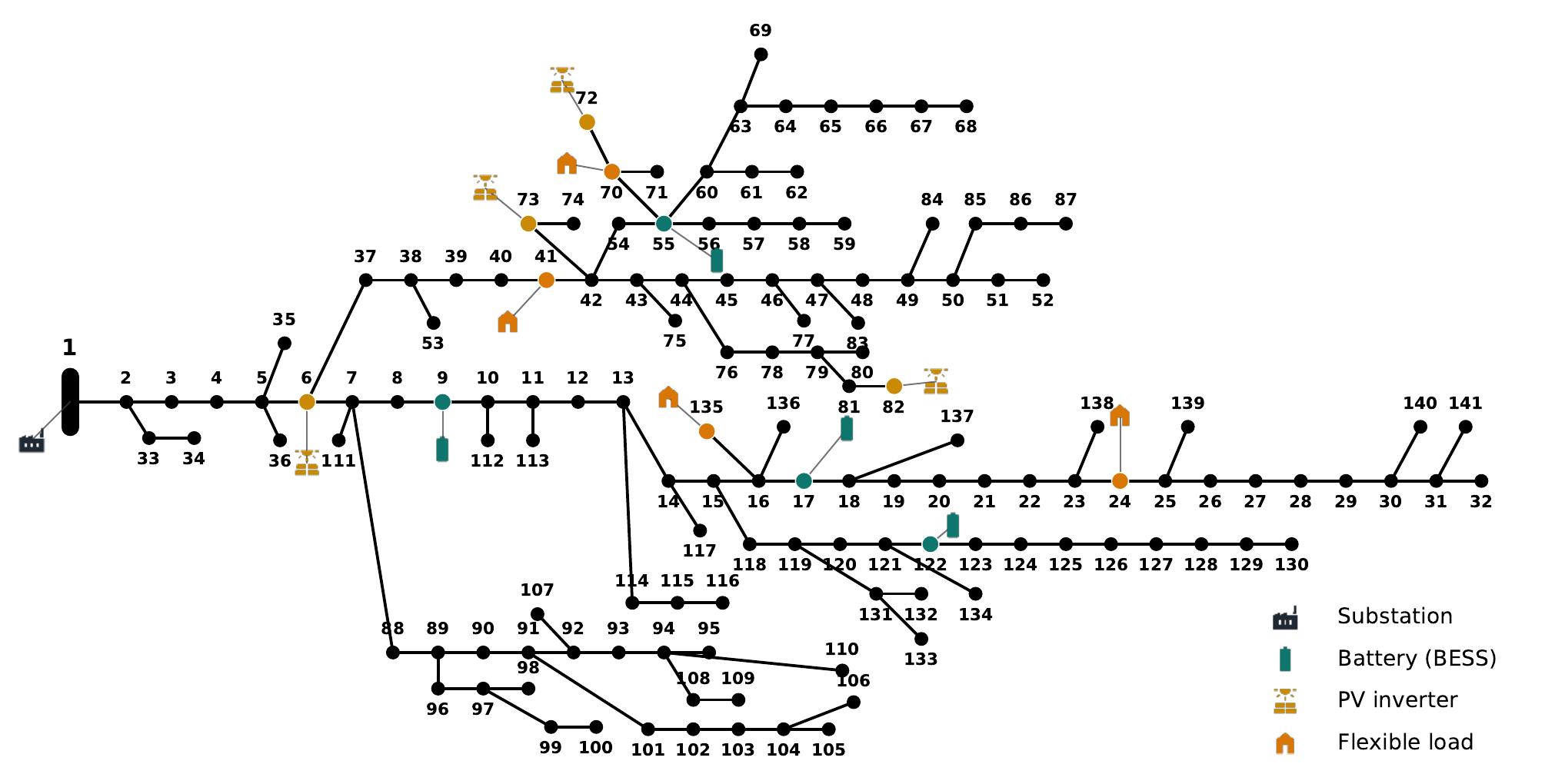}
  \caption{141-bus radial distribution feeder (\texttt{case141}) used by the DERs task. The substation (factory icon) at bus 1 supplies the network through deep radial branches. Twelve DER agents are sited at fixed buses: 4 batteries (BESS, teal vertical bars) at buses 9, 17, 55, 122; 4 PV inverters (sun-panel icons) at buses 6, 72, 73, 82; and 4 flexible loads (orange houses) at buses 24, 41, 70, 135. Small black dots mark passive buses without an attached DER agent. The 12 agents share a team reward (negative network active power loss) and cooperate under partial observability with $K{=}4$ BFS-graph-neighbor windows (Section~\ref{app:E.4}).}
  \label{fig:ders-case141}
\end{figure}

\textbf{Task.}
Twelve heterogeneous distributed energy resources --- four batteries, four PV (photovoltaic) inverters, and four flexible loads --- sit at fixed buses of the 141-bus radial distribution feeder \texttt{case141} (see Figure~\ref{fig:ders-case141}). The horizon is 24 hours at 30-minute resolution, which gives 48 sequential decision steps per episode, and is driven by real Ausgrid demand profiles\footnote{Ausgrid distribution demand traces, \url{https://www.ausgrid.com.au/Industry/Our-Research/Data-to-share/}.} for the bus-level loads and GB solar availability traces\footnote{GB solar from Elexon BMRS, \url{https://bmrs.elexon.co.uk/generation-by-fuel-type}.} for the PV bundles. Each device is its own agent. The agents cooperate to minimize total active power loss while keeping every bus voltage inside the safe band $[0.94, 1.06]$~p.u.\ and respecting line thermal limits. The actions are heterogeneous: the same 2-dimensional action slot means different things for batteries, PV inverters, and flexible loads, so this is a partially observable cooperative MARL task with agents grouped by type.

\textbf{Observation and Action.}
The 141-bus network is too large for every agent to read every bus, so each agent samples the grid only through a small local window. Agent $i$ sits at bus $b(i)$ and reads (i) its own bus voltage, (ii) the voltages at its $K{=}4$ nearest BFS-graph-neighbor buses (the part of the feeder it directly affects), (iii) a three-number summary of the system-wide voltage profile (its only global signal), and (iv) its own device internal state:
\begin{equation}
\label{eq:ders-state-action}
\mathbf{o}_{i,t}=\bigl(V_{b(i),t},\ \mathbf{V}_{\mathcal{N}_{K}(i),t},\ \widetilde{\mathbf{V}}_{t},\ \tau_{t},\ \mathbf{z}^{\mathrm{device}}_{i,t}\bigr) \in \mathbb{R}^{15}.
\end{equation}
The action is per-agent 2-dimensional, $\mathbf{a}_{i,t}\in[-1,1]^{2}$, with type-specific physical interpretation:
\begin{equation}
\mathbf{a}_{i,t}=
\begin{cases}
\bigl(P^{\mathrm{bat}}_{i,t},\ Q^{\mathrm{bat}}_{i,t}\bigr) & \text{battery},\\
\bigl(\kappa^{\mathrm{pv}}_{i,t},\ Q^{\mathrm{pv}}_{i,t}\bigr) & \text{PV inverter},\\
\bigl(d^{\mathrm{curtail}}_{i,t},\ d^{\mathrm{shift}}_{i,t}\bigr) & \text{flexible load}.
\end{cases}
\end{equation}
The five observation entries are:
\begin{itemize}\setlength{\itemsep}{1pt}
  \item $V_{b(i),t}\in\mathbb{R}$: voltage at the agent's own bus, centered and scaled the same way as in the DSO observation;
  \item $\mathbf{V}_{\mathcal{N}_{K}(i),t}\in\mathbb{R}^{K}$: voltages at the agent's $K{=}4$ nearest BFS-graph-neighbor buses, ordered by graph distance on the feeder topology;
  \item $\widetilde{\mathbf{V}}_{t}=(V_{\min},V_{\max},V_{\mathrm{mean}})\in\mathbb{R}^{3}$: min, max, and mean of the full bus voltage vector --- a coarse global signal that does not reveal per-bus details;
  \item $\tau_{t}\in\mathbb{R}^{2}$: sin/cos time-of-day encoding;
  \item $\mathbf{z}^{\mathrm{device}}_{i,t}\in\mathbb{R}^{5}$: the agent's own device internal state, sized to the maximum per-device observation length across DER types and zero-padded for shorter types so all twelve agents share a fixed 15-dimensional input shape under typed parameter sharing. By type: $3$ entries [SoC, normalized $p$, normalized $q$] for a battery; $4$ entries [capacity factor (current solar output relative to nameplate), dispatched $p$, dispatched $q$, curtailment level] for a PV inverter; $5$ entries [current curtailment, outgoing shift fraction, incoming shift fraction, deferred-buffer fill ratio, deferred-buffer energy] for a flexible load.
\end{itemize}
The dimension count is $1 + K + 3 + 2 + 5 = 15$.

The action is type-specific (the same 2-dim slot but different physical meaning):
\begin{itemize}\setlength{\itemsep}{1pt}
  \item \emph{Battery}: $(P^{\mathrm{bat}}_{i,t}, Q^{\mathrm{bat}}_{i,t})$ active and reactive power, denormalized linearly from $[-1,1]^{2}$ to the inverter capability envelope $\{(p,q):p^2+q^2\le(S^{\max}_i)^2\}$ (Section~\ref{app:D.4}) and clipped against the SoC-feasible discharge / charge bounds;
  \item \emph{PV inverter}: $(\kappa^{\mathrm{pv}}_{i,t}, Q^{\mathrm{pv}}_{i,t})$ a curtailment factor $\kappa\in[0,1]$ (fraction of available solar power to use) and a reactive-power setpoint, projected onto the same PQ envelope (Section~\ref{app:D.4}) with $p$ priority;
  \item \emph{Flexible load}: $(d^{\mathrm{curtail}}_{i,t}, d^{\mathrm{shift}}_{i,t})$ curtailment and load-shift fractions exactly as in the DSO task --- clipped to $[0,1]$, scaled by per-device caps, with a shift horizon $H{=}4$ steps.
\end{itemize}

\textbf{Reward and Violation Cost.}
All twelve agents share a single team reward, the negative total active power loss $P^{\mathrm{loss}}_{t}$ in MW summed over all $N_\ell$ branches:
\begin{equation}
\label{eq:ders-reward-cost}
r_{t}=-P^{\mathrm{loss}}_{t},
\qquad
\mathbf{c}_{t}=\bigl(C^{\mathrm{voltage}}_{t},\ C^{\mathrm{thermal}}_{t},\ C^{\mathrm{resource}}_{t}\bigr).
\end{equation}
The three-channel violation cost separately reports:
\begin{itemize}\setlength{\itemsep}{1pt}
  \item $C^{\mathrm{voltage}}_{t}$: count of buses whose voltage falls outside the safe band $[0.94,1.06]$~p.u.\ on the current step;
  \item $C^{\mathrm{thermal}}_{t}$: count of branches whose apparent power flow $|S|$ exceeds the line capacity on the current step;
  \item $C^{\mathrm{resource}}_{t}$: per-step bundle physical cost --- e.g.\ an opportunity cost when a PV inverter curtails available solar power, or a clipping penalty when a battery dispatch hits its SoC bounds.
\end{itemize}
All three channels are reported under Form~1 (budgeted expected cost).

\textbf{Baselines.}
Two heuristic baselines are released:
\begin{itemize}\setlength{\itemsep}{1pt}
  \item \emph{No Control}: every device idle (battery $P{=}Q{=}0$, PV at MPPT with no curtailment, flexible loads serving nominal demand);
  \item \emph{Voltage Droop}: a local reactive-power droop rule keyed to each agent's own bus voltage.
\end{itemize}
Three learned baselines use type-specific parameter sharing in IPPO, with one shared actor-critic per agent type:
\begin{itemize}\setlength{\itemsep}{1pt}
  \item \emph{IPPO}: the reference variant, trained with a voltage-violation penalty of weight 4.0;
  \item \emph{IPPO-rs}: the same with the penalty doubled to 8.0;
  \item \emph{IPPO-Lagrangian}: separate cost critics for the three cost channels with a shared dual variable across types.
\end{itemize}

\textbf{Challenge.}
DERs combines partial observability (each agent only sees a $K{=}4$ local window plus a three-number global summary), heterogeneous agents (batteries, PV, and flexible loads share a reward but not an action interface), and a network-wide constraint (voltage band) that couples all twelve agents through the power flow equations. The reward is a single global signal (loss in MW), while the safety violations can fire from any one of the twelve agents, so credit assignment is hard. These remain open problems for cooperative MARL, especially at feeder sizes where no agent can hold a global view.

\subsection{Data Center Microgrid task (DCMG)}
\label{app:E.5}
\textbf{Task.}
A microgrid powers an AI data center with on-site PV, battery storage, a diesel generator, and a connection to the main grid. A single operator schedules training and fine-tuning workloads, cooling, battery dispatch, diesel generation, and grid import over a 24-hour horizon at 5-minute resolution (288 steps per episode), driven by real Google data center workload traces~\citep{GoogleClusterdata2026}, GB solar availability traces\footnote{GB solar from Elexon BMRS, \url{https://bmrs.elexon.co.uk/generation-by-fuel-type}.} for the on-site PV, and GB MID market grid-import prices\footnote{GB Market Index Data (MID) from Elexon BMRS, \url{https://data.elexon.co.uk/bmrs/api/v1/datasets/MID}.}. The microgrid runs grid-connected by default; under simulated outage scenarios it switches to islanded mode where unmet demand counts as a hard violation. The setting is a long-horizon, multi-objective single-agent RL problem: trade off energy use, cost, and carbon while respecting service-level, thermal, and power-balance constraints.

\textbf{Observation and Action.}
At each step the operator observes a joint workload, thermal, and energy snapshot:
\begin{equation}
\label{eq:dcmg-state-action}
\begin{aligned}
\mathbf{o}_{t} = \bigl(&u^{\mathrm{cpu}}_{t},\ u^{\mathrm{mem}}_{t},\ q^{\mathrm{train}}_{t},\ q^{\mathrm{ft}}_{t},\ \eta_{t},\ T^{\mathrm{zone}}_{t},\ T^{\mathrm{out}}_{t},\ \mathrm{COP}_{t},\ \alpha^{\mathrm{pv}}_{t},\ \mathrm{SoC}_{t},\\
&m^{\mathrm{dg}}_{t},\ p^{\mathrm{load}}_{t},\ p^{\mathrm{net}}_{t},\ \beta^{\mathrm{dis}}_{t},\ \beta^{\mathrm{chg}}_{t},\ \pi^{\mathrm{grid}}_{t},\ \overline{\pi}^{\mathrm{grid},6h}_{t},\ \mathbf{a}_{t-1},\ \tau_{t}\bigr) \in \mathbb{R}^{24}.
\end{aligned}
\end{equation}
\begin{equation}
\mathbf{a}_{t}=\bigl(a^{\mathrm{train}}_{t},\ a^{\mathrm{ft}}_{t},\ a^{\mathrm{cool}}_{t},\ a^{\mathrm{batt}}_{t},\ a^{\mathrm{dg}}_{t}\bigr) \in [0,1]^{3}\times[-1,1]\times[0,1].
\end{equation}
The 24 observation entries split into five thematic blocks plus the previous action and a time encoding:
\begin{itemize}\setlength{\itemsep}{1pt}
  \item \emph{Workload state} (5 dims): $u^{\mathrm{cpu}}_t, u^{\mathrm{mem}}_t \in [0,1]$ are GPU compute and memory utilization; $q^{\mathrm{train}}_t, q^{\mathrm{ft}}_t \in [0,1]$ are the training and fine-tuning queue fills (waiting GPU demand normalized by total GPU count); $\eta_t \in [-1,1]$ is the deadline urgency of the most pressing waiting job (negative means already overdue);
  \item \emph{Thermal state} (3 dims): $T^{\mathrm{zone}}_t \in [0,1]$ the server-zone temperature, $T^{\mathrm{out}}_t \in [0,1]$ the outdoor temperature (both centered and scaled to the operating range), and $\mathrm{COP}_t \in [0,1]$ the chiller's coefficient of performance (higher means more cooling delivered per unit electricity);
  \item \emph{Energy assets} (3 dims): $\alpha^{\mathrm{pv}}_t \in [0,1]$ the PV capacity factor (current solar output relative to nameplate), $\mathrm{SoC}_t \in [0,1]$ the battery state of charge, and $m^{\mathrm{dg}}_t \in [0,1]$ the remaining diesel headroom;
  \item \emph{Power balance and battery headroom} (4 dims): $p^{\mathrm{load}}_t \in [0,1]$ the total data-center electrical load (normalized by total supply capacity), $p^{\mathrm{net}}_t \in [-1,1]$ the residual net load after PV (signed), and $\beta^{\mathrm{dis}}_t, \beta^{\mathrm{chg}}_t \in [0,1]$ the battery's discharge and charge headroom as fractions of rated power;
  \item \emph{Grid price signal} (2 dims): $\pi^{\mathrm{grid}}_t \in [0,1]$ the current GB MID grid-import price (normalized) and $\overline{\pi}^{\mathrm{grid},6h}_t \in [0,1]$ the maximum grid price expected over the next six hours, giving the policy a forward look-ahead for arbitrage;
  \item \emph{Previous action and time encoding} (5 + 2 dims): $\mathbf{a}_{t-1} \in \mathbb{R}^5$ the previous-step action and $\tau_t \in \mathbb{R}^2$ the sin/cos time-of-day encoding.
\end{itemize}
The dimension count is $11 + 6 + 5 + 2 = 24$.

The 5-dimensional action has mixed bounds:
\begin{itemize}\setlength{\itemsep}{1pt}
  \item $a^{\mathrm{train}}_t, a^{\mathrm{ft}}_t \in [0,1]$: fractions of the training and fine-tuning queues dispatched in the next 5-minute slot;
  \item $a^{\mathrm{cool}}_t \in [0,1]$: cooling setpoint command (lower means a colder zone setpoint, requiring more cooling power);
  \item $a^{\mathrm{batt}}_t \in [-1,1]$: battery dispatch signal (negative charges, positive discharges; scaled by rated power and clipped against SoC-feasible bounds inside the bundle);
  \item $a^{\mathrm{dg}}_t \in [0,1]$: diesel generator preference --- realized as a same-step residual slack actuator after the workload, cooling, and battery decisions are applied, so $a^{\mathrm{dg}}$ acts as a soft preference rather than a strict setpoint.
\end{itemize}

\textbf{Reward and Violation Cost.}
The reward sums an energy-use term, a grid-cost term, and a carbon-emissions term, with scalar weights $w_{\mathrm{cost}}, w_{\mathrm{carbon}}$ that the operator can pick to encode a Pareto preference (the canonical benchmark uses $w_{\mathrm{cost}}=0.2$, $w_{\mathrm{carbon}}=0.1$):
\begin{equation}
\label{eq:dcmg-reward-cost}
r_{t}=r^{\mathrm{energy}}_{t}+w_{\mathrm{cost}}\,r^{\mathrm{cost}}_{t}+w_{\mathrm{carbon}}\,r^{\mathrm{carbon}}_{t},
\qquad
\mathbf{c}_{t}=\bigl(C^{\mathrm{SLA}}_{t},\ C^{\mathrm{temp}}_{t},\ C^{\mathrm{pwr}}_{t}\bigr).
\end{equation}
The three-channel violation cost separately reports:
\begin{itemize}\setlength{\itemsep}{1pt}
  \item $C^{\mathrm{SLA}}_t$: workload incompletion --- the per-step count of expired tasks normalized by the GPU count;
  \item $C^{\mathrm{temp}}_t$: server-zone over-temperature, $C^{\mathrm{temp}}_t = (T^{\mathrm{zone}}_t - T^{\max})_+$;
  \item $C^{\mathrm{pwr}}_t$: per-step power deficit --- the shortfall when on-site generation, battery, and grid import together cannot serve the data-center load. This channel is the hard constraint that becomes binding under simulated outage scenarios when the microgrid is forced into islanded mode.
\end{itemize}
All three channels are reported under Form~1 (budgeted expected cost).

\textbf{Baselines.}
Three heuristic baselines are released:
\begin{itemize}\setlength{\itemsep}{1pt}
  \item \emph{No Control}: always-on policy with no scheduling;
  \item \emph{Max renewable}: serve workload from PV first, then battery, then diesel;
  \item \emph{Rule-based}: a heuristic scheduler that respects SLA, temperature, and SoC bands.
\end{itemize}
Two learned baselines are trained with fixed penalty weights on the SLA, over-temperature, and deficit channels added to the reward:
\begin{itemize}\setlength{\itemsep}{1pt}
  \item \emph{PPO} with a Beta distribution policy head for bounded actions;
  \item \emph{SAC} with a tanh-squashed Gaussian.
\end{itemize}

\textbf{Challenge.}
DCMG is a long-horizon (288 steps), multi-objective single-agent task with three hard constraint channels. The objectives (energy, cost, carbon) and the constraints (SLA, over-temperature, power deficit) are cross-coupled: deferring training jobs saves energy but risks SLA misses; running cooling harder protects servers but burns diesel; charging the battery during PV peaks improves later resilience but uses up current SLA headroom. Credit must be propagated across hundreds of 5-minute steps under stochastic workload, weather, and PV traces, which is precisely the regime that current model-free RL handles poorly.

\FloatBarrier

\section{Data Sources and Splits}
\label{app:F}

\subsection{Data and splits}
\label{app:F.1}

Table~\ref{tab:data-provenance} summarizes the external time-series traces used in the reported results, the network case for each task, the splits, and the seed and episode budget. Processed copies of these traces are shipped with the PowerZooJax code release.

\begingroup
\footnotesize
\setlength{\tabcolsep}{3pt}
\setlength{\LTleft}{0pt}
\setlength{\LTright}{0pt}
\setlength{\LTcapwidth}{\linewidth}
\begin{longtable}{@{}>{\raggedright\arraybackslash}p{0.08\linewidth}>{\raggedright\arraybackslash}p{0.14\linewidth}>{\raggedright\arraybackslash}p{0.22\linewidth}>{\raggedright\arraybackslash}p{0.32\linewidth}>{\raggedright\arraybackslash}p{0.18\linewidth}@{}}
  \caption{External time-series traces, splits, and seed and episode budgets per task. GB denotes Great Britain. Bold marks the primary split used for the headline metric reported in Section~\ref{app:I}; for DCMG the appendix-only splits follow the semicolon.}
  \label{tab:data-provenance}\\
  \toprule
  Task & System / case & Time-series source(s) & Splits & Seed budget \\
  \midrule
  \endfirsthead
  \toprule
  Task & System / case & Time-series source(s) & Splits & Seed budget \\
  \midrule
  \endhead
  \midrule
  \endfoot
  \bottomrule
  \endlastfoot
  GenCos & 5-bus market         & GB demand                                & train, \textbf{in-distribution}, demand shift, renewable shock                                                                           & 5 seeds, 30 episodes/seed \\
  \midrule
  TSO    & 118-bus transmission & GB demand, wind, and solar               & train, \textbf{in-distribution}, load stress, line tightening                                                                            & 5 seeds, 50 episodes/seed \\
  \midrule
  DSO    & 33-bus feeder        & Ausgrid distribution demand              & \textbf{in-distribution}                                                                                                                  & 5 seeds, 50 episodes/seed \\
  \midrule
  DERs   & 141-bus feeder       & Ausgrid distribution demand and GB solar & train, \textbf{in-distribution}, voltage tightening, PV shift, load stress                                                                & 5 seeds, 30 episodes/seed \\
  \midrule
  DCMG   & Behind-the-meter microgrid & Google workload (Azure and Alibaba for workload-OOD splits); GB solar; GB MID price & train, \textbf{in-distribution}, cooling stress, renewable drought; appendix: workload swap, workload shock, dg derating, SLA tightening & 5 seeds, 10 episodes/seed \\
\end{longtable}
\endgroup

\subsection{Split and stress-test taxonomy}
\label{app:F.2}

Each split asks a different physical question. Table~\ref{tab:shift-taxonomy} groups them by the mechanism that changes between train and evaluation or between routine and stressed evaluation, and states the role of each split.

\begingroup
\footnotesize
\setlength{\tabcolsep}{3pt}
\setlength{\LTleft}{0pt}
\setlength{\LTright}{0pt}
\setlength{\LTcapwidth}{\linewidth}
\begin{longtable}{@{}>{\raggedright\arraybackslash}p{0.17\linewidth}>{\raggedright\arraybackslash}p{0.25\linewidth}>{\raggedright\arraybackslash}p{0.26\linewidth}>{\raggedright\arraybackslash}p{0.28\linewidth}@{}}
  \caption{Split and stress-test taxonomy. Each row states the physical mechanism and the role of the split.}
  \label{tab:shift-taxonomy}\\
  \toprule
  Split type & Tasks / split names & Physical mechanism & Role \\
  \midrule
  \endfirsthead
  \toprule
  Split type & Tasks / split names & Physical mechanism & Role \\
  \midrule
  \endhead
  \midrule
  \endfoot
  \bottomrule
  \endlastfoot
  Routine in-distribution & TSO, DSO, DERs, GenCos, DCMG / \textbf{in-distribution} & Held-out episodes from the routine evaluation regime & Main task comparison for the primary metric \\
  \midrule
  Load / demand shift & TSO / load stress; DERs / load stress; GenCos / demand shift & Higher or shifted demand changes reserve pressure, feeder loading, or market clearing & Whether an in-distribution policy remains useful when demand changes \\
  \midrule
  Renewable / PV shift & DERs / PV shift; GenCos / renewable shock; DCMG / renewable drought & Renewable generation changes the controllable/exogenous balance & Sensitivity to renewable availability \\
  \midrule
  Network / security tightening & TSO / line tightening; DERs / voltage tightening & Tighter physical operating limits (line capacity or voltage band) & Whether lower cost or lower loss survives under tighter operating limits \\
  \midrule
  Cooling / workload stress & DCMG / cooling stress (main); workload swap, workload shock, SLA tightening (appendix) & Data center thermal load, workload mix, or service deadlines change & Long-horizon resource scheduling under thermal, workload, or service stress \\
  \midrule
  Resource-availability stress & DCMG / dg derating & Backup-generation capacity is reduced & Sensitivity to reduced backup capacity \\
\end{longtable}
\endgroup

\section{Quickstart and API}
\label{app:G}

This section is a copy-pasteable walk-through of how to install PowerZooJax, run a minimal rollout, read a task config, read a training config, and trigger a benchmark run from the CLI. Listings~E.1--E.3 are minimal Python demos and the \texttt{TaskSpec} Protocol surface, Listings~E.4--E.5 are abridged from the released config files, and the bash blocks in Sections~\ref{app:G.1} and~\ref{app:G.5} cover install and CLI usage.

\subsection{Installation}
\label{app:G.1}

PowerZooJax requires Python 3.10--3.12. The base install ships a CPU-only JAX; CUDA 12 support is opt-in via the \texttt{cuda12} extra; the JAX single-agent RL stack lives behind \texttt{rl} (\texttt{rejax}, \texttt{distrax}) and the multi-agent stack behind \texttt{marl} (\texttt{jaxmarl}); the benchmark workflow lives behind \texttt{benchmarks} (Stable-Baselines3, SBX, PettingZoo).

\begin{bashplain}
git clone https://github.com/powerzoojax/PowerZooJax && cd PowerZooJax
pip install -e ".[rl,marl,benchmarks]"               # CPU JAX
pip install -e ".[cuda12,rl,marl,benchmarks]"        # Linux + CUDA 12

# JAX preallocates ~75% of VRAM on first GPU use; disable if sharing the GPU:
export XLA_PYTHON_CLIENT_PREALLOCATE=false
\end{bashplain}

\subsection{Minimal rollout}
\label{app:G.2}

The example below opens a transmission environment on \texttt{case5}, resets it, then runs a 48-step rollout under \texttt{jax.lax.scan} with a fixed action schedule. Every transition stays inside JIT.

\begin{lstbox}{Listing E.1: Minimal rollout on case5 (Python)}
\begin{lstlisting}[style=pzjpython]
import jax, jax.numpy as jnp
from powerzoojax.case import create_case5
from powerzoojax.envs import TransGridEnv, make_trans_params
from powerzoojax.utils import scan_rollout

env = TransGridEnv()
params = make_trans_params(create_case5())
key = jax.random.PRNGKey(0)
reset_key, rollout_key = jax.random.split(key)
obs, state = env.reset(reset_key, params)
actions = jnp.zeros((48, *env.action_space(params).shape), dtype=jnp.float32)
final_state, obs_traj, reward_traj, cost_traj, done_traj, info_traj = scan_rollout(
    env, rollout_key, state, params, actions
)
\end{lstlisting}
\end{lstbox}

The same env contract works for the other physical models via the task-level wrapper:

\begin{lstbox}{Listing E.2: Task-level interface (Python)}
\begin{lstlisting}[style=pzjpython]
from powerzoojax.tasks.tso import TSOTask
import jax
task = TSOTask()
env = task.make_env(split="iid")
params = task.episode_params(
    "iid", episode_idx=0, n_episodes=1, max_steps=48,
    strategy="seeded", seed=0,
)
obs, state = env.reset(jax.random.PRNGKey(0), params)
\end{lstlisting}
\end{lstbox}

The five task classes live in \texttt{powerzoojax.tasks.\{tso,dso,ders,gencos,dc\_microgrid\}} and all satisfy the \texttt{TaskSpec} Protocol (Listing~E.3, abridged from \texttt{powerzoojax/tasks/base.py}).

\begin{lstbox}{Listing E.3: \texttt{TaskSpec} Protocol surface (\texttt{powerzoojax/tasks/base.py}, abridged)}
\begin{lstlisting}[style=pzjpython]
from typing import Any, Literal, Protocol, runtime_checkable

@runtime_checkable
class TaskSpec(Protocol):
    """Structural interface for the five benchmark tasks."""

    task_name: str                          # "dso" | "tso" | "ders" | "gencos" | "dc_microgrid"
    default_splits: tuple[str, ...]         # e.g. ("train", "iid", "load_stress", ...)

    def make_env(self, split: str = "train") -> Any: ...

    def episode_params(
        self,
        split: str, episode_idx: int, n_episodes: int, max_steps: int,
        *, strategy: Literal["uniform", "seeded"] = "uniform", seed: int = 0,
    ) -> Any: ...

    def rollout(self, env: Any, params: Any, key: Any, policy_fn: Any) -> Any: ...
    def constraint_spec(self) -> "ConstraintSpec": ...
\end{lstlisting}
\end{lstbox}

\subsection{Reading a task config}
\label{app:G.3}

Every benchmark task is specified by a YAML file at \texttt{benchmarks/<task>/configs/task.yaml}. The TSO file is reproduced below in abridged form.

\begin{lstbox}{Listing E.4: TSO task config (\texttt{benchmarks/tso/configs/task.yaml}, abridged)}
\begin{lstlisting}[style=pzjyaml]
task: tso
case: case118                 # IEEE 118-bus / 186-line transmission case
n_units: 54                   # Thermal units committed by SCUC
max_steps: 48                 # 48 half-hour intervals = 24 h horizon
dt_hours: 0.5

data_source: gb               # Real GB transmission demand & wind trace
gb_train_start: "2025-04-01"
gb_train_end:   "2025-12-31"
gb_iid_start:   "2026-01-01"
gb_iid_end:     "2026-03-31"

reserve_margin_frac: 0.05     # SCUC reserve margin
reward_scale: 0.0001          # reward scaling factor
solver_mode: 1                # DC-OPF solver mode
dcopf_max_iter: 200
dcopf_tol: 0.0001
forecast_horizon_steps: 4     # 4 steps = 2 h; sets TSO obs to 410-dim (Section C.2)

baseline_set: [all_on, merit_order]
eval_splits:  [train, iid, load_stress, line_tightening]
primary_split: iid
seeds: [0, 1, 2, 3, 4]
eval_episodes: 50

safety_thresholds:
  reserve_shortfall_rate: 0.0
  thermal_violation_rate: 0.0
\end{lstlisting}
\end{lstbox}

\subsection{Reading a training config}
\label{app:G.4}

Training-side hyperparameters live next to the task config as \texttt{train\_<algo>.yaml}. The TSO PPO file is below; the master per-algorithm-per-task hyperparameter table is in Section~\ref{app:H.2}.

\begin{lstbox}{Listing E.5: TSO PPO config (\texttt{benchmarks/tso/configs/train\_ppo.yaml})}
\begin{lstlisting}[style=pzjyaml]
algo: ppo
total_timesteps: 20000000
num_envs: 256                 # GPU parallelism for case118 (54 units, 410-dim obs)
n_steps: 48                   # one episode per update
hidden_dims: [256, 256]
gamma: 0.995
gae_lambda: 0.95
lr: 0.0003
normalize_observations: true
wrapper: log                  # use 'safe' for CMDP cost channel
eval_freq: 100000
eval_episodes: 8              # in-training monitor only; benchmark eval uses task.yaml::eval_episodes
record_eval_wall_time: true   # logs per-eval walltime for cross-backend curves
\end{lstlisting}
\end{lstbox}

\subsection{CLI commands}
\label{app:G.5}

The benchmark workflow exposes five entry points per task: \texttt{baseline} (run non-learning baselines), \texttt{train} (run a learning algorithm), \texttt{eval} (evaluate a trained run on a split), \texttt{summarize} (rebuild result summaries), and \texttt{plots} (regenerate paper figures). The TSO end-to-end example is

\begin{bashplain}
# Sanity check: run 5 seeds of the All-On and Merit-Order baselines.
python benchmarks/tso/run.py baseline --seeds 0,1,2,3,4

# Train PPO from train_ppo.yaml.
python benchmarks/tso/run.py train --algo ppo --seed 0

# Evaluate a trained run on the in-distribution split and the line-tightening stress split.
python benchmarks/tso/run.py eval --run-id <id> --split iid
python benchmarks/tso/run.py eval --run-id <id> --split line_tightening

# Rebuild result tables.
python benchmarks/tso/run.py summarize

# Regenerate figures from the result summary.
python benchmarks/tso/run.py plots
\end{bashplain}

A lighter ad-hoc CLI lives at \texttt{python -m powerzoojax} for preset and YAML-driven runs:

\begin{bashplain}
python -m powerzoojax --list-presets
python -m powerzoojax --preset case5-economic-dispatch --seed 0
python -m powerzoojax --preset case5-economic-dispatch --config experiment.yaml --output result.json
\end{bashplain}

\FloatBarrier

\section{Training Setup and Hyperparameters}
\label{app:H}

This section gives the algorithm-choice rationale, full per-task per-algorithm hyperparameters, network architectures, and evaluation protocol behind every reported number in the paper.

\subsection{Algorithm choices}
\label{app:H.1}

\begin{itemize}
\item \textbf{TSO, DSO, DCMG (single-agent).} PPO~\citep{ProximalPolicyOptimization2017schulman} as the unconstrained baseline; PPO-Lagrangian as the safe-RL baseline on TSO and DSO, with Sauté PPO~\citep{SauteRLAlmost2022sootla} additionally reported on DSO. DSO and DCMG additionally train SAC~\citep{SoftActorCriticAlgorithms2018haarnoja}; on DCMG, SAC supplies the appendix PPO-vs-SAC comparison in Section~\ref{app:I.5} and the representative episode in main-text Section~\ref{sec:6.4}.
\item \textbf{DERs (cooperative MARL).} IPPO~\citep{IndependentLearningAll2020witt} with type-specific parameter sharing: the 12 agents are partitioned into three types (4 batteries, 4 PV inverters, 4 flexible loads) and each type carries its own actor-critic. IPPO-rs (reward-shaped IPPO with the voltage penalty doubled) and IPPO-Lagrangian (separate cost critic per channel, one shared team-level dual variable) are the safety variants.
\item \textbf{GenCos (competitive MARL).} IPPO with full parameter sharing: a single ActorCritic is shared across the five GenCos, where each agent is conditioned on its private 12-dim observation and trained on its own profit reward. Truthful, uniform-mid, and max-markup are reported as non-learning strategic baselines.
\end{itemize}

PPO-Lagrangian on both TSO and DSO uses a zero cost-budget threshold ($b_i=0$). The Sauté PPO observation-augmentation budget, per-algorithm dual learning rates, and other safe-RL specifics are listed with the master table in Section~\ref{app:H.2}.

\paragraph{TSO reward scaling.} TSO trains with a reward-scaling factor $\alpha_r = 10^{-4}$; the per-step reward seen by the optimizer is
\begin{equation}
  r_t \;=\; -\alpha_r\,\bigl(C^{\mathrm{gen}}_t + C^{\mathrm{startup}}_t + C^{\mathrm{noload}}_t\bigr).
  \label{eq:tso-reward-scale}
\end{equation}
The operating-cost numbers reported in main-text Section~\ref{sec:6.3} and in Section~\ref{app:I.2} are the unscaled per-day operating cost.

\subsection{Per-task per-algorithm hyperparameters}
\label{app:H.2}

Table~\ref{tab:hparams} summarises the key training hyperparameters.

\begin{table}[h]
  \caption{Master hyperparameter table. Columns: $T_\text{tot}$ total environment steps (millions); $N_\text{envs}$ parallel envs; $n_s$ steps-per-update; $E$ PPO epochs (SAC update epochs); LR learning rate; HD hidden dims; CL clip-eps; EC entropy coef; $\gamma$ discount; $\lambda$ GAE lambda. Optimizer is Adam throughout.}
  \label{tab:hparams}
  \centering
  \footnotesize
  \begin{tabular}{llrrrrrlrrrr}
    \toprule
    Task & Algo & $T_\text{tot}$ & $N_\text{envs}$ & $n_s$ & $E$ & LR & HD & CL & EC & $\gamma$ & $\lambda$ \\
    \midrule
    GenCos      & IPPO       & 5  & 256 & 48  & 4   & 3e-4 & [128,128] & 0.2 & 0.01  & 0.995 & 0.95 \\
    TSO         & PPO        & 20 & 256 & 48  & 4   & 3e-4 & [256,256] & 0.2 & 0.01  & 0.995 & 0.95 \\
    TSO         & PPO-Lag    & 20 & 256 & 48  & 4   & 5e-5 & [256,256] & 0.2 & 0.005 & 0.995 & 0.95 \\
    DSO         & PPO        & 3  & 128 & 48  & 4   & 3e-4 & [128,128] & 0.2 & 0.01  & 0.995 & 0.95 \\
    DSO         & SAC        & 3  & 64  & --- & 1   & 3e-4 & [128,128] & --- & ---   & 0.995 & ---  \\
    DSO         & Sauté PPO  & 3  & 128 & 48  & 4   & 3e-4 & [128,128] & 0.2 & 0.01  & 0.995 & 0.95 \\
    DSO         & PPO-Lag    & 3  & 128 & 48  & 4   & 3e-4 & [128,128] & 0.2 & 0.01  & 0.995 & 0.95 \\
    DERs        & IPPO       & 10 & 128 & 48  & 4   & 1e-4 & [128,128] & 0.2 & 0.02  & 0.995 & 0.95 \\
    DERs        & IPPO-rs    & 10 & 128 & 48  & 4   & 1e-4 & [128,128] & 0.2 & 0.01  & 0.995 & 0.95 \\
    DERs        & IPPO-Lag   & 10 & 128 & 48  & 4   & 1e-4 & [128,128] & 0.2 & 0.01  & 0.995 & 0.95 \\
    DCMG        & PPO        & 1  & 64  & 288 & 12  & 1e-4 & [256,256] & 0.1 & 0.001 & 0.99  & 0.95 \\
    DCMG        & SAC        & 1  & 64  & --- & 2   & 3e-4 & [256,256] & --- & ---   & 0.99  & ---  \\
    \bottomrule
  \end{tabular}
\end{table}

Safe-RL specifics not in the table: the dual learning rate is $5\times 10^{-3}$ for TSO PPO-Lagrangian, $2\times 10^{-5}$ for DSO PPO-Lagrangian, and $5\times 10^{-5}$ for DERs IPPO-Lagrangian (three independent cost channels, one shared team-level dual variable). The Sauté PPO observation-augmentation budget on DSO is $192$ violation-step-equivalents. IPPO-rs on DERs sets the voltage penalty to $8.0$, twice the $4.0$ of IPPO.

\subsection{Network architectures}
\label{app:H.3}

\begin{itemize}
\item \textbf{MLP backbone.} Actor and critic are independent MLPs with hidden sizes from Table~\ref{tab:hparams}. IPPO (GenCos, DERs) and the CMDP wrapper used for PPO-Lagrangian and Sauté PPO use \texttt{tanh} activations; the rejax-based PPO and SAC baselines use \texttt{swish} and \texttt{relu}, respectively.
\item \textbf{Continuous policy heads.} Plain PPO via rejax (TSO PPO, DSO PPO) emits an unbounded diagonal Gaussian and clips the sampled action to the action box; the CMDP wrapper applies a \texttt{tanh} squash to the action mean before adding diagonal Gaussian noise; rejax SAC (DSO SAC, DCMG SAC) uses a \texttt{tanh}-squashed Gaussian whose support equals the action box; IPPO (GenCos, DERs) emits an unbounded diagonal Gaussian whose samples are clipped to $[-1,1]$ before the environment step. DCMG PPO uses a bounded Beta policy over the $[0,1]^3 \times [-1,1] \times [0,1]$ action box (Section~\ref{app:E.5}).
\item \textbf{Mixed action space (TSO).} The policy emits a single 108-dim continuous head; the environment decomposes the output into 54 sign-thresholded commitment intents and 54 dispatch preferences that bias the OPF cost intercept (Section~\ref{app:E.2}).
\item \textbf{MARL parameter sharing.} DERs partitions the 12 agents by type (4 batteries, 4 PV inverters, 4 flexible loads) and uses one shared actor-critic per type; GenCos shares one actor-critic across the five generation companies.
\item \textbf{Observation normalization.} The single-agent baselines train with running-mean / running-std observation normalization; the MARL baselines (GenCos and DERs IPPO variants) train on raw observations.
\end{itemize}

\subsection{Seeds and evaluation protocol}
\label{app:H.5}

Each task uses 5 seeds (\texttt{0,1,2,3,4}) and an evaluation budget of 10--50 episodes per seed (Table~\ref{tab:data-provenance}). Episode-level metrics --- operating cost, network loss, profit, and episode return --- are reported as across-seed mean $\pm$ 95\% bootstrap CI; per-task comparisons on the primary metric use a paired sign-flip permutation test on matched (seed, episode-index) pairs. Per-step rate metrics (thermal overload, reserve shortfall, voltage violation, SLA deficit) are reported as across-seed mean rates; the TSO reserve and thermal channels are reported under the zero-violation constraint of Section~\ref{app:C.2}. Wall-clock time is recorded at every evaluation point so that learning curves can be aligned across backends.

\FloatBarrier

\section{Additional Empirical Results}
\label{app:I}

This section reports the per-task supplementary tables and figures, one subsection per task (Sections~\ref{app:I.1}--\ref{app:I.5}). Task definitions are in Section~\ref{app:E}; split meanings are in Table~\ref{tab:shift-taxonomy}; per-task seed and evaluation-episode budgets are in Table~\ref{tab:data-provenance}; the across-seed mean $\pm$ 95\% bootstrap CI and the paired sign-flip permutation test are defined in Section~\ref{app:H.5}.

\subsection{GenCos}
\label{app:I.1}

\begin{figure}[!htbp]
  \centering
  \includegraphics[width=1.00\linewidth]{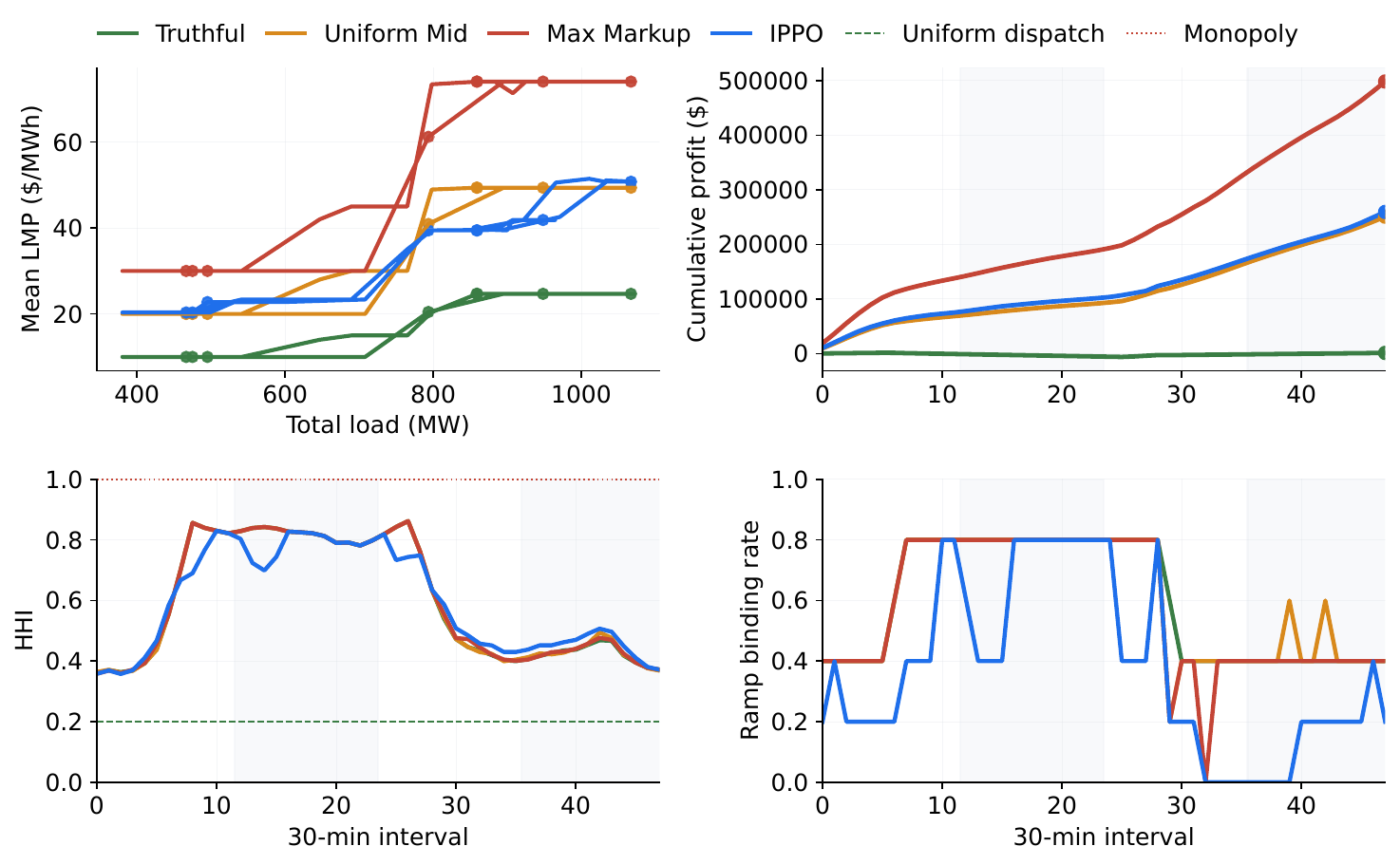}
  \caption{GenCos behavior on a representative in-distribution episode under the same physical load. Top-left: locational marginal price as a function of system load. Top-right: cumulative profit per step. Bottom-left: Herfindahl--Hirschman Index of cleared-energy shares. Bottom-right: per-step ramp-binding indicator.}
  \label{fig:gencos-policy-appendix}
\end{figure}

The task definition is given in Section~\ref{app:E.1}. We report the in-distribution baseline comparison (Table~\ref{tab:gencos-iid-appendix}), the cross-split profit table (Table~\ref{tab:gencos-splits}), and a representative-episode behavior figure (Figure~\ref{fig:gencos-policy-appendix}). Each method aggregates 5 seeds and 30 evaluation episodes per seed with 95\% bootstrap CIs (Section~\ref{app:H.5}).

The in-distribution mean total profit places IPPO ($\mathrm{GBP}\,395{,}030$, 95\% CI $[291{,}174,\ 499{,}652]$) strictly between Uniform-mid ($302{,}972$, $[297{,}376,\ 309{,}015]$) and Max-markup ($597{,}864$, $[587{,}181,\ 609{,}873]$), and well above Truthful ($6{,}934$, $[6{,}412,\ 7{,}432]$). Under the paired sign-flip permutation test of Section~\ref{app:H.5} on $5\,\text{seeds}\times 30\,\text{episodes}=150$ matched (seed, episode-index) pairs, the IPPO advantage over Truthful ($+388{,}096$) and Uniform-mid ($+92{,}058$) and the IPPO shortfall against Max-markup ($-202{,}834$) are all significant at $p<10^{-4}$. The cross-split table preserves the same $\textrm{Truthful}<\textrm{Uniform-mid}<\textrm{IPPO}<\textrm{Max-markup}$ ordering on the demand-shift and renewable-shock splits.

Beyond the ranking, the in-distribution results show two behavioral differences from the heuristics. First, the three non-learning baselines all bid the same shape and produce nearly identical Herfindahl--Hirschman Index values ($0.5664$, $0.5663$, $0.5662$ for Truthful, Uniform-mid, and Max-markup), while IPPO learns a less concentrated outcome ($\mathrm{HHI}=0.5384$). Second, the ramp-binding rate (the fraction of clearing steps in which the SCED ramp constraint is active) drops from $0.58$ for the heuristics to $0.35$ under IPPO. Figure~\ref{fig:gencos-policy-appendix} resolves these two signals over a single representative in-distribution episode: the top-left panel shows that IPPO clears at lower locational marginal prices than Max-markup at high system load, the top-right panel tracks the per-step cumulative profit gap, the bottom-left panel plots the HHI over the episode, and the bottom-right panel plots the ramp-binding indicator.

\begin{table}[h]
  \caption{GenCos in-distribution baseline comparison on \texttt{case5}, 5 seeds, 30 evaluation episodes per seed. Higher total profit is better. Profit is the across-seed mean with a 95\% bootstrap CI (Section~\ref{app:H.5}). HHI denotes the Herfindahl--Hirschman Index of cleared-energy shares.}
  \label{tab:gencos-iid-appendix}
  \centering
  \footnotesize
  \begin{tabular}{lrlr}
    \toprule
    Method & Profit mean (GBP) & Profit 95\% CI (GBP) & HHI \\
    \midrule
    Truthful     &   6{,}934 & $[6{,}412,\ 7{,}432]$         & 0.5664 \\
    Uniform-mid  & 302{,}972 & $[297{,}376,\ 309{,}015]$     & 0.5663 \\
    IPPO         & 395{,}030 & $[291{,}174,\ 499{,}652]$     & 0.5384 \\
    Max-markup   & 597{,}864 & $[587{,}181,\ 609{,}873]$     & 0.5662 \\
    \bottomrule
  \end{tabular}
\end{table}

\begin{table}[h]
  \caption{GenCos total-profit mean (GBP) across splits. Each entry aggregates 5 seeds and 30 evaluation episodes per seed.}
  \label{tab:gencos-splits}
  \centering
  \footnotesize
  \begin{tabular}{lrrrr}
    \toprule
    Split & Truthful & Uniform-mid & IPPO & Max-markup \\
    \midrule
    In-distribution &  6{,}934 & 302{,}972 & 395{,}030 & 597{,}864 \\
    Demand shift    &  9{,}634 & 354{,}978 & 475{,}620 & 702{,}906 \\
    Renewable shock &  7{,}977 & 329{,}873 & 434{,}642 & 650{,}409 \\
    \bottomrule
  \end{tabular}
\end{table}

\FloatBarrier

\subsection{TSO}
\label{app:I.2}

\begin{figure}[!htbp]
  \centering
  \includegraphics[width=1.00\linewidth]{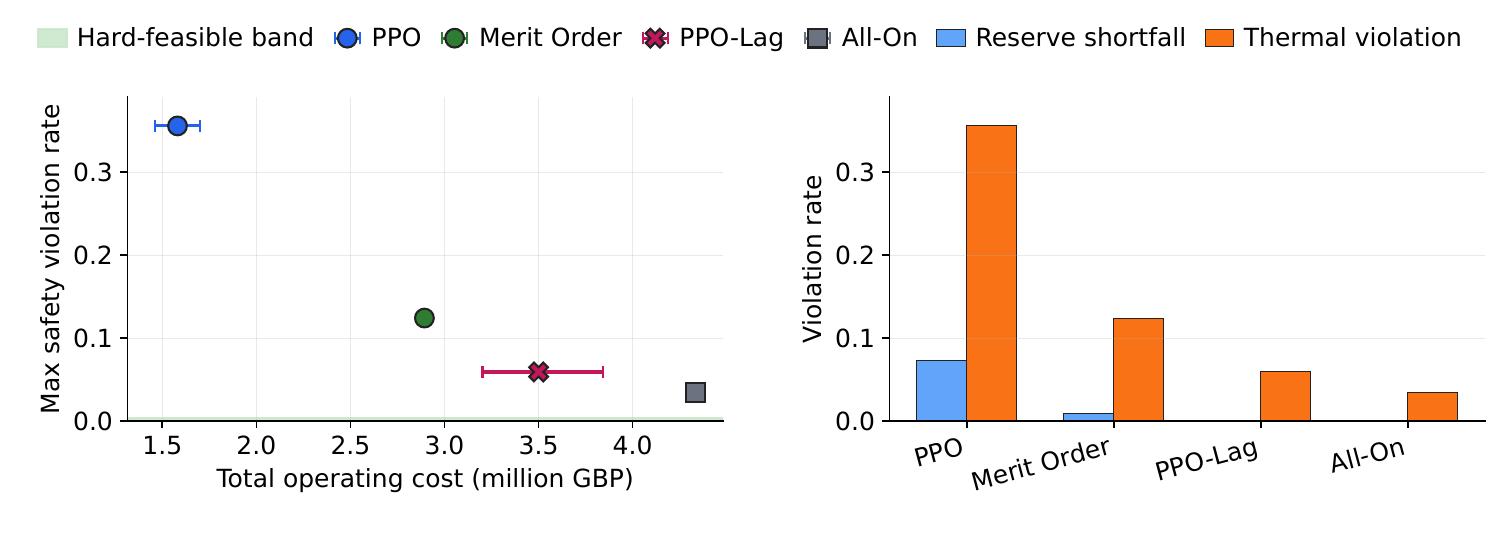}
  \caption{TSO cost--safety summary. Left: total operating cost vs the worst safety-violation rate (the maximum of the reserve-shortfall and thermal-overload rates). Right: the same methods separated by violation channel, showing that the reserve channel is satisfied for PPO-Lagrangian and All-On while the thermal channel is not satisfied for any method.}
  \label{fig:tso-safety-test-appendix}
\end{figure}

Table~\ref{tab:tso-frontier-appendix} reports operating cost and the two per-step violation rates under the zero-violation constraint of Section~\ref{app:C.2}. No method satisfies it on both channels: PPO-Lagrangian and All-On record zero reserve shortfall, while every method records a non-zero thermal-overload rate. The paired sign-flip permutation test on per-episode operating cost (Section~\ref{app:H.5}, $n=250$ pairs, $p<10^{-4}$ for the reported comparisons) confirms that the cost differences are significant. The task card and 24-hour SCUC formulation are described in Section~\ref{app:E.2}; Figure~\ref{fig:tso-safety-test-appendix} decomposes the cost-versus-violation pattern into worst-rate and per-channel views, separating the reserve channel (PPO-Lagrangian and All-On at zero) from the thermal channel (every method above zero).

\begin{table}[h]
  \caption{TSO cost and safety results. Operating cost is reported per 24-hour episode in \pounds M; reserve and thermal columns are per-step violation rates.}
  \label{tab:tso-frontier-appendix}
  \centering
  \footnotesize
  \setlength{\tabcolsep}{3pt}
  \begin{tabularx}{\linewidth}{llrrr}
    \toprule
    Split & Method & Operating cost (\pounds M/day) & Reserve-shortfall rate & Thermal-overload rate \\
    \midrule
    in-distribution  & PPO         & 1.58 & 0.0728 & 0.3554 \\
    in-distribution  & Merit Order & 2.89 & 0.0096 & 0.1242 \\
    in-distribution  & PPO-Lag     & 3.50 & 0.0000 & 0.0593 \\
    in-distribution  & All-On      & 4.34 & 0.0000 & 0.0346 \\
    line-tightening  & PPO         & 1.57 & 0.0547 & 0.4932 \\
    line-tightening  & Merit Order & 2.89 & 0.0096 & 0.3371 \\
    line-tightening  & PPO-Lag     & 3.50 & 0.0000 & 0.1560 \\
    line-tightening  & All-On      & 4.34 & 0.0000 & 0.0504 \\
    load-stress      & PPO         & 1.82 & 0.0894 & 0.4118 \\
    load-stress      & Merit Order & 3.31 & 0.0183 & 0.1675 \\
    load-stress      & PPO-Lag     & 3.78 & 0.0000 & 0.0889 \\
    load-stress      & All-On      & 4.68 & 0.0000 & 0.0754 \\
    \bottomrule
  \end{tabularx}
\end{table}

\begin{figure}[!htbp]
  \centering
  \includegraphics[width=0.95\linewidth]{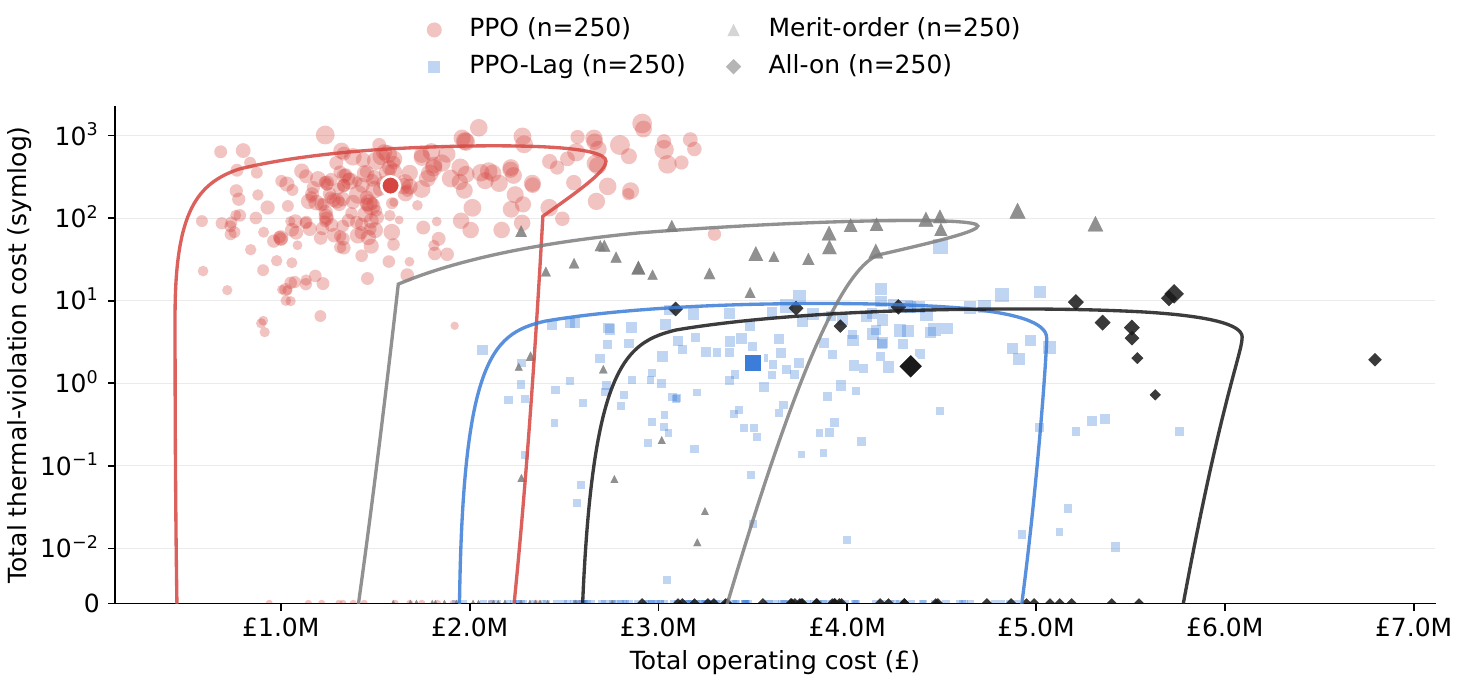}
  \caption{TSO per-episode operating cost vs thermal-overload cost on the in-distribution split (4 algorithms $\times$ 5 seeds $\times$ 50 episodes $=$ 1{,}000 points). Marker area is proportional to the per-episode thermal-violation rate; the ellipse around each method covers its 95\% covariance region.}
  \label{fig:tso-episode-scatter-appendix}
\end{figure}

Figure~\ref{fig:tso-episode-scatter-appendix} expands the method-level numbers in Table~\ref{tab:tso-frontier-appendix} into per-episode dispersion. PPO has the lowest mean operating cost (\pounds1.58M per 24-hour episode) but a wide spread along the thermal-cost axis: its mean per-episode total thermal cost is $247.67$, and the worst-decile episodes carry thermal cost more than an order of magnitude above this mean. PPO-Lagrangian shifts the cloud toward higher cost (\pounds3.50M per episode) while compressing mean thermal cost to $1.76$, roughly two orders of magnitude below PPO. Merit Order and All-On occupy the high-cost low-thermal corner.

\begin{figure}[!htbp]
  \centering
  \includegraphics[width=1.00\linewidth]{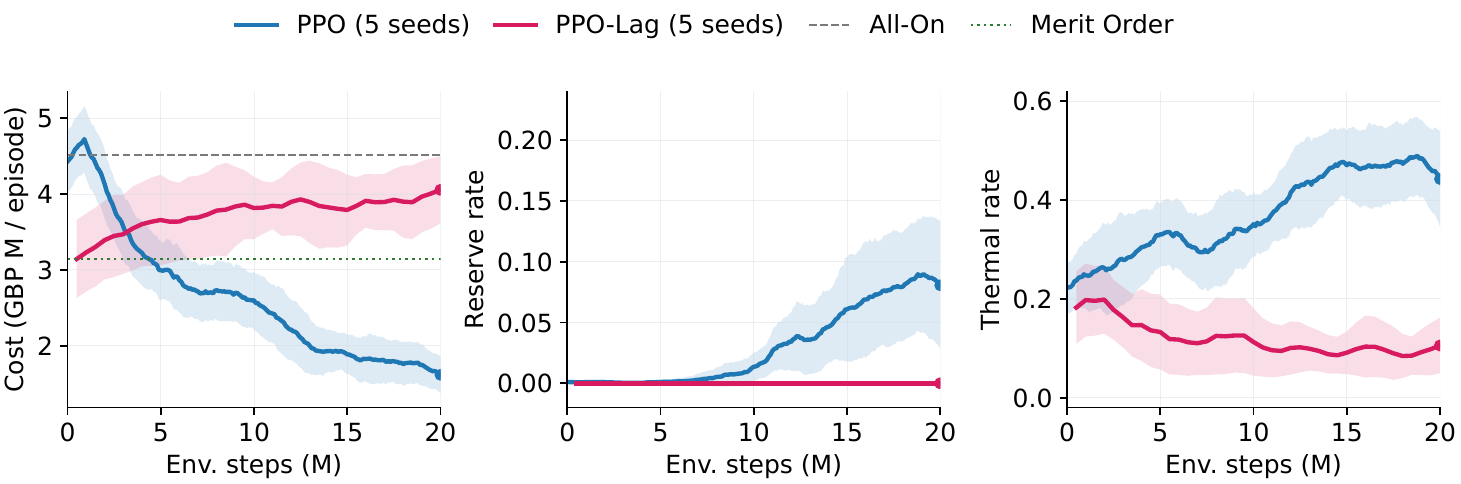}
  \caption{TSO train-split checkpoint monitor over 20\,M environment steps. Panels show evaluation operating cost, reserve-shortfall incidence, and thermal-overload incidence. Curves are smoothed means over five seeds; shaded bands show one standard deviation.}
  \label{fig:tso-learning-curves-appendix}
\end{figure}

Figure~\ref{fig:tso-learning-curves-appendix} complements Table~\ref{tab:tso-frontier-appendix} with train-split checkpoint monitoring curves over the 20\,M-step training budget. The curves aggregate the five seeds and show smoothed mean $\pm$ one standard deviation over sparse greedy-evaluation checkpoints. PPO drives operating cost down toward its held-out in-distribution value of \pounds1.58M per episode (Table~\ref{tab:tso-frontier-appendix}), but its safety panels move in the opposite direction, with both reserve-shortfall and thermal-overload incidence rising over training. PPO-Lagrangian holds reserve-shortfall incidence at zero throughout and lowers final thermal-overload incidence well below PPO, while its operating cost stays above both PPO and the merit-order reference. The learning dynamics therefore explain how the cost--safety frontier is reached.

\FloatBarrier

\subsection{DSO}
\label{app:I.3}

DSO reports raw in-distribution physical metrics over 5 seeds $\times$ 50 evaluation episodes; the voltage channel is reported under Form~1 with $b_i{=}0$ (Section~\ref{app:C.2}). Loss reduction in Table~\ref{tab:dso-iid-appendix} is computed relative to no control.

\begin{table}[h]
  \caption{DSO in-distribution physical metrics. Lower total network loss is better; the voltage column is the per-step count of buses outside the $[0.94,1.06]$~p.u.\ band.}
  \label{tab:dso-iid-appendix}
  \centering
  \small
  \begin{tabular}{lrrr}
    \toprule
    Method & Total loss (MWh) & Voltage violations / step & Loss reduction \\
    \midrule
    PPO        & 1.9355 & 0.00033 & 34.45\% \\
    Sauté PPO  & 1.9370 & 0.00292 & 34.35\% \\
    SAC        & 2.3178 & 0.05475 & 20.54\% \\
    PPO-Lag    & 2.4108 & 0.01733 & 15.84\% \\
    Droop      & 2.8593 & 0.07042 &  0.74\% \\
    No control & 2.8909 & 0.20667 &  0.00\% \\
    \bottomrule
  \end{tabular}
\end{table}

\begin{figure}[!htbp]
  \centering
  \includegraphics[width=0.95\linewidth]{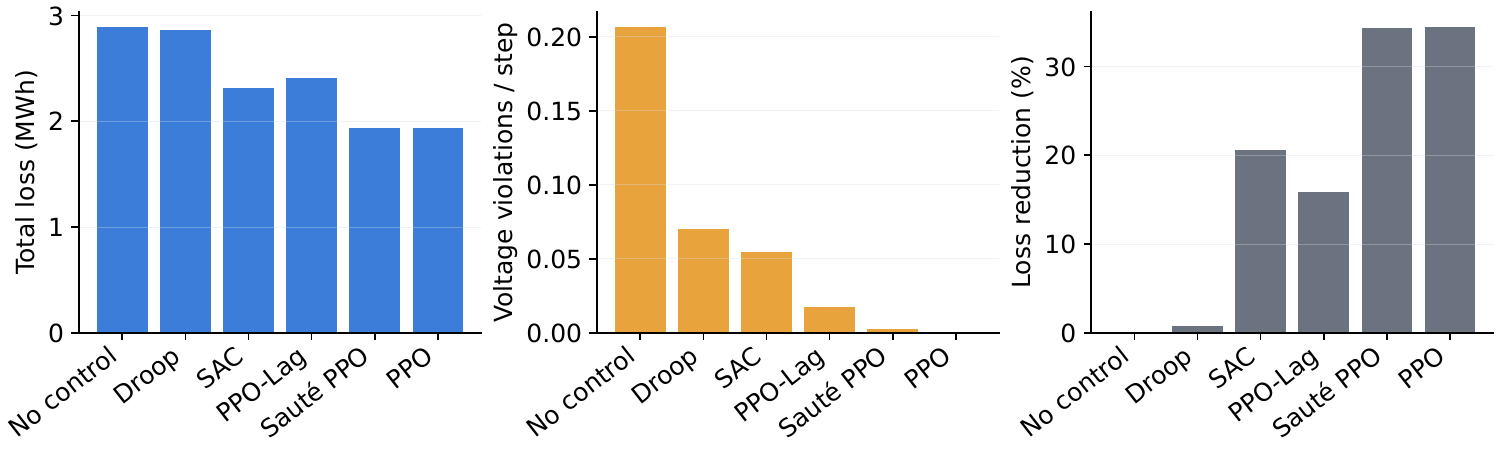}
  \caption{DSO in-distribution physical metrics. Panels report total network loss, voltage-violation count per step, and loss reduction relative to no control.}
  \label{fig:dso-iid}
\end{figure}

Figure~\ref{fig:dso-iid} visualizes the three columns of Table~\ref{tab:dso-iid-appendix} as bar panels.

\begin{figure}[!htbp]
  \centering
  \includegraphics[width=1.00\linewidth]{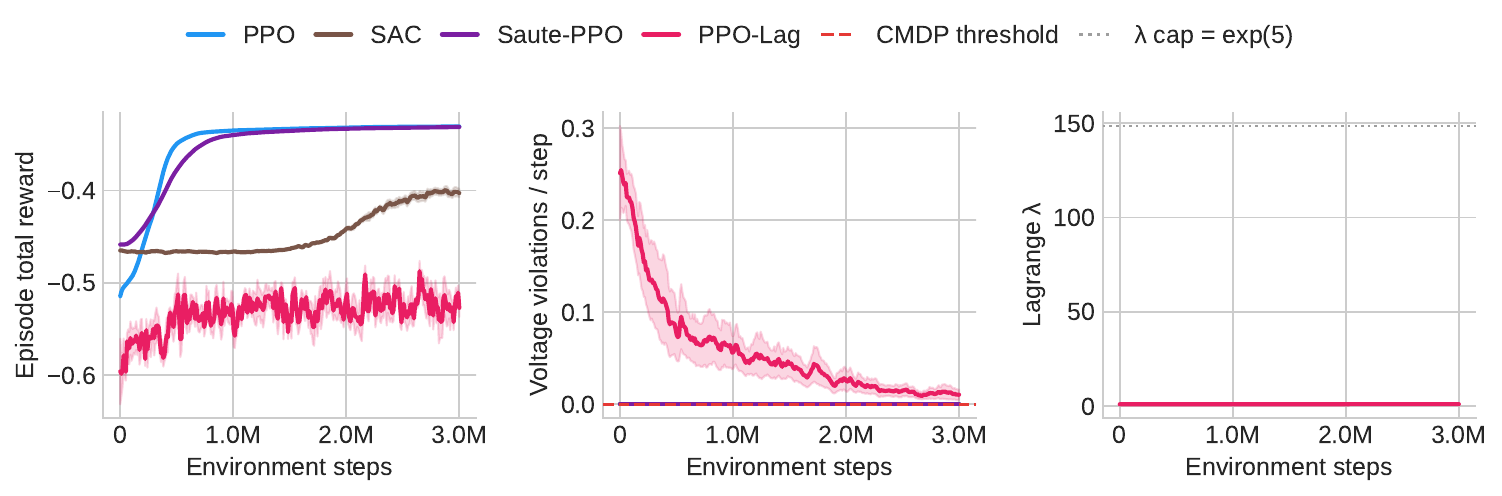}
  \caption{DSO learning curves over 3\,M environment steps. Left: episode total reward across the four learned algorithms. Middle: per-step voltage-violation count. Right: PPO-Lagrangian Lagrange multiplier $\boldsymbol{\lambda}$.}
  \label{fig:dso-learning-curves-appendix}
\end{figure}

\begin{figure}[!htbp]
  \centering
  \includegraphics[width=1.00\linewidth]{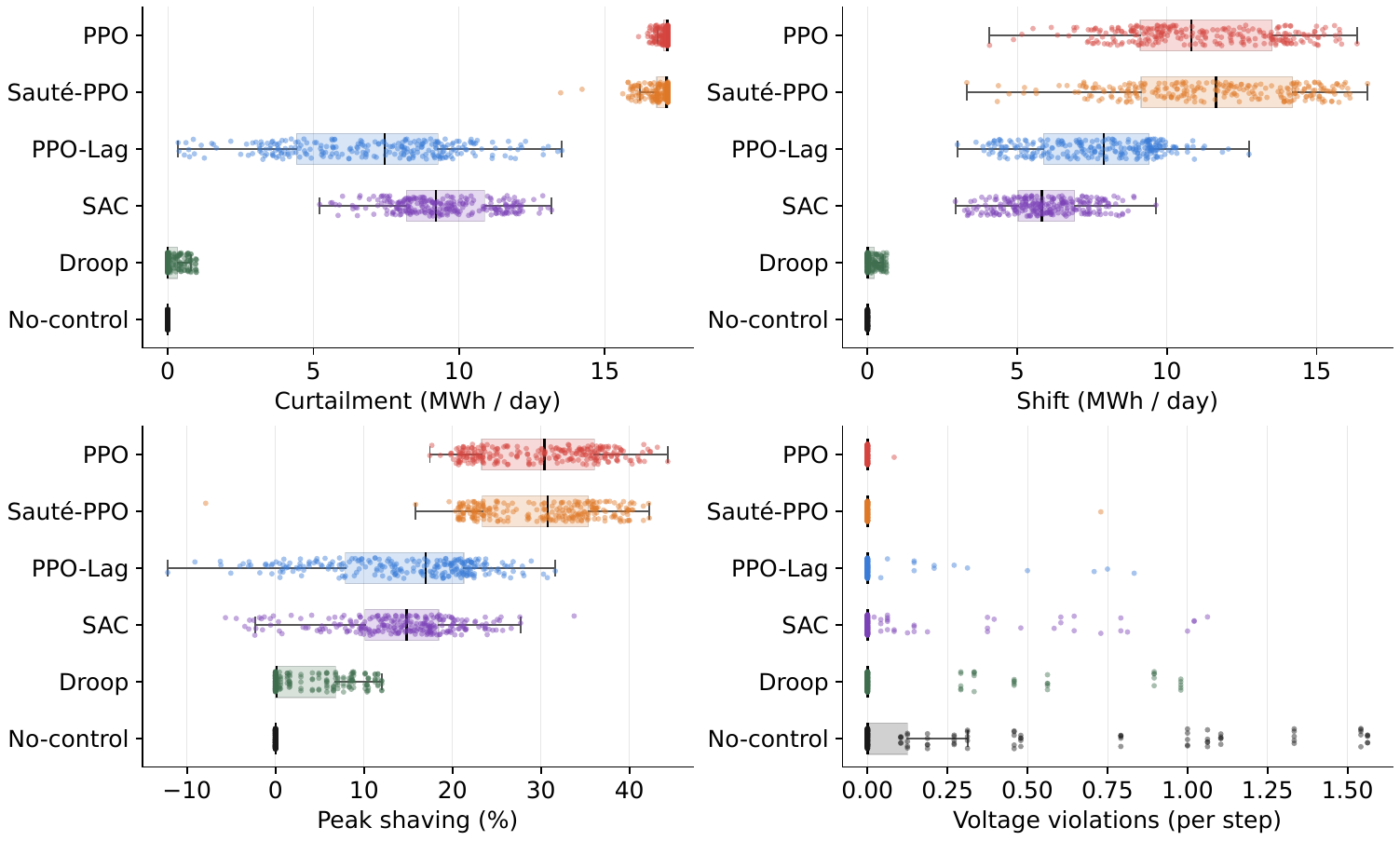}
  \caption{DSO per-episode in-distribution behavior distributions over 6 algorithms $\times$ 5 seeds $\times$ 50 episodes. Each panel reports a single behavior metric; boxes are overlaid by jittered individual episodes so distributional tails (especially the voltage-violation panel) remain visible.}
  \label{fig:dso-episode-distributions-appendix}
\end{figure}

Figure~\ref{fig:dso-episode-distributions-appendix} expands Table~\ref{tab:dso-iid-appendix} into per-episode behavior distributions over the six methods, with each box overlaid by the 250 individual episodes so distributional tails remain visible. PPO and Sauté PPO use roughly $17$\,MWh per day of curtailment and $11$\,MWh per day of shifting, achieving close to $30\%$ peak shaving with voltage violations driven to near zero on most episodes; tail steps are still visible above zero in the bottom-right panel. PPO-Lagrangian curtails and shifts substantially less, yet has a heavier voltage-violation tail than PPO. SAC reaches lower total loss than PPO-Lagrangian on every paired episode but carries the highest per-step voltage-violation rate of the four learned methods. Droop's curtailment and shifting stay well below $1$\,MWh per day, an order of magnitude under the four learned policies.

\paragraph{PPO-Lagrangian training dynamics.}
Under no control the in-distribution DSO carries roughly $0.21$ voltage violations per step (Table~\ref{tab:dso-iid-appendix}). PPO and Sauté PPO drive the in-distribution rate to near zero within the first half of the training budget; once the per-step rate is small, the constraint signal that the dual update consumes becomes sparse, and PPO-Lagrangian converges to a conservative region with low loss reduction and a heavier violation tail than PPO. The corresponding learning curves are shown in Figure~\ref{fig:dso-learning-curves-appendix}.

\FloatBarrier

\subsection{DERs}
\label{app:I.4}

On the in-distribution split, every learned and non-learning method reports zero or near-zero voltage violations (Table~\ref{tab:ders-iid-appendix}). Across all five algorithms and all five evaluation splits, the maximum bus voltage stays at the slack reference $V_{\max}=1.000$~p.u., so the safe band is exercised only on its under-voltage side and the operating regime never enters PV-driven over-voltage; this remains true under the PV-shift split, where the PV bundle capacity is doubled. The voltage-tightening split (Table~\ref{tab:ders-tightening-appendix}), which narrows the band from $[0.94, 1.06]$~p.u.\ to $[0.96, 1.04]$~p.u., is therefore the split on which the methods separate on safety.

For the in-distribution loss test, IPPO and IPPO-rs reduce active-power loss to $0.1974$~MW versus $0.2052$~MW for no control --- a $3.8\%$ reduction --- and beat both no control and voltage droop at $p<10^{-4}$ under the paired sign-flip test of Section~\ref{app:H.5}. IPPO-Lagrangian has higher loss ($0.2051$~MW): it is statistically indistinguishable from no control on the loss test ($p{=}0.87$) and is statistically worse than voltage droop (mean difference $+0.0020$~MW, $p<10^{-4}$).

\begin{figure}[!htbp]
  \centering
  \includegraphics[width=0.92\linewidth]{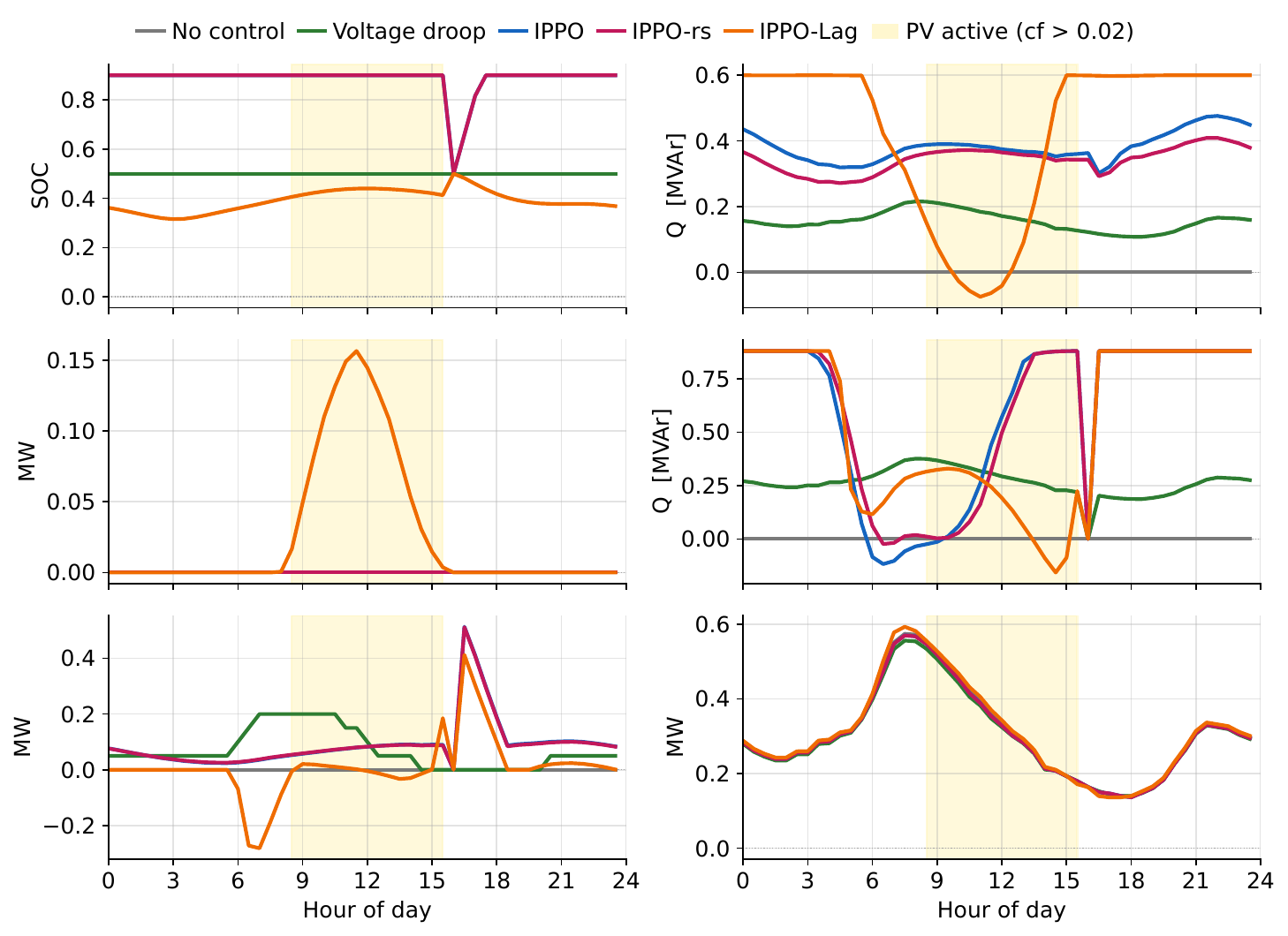}
  \caption{DERs schedules on a single in-distribution episode. Panels compare battery state of charge, battery reactive power commands, PV curtailment, PV reactive-power commands, flexible load actions, and active power loss across IPPO, IPPO-rs, IPPO-Lagrangian, voltage droop, and no control.}
  \label{fig:ders-policy-appendix}
\end{figure}

Under tightening, IPPO and IPPO-rs cut the violation count by $44\%$ relative to no control ($4.91$ vs $8.73$ steps per episode); voltage droop cuts it by $32\%$ ($5.97$); IPPO-Lagrangian cuts it by $9\%$ ($7.94$), so the constrained variant trails IPPO and IPPO-rs on both the in-distribution loss test and the tightening split. The load-stress split shows the same ordering qualitatively: per-episode violation steps are $0.23$ for IPPO and IPPO-rs, $0.33$ for voltage droop, $0.83$ for IPPO-Lagrangian, and $0.93$ for no control. Figure~\ref{fig:ders-policy-appendix} traces battery state of charge, battery reactive-power command, photovoltaic curtailment, photovoltaic reactive-power command, flexible-load action, and active-power loss for each policy on the same in-distribution episode, showing that IPPO and IPPO-rs realize their loss reduction through coordinated reactive-power dispatch rather than active-power curtailment, while IPPO-Lagrangian holds the cost critic near its budget and dispatches less aggressively on both channels.

\begin{table}[h]
  \caption{DERs in-distribution results. Lower active-power loss is better. Loss values are mean across $5$ seeds $\times$ $30$ episodes.}
  \label{tab:ders-iid-appendix}
  \centering
  \footnotesize
  \begin{tabular}{lrrr}
    \toprule
    Method & Active loss (MW) & Violation steps & Violation rate \\
    \midrule
    IPPO-rs       & 0.1974 & 0.00   & 0.0000  \\
    IPPO          & 0.1974 & 0.00   & 0.0000  \\
    Voltage droop & 0.2031 & 0.00   & 0.0000  \\
    IPPO-Lag      & 0.2051 & 0.0067 & 0.00014 \\
    No control    & 0.2052 & 0.00   & 0.0000  \\
    \bottomrule
  \end{tabular}
\end{table}

\begin{table}[h]
  \caption{DERs voltage-tightening stress split (band tightened from $[0.94,1.06]$ to $[0.96,1.04]$~p.u.). Violation steps are per episode; rates use the $48$-step horizon as denominator. Active loss matches the in-distribution column because tightening changes only the constraint band, not dispatch decisions.}
  \label{tab:ders-tightening-appendix}
  \centering
  \footnotesize
  \begin{tabular}{lrrr}
    \toprule
    Method & Active loss (MW) & Violation steps & Violation rate \\
    \midrule
    IPPO-rs       & 0.1974 & 4.91 & 0.1022 \\
    IPPO          & 0.1974 & 4.91 & 0.1024 \\
    Voltage droop & 0.2031 & 5.97 & 0.1243 \\
    IPPO-Lag      & 0.2051 & 7.94 & 0.1654 \\
    No control    & 0.2052 & 8.73 & 0.1819 \\
    \bottomrule
  \end{tabular}
\end{table}

\FloatBarrier

\subsection{DCMG}
\label{app:I.5}

\begin{figure}[!htbp]
  \centering
  \includegraphics[width=1.00\linewidth]{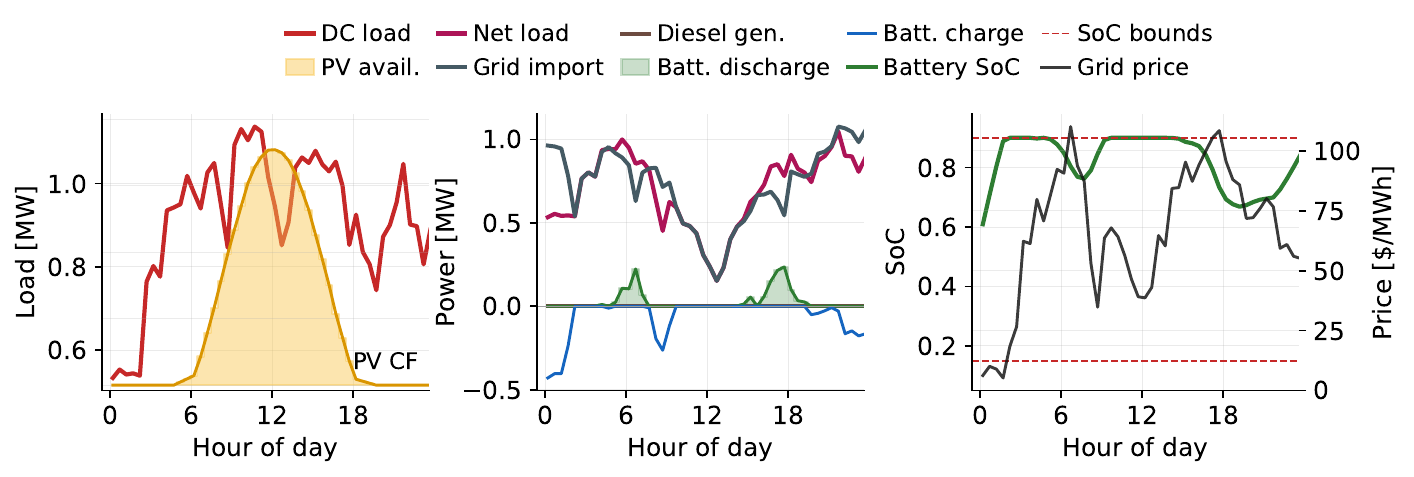}
  \caption{Representative DCMG dispatch from the SAC policy on one in-distribution episode, displayed as 30-minute means for readability. \textbf{Left:} real GB solar capacity factor and the data center IT load. \textbf{Middle:} net load after PV, grid import, diesel output, and battery charge/discharge. \textbf{Right:} battery state of charge with SoC bounds, alongside the GB MID grid price.}
  \label{fig:dc-dispatch-appendix}
\end{figure}

DCMG (Section~\ref{app:E.5}) is reported on the in-distribution split with all three cost channels (SLA, over-temperature, power deficit) at a zero threshold. Episode returns are in Table~\ref{tab:dc-iid-appendix}. Figure~\ref{fig:dc-dispatch-appendix} shows a representative SAC dispatch on one in-distribution episode; the same SAC trajectory underlies main-text Section~\ref{sec:6.4}.

PPO records zero violations on all three channels; SAC and the three heuristic methods show SLA rates of $10^{-7}$ to $10^{-6}$ and zero on the other two channels.

\begin{table}[h]
  \caption{DCMG in-distribution results. Higher episode return is better. SLA rate is the per-step rate of expired tasks; spill cost is the per-step over-generation penalty. The power-deficit and over-temperature rates are $0$ for every method and are not shown.}
  \label{tab:dc-iid-appendix}
  \centering
  \footnotesize
  \begin{tabular}{lrrr}
    \toprule
    Method & Episode return & SLA rate & Spill cost \\
    \midrule
    SAC              & $-2594.59$  & $4.17{\times}10^{-7}$ & $9.05{\times}10^{-5}$ \\
    PPO              & $-2845.29$  & $0$                  & $1.03{\times}10^{-3}$ \\
    No control       & $-3110.68$  & $1.11{\times}10^{-6}$ & $0$ \\
    Rule-based       & $-5226.24$  & $8.33{\times}10^{-7}$ & $4.58{\times}10^{-4}$ \\
    Max renewable    & $-6010.68$  & $1.39{\times}10^{-7}$ & $6.30{\times}10^{-3}$ \\
    \bottomrule
  \end{tabular}
\end{table}

\FloatBarrier

\section{Execution Scaling and Backend Comparison}
\label{app:J}

This section decomposes the suite-level speed result of Section~\ref{sec:6.1} along three axes: backend architecture, parallelism range, and wall-clock learning efficiency.

\subsection{Backend decomposition at \texorpdfstring{$N_{\rm env}=2^{12}$}{Nenv=2\string^12}}
\label{app:J.1}

The three training backends differ in where the environment transition runs. SB3 runs a PyTorch policy on the GPU and the PowerZooPy environment on the CPU. SBX runs a JAX policy and optimizer on the GPU but keeps the same CPU-hosted PowerZooPy environment. PowerZooJax runs the policy, optimizer, and environment transition end-to-end inside a compiled JAX program on the GPU. Table~\ref{tab:execution-endpoint-backend} decomposes throughput at $N_{\rm env}=2^{12}$ across these three stacks: the SB3-to-SBX delta isolates the policy-side acceleration (PyTorch~$\to$~JAX), and the SBX-to-PowerZooJax delta, reported in the rightmost column ``$\times$ vs SBX'', isolates the environment-side acceleration. The environment-side gain alone is $1{,}145\times$ on GenCos and $2{,}040\times$ on DERs, both of which call CPU-hosted OPF or power-flow solvers in PowerZooPy that PowerZooJax fuses into the compiled JAX graph. The gain is smaller on DSO ($39\times$) and TSO ($33\times$), whose CPU steps already run JAX-friendly kernels rather than an external solver. DCMG ($22\times$) is the smallest case: its step is dominated by small arithmetic device updates without an external solver call, so the CPU-hosted environment is already fast.

\begin{table}[h]
  \caption{Throughput at $N_{\rm env}=2^{12}$ parallel environments, with the environment-side speedup isolated. The headline speedup against the slower CPU baseline is in main-text Table~\ref{tab:speed}.}
  \label{tab:execution-endpoint-backend}
  \centering
  \footnotesize
  \setlength{\tabcolsep}{4pt}
  \begin{tabular}{lrrrr}
    \toprule
    Task & PowerZooJax & SBX/CUDA & SB3/CUDA & $\times$ vs SBX \\
    \midrule
    GenCos & 506{,}134 & 442     & 244     & 1{,}145$\times$ \\
    DERs   & 414{,}506 & 203     & 233     & 2{,}040$\times$ \\
    DSO    & 134{,}660 & 3{,}435 & 2{,}207 & 39$\times$ \\
    TSO    &  53{,}724 & 1{,}649 & 411     & 33$\times$ \\
    DCMG   &  25{,}132 & 1{,}167 & 1{,}298 & 22$\times$ \\
    \bottomrule
  \end{tabular}
\end{table}

\subsection{Per-task throughput scaling}
\label{app:J.2}

\begin{figure}[t]
  \centering
  \includegraphics[width=\linewidth]{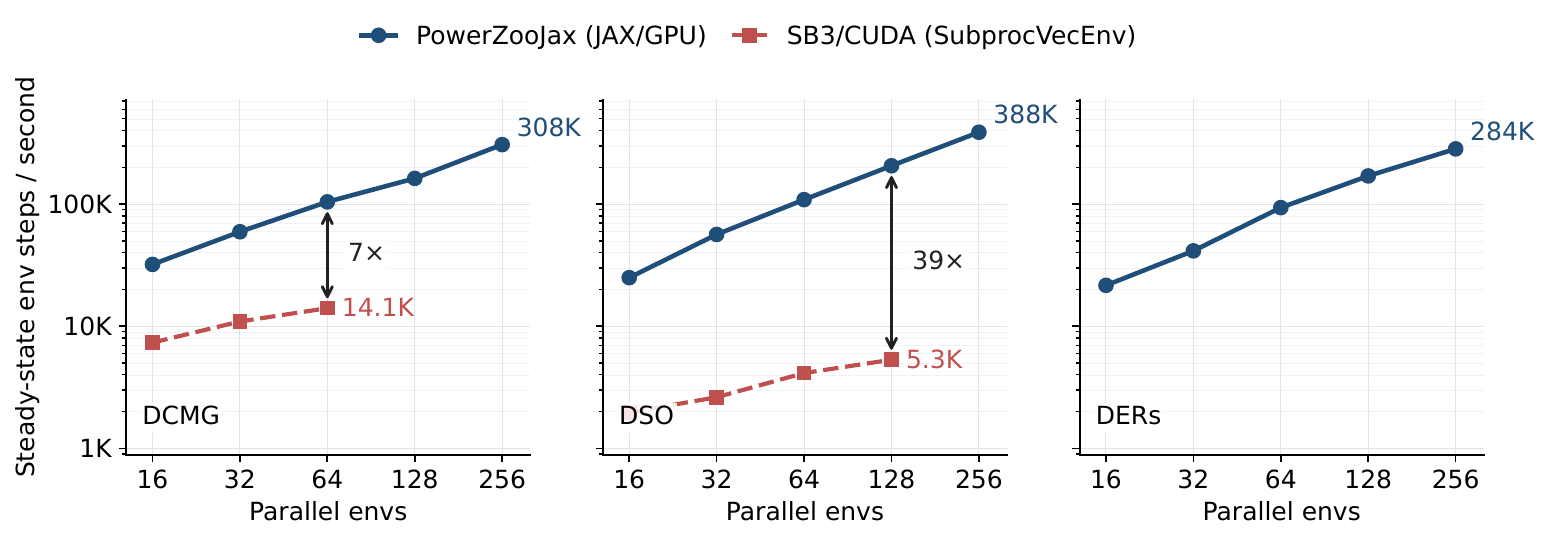}
  \caption{Per-task throughput scaling. PowerZooJax JAX/GPU is plotted over $\text{nenv}\in\{16, 32, 64, 128, 256\}$ for every task; SB3/CUDA is overlaid where its \texttt{SubprocVecEnv} backend completes (through $\text{nenv}=64$ for DCMG, through $\text{nenv}=128$ for DSO; DERs has no paired SB3 sweep). Endpoint labels mark the JAX/GPU and SB3/CUDA throughput at the right end of each curve; arrows mark the JAX-vs-SB3 ratio at the top of each matched range.}
  \label{fig:execution-scaling-combined}
\end{figure}

Figure~\ref{fig:execution-scaling-combined} reports throughput against parallelism for DCMG, DSO, and DERs. PowerZooJax JAX/GPU is swept over $\text{nenv}\in\{16, 32, 64, 128, 256\}$ on all three; SB3/CUDA is overlaid for DCMG and DSO under the same warm-up and measurement protocol over the range where its \texttt{SubprocVecEnv} backend completes. JAX/GPU throughput rises with parallelism on every task: DCMG from $32.1$k to $307.7$k env steps/s, DSO from $25.0$k to $388.4$k, and DERs from $21.6$k to $283.8$k. The matched-range speedup over SB3/CUDA reaches $7.4\times$ for DCMG (at $\text{nenv}=64$) and $39\times$ for DSO (at $\text{nenv}=128$), the largest parallelism each SB3 path completes; SB3 throughput tops out at $14.1$k on DCMG and $5.3$k on DSO. The high-parallel endpoint in Table~\ref{tab:execution-endpoint-backend} reports $25.1$k, $134.7$k, and $414.5$k at $N_{\rm env}=2^{12}$ for DCMG, DSO, and DERs respectively under a separate measurement protocol.

\subsection{First-compile cost}
\label{app:J.3}

The scaling sweeps record the first JAX compilation separately from the steady-state update loop. Compile cost is single-digit seconds for the three tasks in Figure~\ref{fig:execution-scaling-combined} (Table~\ref{tab:jax-compile-time}); subsequent updates run from the cached compiled program.

\begin{table}[h]
  \caption{First JAX compilation time in task-specific scaling sweeps. Values are means over seeds.}
  \label{tab:jax-compile-time}
  \centering
  \small
  \begin{tabular}{lrrr}
    \toprule
    Task & Measured nenv range & Compile time & JAX throughput at $\text{nenv}{=}256$ \\
    \midrule
    DCMG & 16--256 & 8.2\,s & 307.7k steps/s \\
    DSO  & 16--256 & 2.9\,s & 388.4k steps/s \\
    DERs & 16--256 & 2.1\,s & 283.8k steps/s \\
    \bottomrule
  \end{tabular}
\end{table}

\subsection{Wall-clock learning curves for DSO and TSO}
\label{app:J.4}

\begin{figure}[t]
  \centering
  \includegraphics[width=0.95\linewidth]{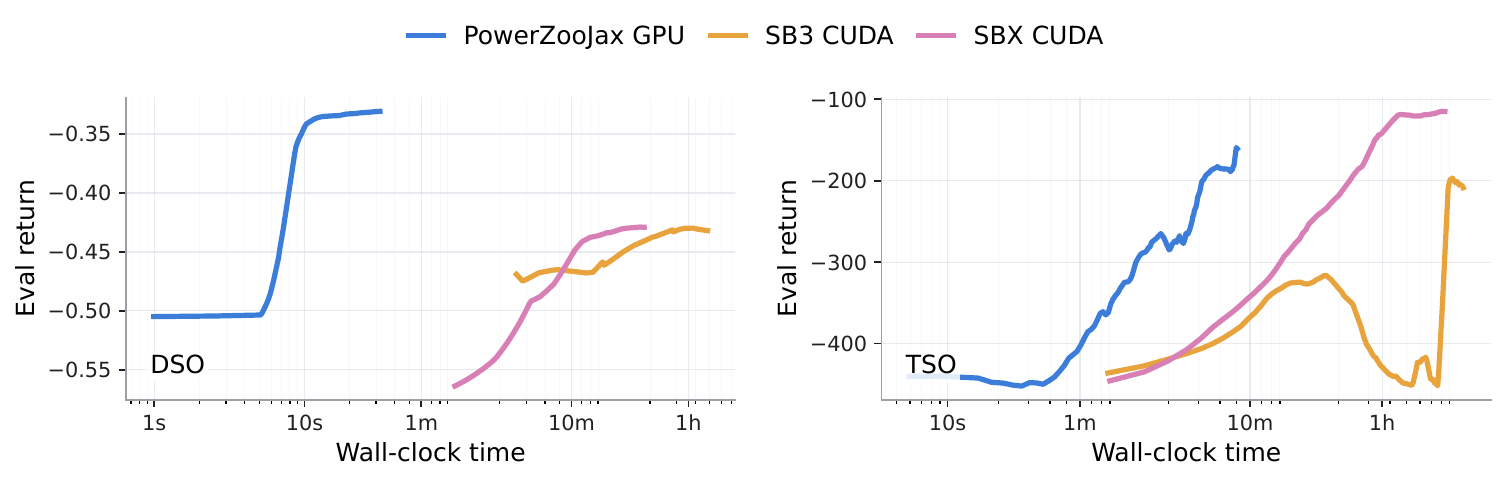}
  \caption{Evaluation return against elapsed wall-clock time for DSO and TSO across PowerZooJax, SBX/CUDA, and SB3/CUDA.}
  \label{fig:wallclock-eval-return-clean}
\end{figure}

For DSO and TSO, all three backends (PowerZooJax, SBX/CUDA, SB3/CUDA) ran to the same training budget and recorded per-evaluation-point wall-clock timestamps, which allows learning curves to be plotted against time rather than environment steps. Figure~\ref{fig:wallclock-eval-return-clean} traces evaluation return against elapsed time for the three backends on these two tasks at the matched budgets of main-text Table~\ref{tab:speed}. The PowerZooJax run reaches the same training budget in a fraction of the time required by either CPU-hosted backend.

\end{document}